\documentclass[12pt,hidelinks]{article}
	\usepackage[explicit]{titlesec}
	\usepackage{titletoc}
	\usepackage{tocloft}
	\usepackage{charter}
	\usepackage[many]{tcolorbox}
	\usepackage{amsfonts, mathtools, amsmath}
        \usepackage{multirow, multicol, makecell} 
        
	\usepackage{graphicx}
	\usepackage{xcolor}
	\usepackage{tikz,lipsum,lmodern}
	\usetikzlibrary{calc}
	\usepackage[english]{babel}
	\usepackage{fancyhdr}
	\usepackage{mathrsfs}
	\usepackage{empheq}
	\usepackage{fourier}% change to lmodern if fourier is no available
	\usepackage{wrapfig}
	\usepackage{fancyref}
	\usepackage{hyperref}
	\usepackage{cleveref}
	\usepackage{listings}
	\usepackage{varwidth}
	\usepackage{longfbox}
	\usepackage{geometry}
	\usepackage{marginnote}

	\tcbuselibrary{theorems}
	\tcbuselibrary{breakable, skins}
	\tcbuselibrary{listings, documentation}
\graphicspath{{./images/}}
\addto{\captionsenglish}{}
\renewcommand{\thesection}{\arabic{section}}
\newcommand\sectionnumfont{% font specification for the number
	\fontsize{380}{130}\color{myblueii}\selectfont}
\newcommand\sectionnamefont{% font specification for the name "PART"
	\normalfont\color{white}\scshape\small\bfseries }
\definecolor{mordantred19}{rgb}{0.68, 0.05, 0.0}
\definecolor{st.patrick\'sblue}{rgb}{0.14, 0.16, 0.48}
\definecolor{teal}{rgb}{0.0, 0.5, 0.5}
\definecolor{beaublue}{rgb}{0.74, 0.83, 0.9}
\definecolor{mybluei}{RGB}{0,173,239}
\definecolor{myblueii}{RGB}{63,200,244}
\definecolor{myblueiii}{RGB}{199,234,253}
\definecolor{blond}{rgb}{0.98, 0.94, 0.75}
\definecolor{cream}{rgb}{1.0, 0.99, 0.82}
\definecolor{emerald}{rgb}{0.31, 0.78, 0.47}
\definecolor{darkspringgreen}{rgb}{0.09, 0.45, 0.27}
\definecolor{ghostwhite}{rgb}{0.97, 0.97, 1.0}
\definecolor{splashedwhite}{rgb}{1.0, 0.99, 1.0}
\definecolor{whitesmoke}{rgb}{0.96, 0.96, 0.96}
\definecolor{lightgray}{rgb}{0.92, 0.92, 0.92}
\definecolor{floralwhite}{rgb}{1.0, 0.98, 0.94}
\titleformat{\section}
{\normalfont\huge\filleft}
{}
{20pt}
{\begin{tikzpicture}[remember picture,overlay]
	\fill[myblueiii] 
	(current page.north west) rectangle ([yshift=-13cm]current page.north east);   
\node[
	fill=mybluei,
	text width=2\paperwidth,
	rounded corners=6cm,
	text depth=18cm,
	anchor=center,
	inner sep=0pt] at (current page.north east) (parttop)
	{\thepart};%
\node[
	anchor=south east,
	inner sep=0pt,
	outer sep=0pt] (partnum) at ([xshift=-20pt]parttop.south) 
	{\sectionnumfont\thesection};
\node[
	anchor=south,
	inner sep=0pt] (partname) at ([yshift=2pt]partnum.south)   
	{\sectionnamefont SECTION};
\node[
	anchor=north east,
	align=right,
	inner xsep=0pt] at ([yshift=-0.5cm]partname.east|-partnum.south) 
	{\parbox{.7\textwidth}{\raggedleft#1}};
\end{tikzpicture}%
}
\hypersetup{
	colorlinks,
	linkcolor={red!50!black},
	citecolor={blue!50!black},
	urlcolor={blue!80!black}
}
\tcbset{
	defstyle/.style={
		fonttitle=\bfseries\upshape, 
		fontupper=\slshape,
		arc=0mm, 
		beamer,
		colback=blue!5!white,
		colframe=blue!75!black},
	theostyle/.style={
		fonttitle=\bfseries\upshape, 
		fontupper=\slshape,
		colback=red!10!white,
		colframe=red!75!black},
	visualstyle/.style={
		height=6.5cm,
		breakable,
		enhanced,
		leftrule=0pt,
		rightrule=0pt,
		bottomrule=0pt,
		outer arc=0pt,
		arc=0pt,
		colframe=mordantred19,
		colback=lightgray,
		attach boxed title to top left,
		boxed title style={
			colback=mordantred19,
			outer arc=0pt,
			arc=0pt,
			top=3pt,
			bottom=3pt,
		},
		fonttitle=\sffamily,},
	discussionstyle/.style={
		height=6.5cm,
		breakable,
		enhanced,
		rightrule=0pt,
		toprule=0pt,
		outer arc=0pt,
		arc=0pt,
		colframe=mordantred19,
		colback=lightgray,
		attach boxed title to top left,
		boxed title style={
			colback=mordantred19,
			outer arc=0pt,
			arc=0pt,
			top=3pt,
			bottom=3pt,
		},
		fonttitle=\sffamily},
	mystyle/.style={
		height=6.5cm,
		breakable,
		enhanced,
		rightrule=0pt,
		leftrule=0pt,
		bottomrule=0pt,
		outer arc=0pt,
		arc=0pt,
		colframe=mordantred19,
		colback=lightgray,
		attach boxed title to top left,
		boxed title style={
			colback=mordantred19,
			outer arc=0pt,
			arc=0pt,
			top=3pt,
			bottom=3pt,
		},
		fonttitle=\sffamily},
	aastyle/.style={
			height=3.5cm,
			enhanced,
			colframe=teal,
			colback=lightgray,
			colbacktitle=floralwhite,
			fonttitle=\bfseries,
			coltitle=black,
		attach boxed title to top center={
	  		yshift=-0.25mm-\tcboxedtitleheight/2,
	   		yshifttext=2mm-\tcboxedtitleheight/2}, 
		boxed title style={boxrule=0.5mm,
			frame code={ \path[tcb fill frame] ([xshift=-4mm]frame.west)
				-- (frame.north west) -- (frame.north east) -- ([xshift=4mm]frame.east)
				-- (frame.south east) -- (frame.south west) -- cycle; },
			interior code={ 
				\path[tcb fill interior] ([xshift=-2mm]interior.west)
				-- (interior.north west) -- (interior.north east)
				-- ([xshift=2mm]interior.east) -- (interior.south east) -- (interior.south west)
				-- cycle;} }
				},
	examstyle/.style={
		height=9.5cm,
		breakable,
		enhanced,
		rightrule=0pt,
		leftrule=0pt,
		bottomrule=0pt,
		outer arc=0pt,
		arc=0pt,
		colframe=mordantred19,
		colback=lightgray,
		attach boxed title to top left,
		boxed title style={
			colback=mordantred19,
			outer arc=0pt,
			arc=0pt,
			top=3pt,
			bottom=3pt,
		},
		fonttitle=\sffamily},
	doc head command={
		interior style={
			fill,
			left color=yellow!20!white, 
			right color=white}},
	doc head environment={
		boxsep=4pt,
		arc=2pt,
		colback=yellow!30!white,
		},
	doclang/environment content=text
}
\newtcolorbox[auto counter,number within=section]{example}[1][]{
	mystyle,
	title=Example~\thetcbcounter,
	overlay unbroken and first={
		\path
		let
		\p1=(title.north east),
		\p2=(frame.north east)
		in
		node[anchor=
			west,
			font=\sffamily,
			color=st.patrick\'sblue,
			text width=\x2-\x1] 
		at (title.east) {#1};
	}
}
\newtcolorbox[auto counter,number within=section]{longexample}[1][]{
	examstyle,
	title=Example~\thetcbcounter,
	overlay unbroken and first={
		\path
		let
		\p1=(title.north east),
		\p2=(frame.north east)
		in
		node[anchor=
		west,
		font=\sffamily,
		color=st.patrick\'sblue,
		text width=\x2-\x1] 
		at (title.east) {#1};
	}
}
\newtcolorbox[auto counter,number within=section]{example2}[1][]{
	aastyle,
	title=Example~\thetcbcounter,{}
}
\newtcolorbox[auto counter,number within=section]{discussion}[1][]{
	discussionstyle,
	title=Discussion~\thetcbcounter,
	overlay unbroken and first={
		\path
		let
		\p1=(title.north east),
		\p2=(frame.north east)
		in
		node[anchor=
		west,
		font=\sffamily,
		color=st.patrick\'sblue,
		text width=\x2-\x1] 
		at (title.east) {#1};
	}
}
\newtcolorbox[auto counter,number within=section]{visualization}[1][]{
	visualstyle,
	title=Visualization~\thetcbcounter,
	overlay unbroken and first={
		\path
		let
		\p1=(title.north east),
		\p2=(frame.north east)
		in
		node[anchor=
		west,
		font=\sffamily,
		color=st.patrick\'sblue,
		text width=\x2-\x1] 
		at (title.east) {#1};
	}
}
\newtcbtheorem[number within=subsection,crefname={definition}{definitions}]%
	{Definition}{Definition}{defstyle}{def}%
\newtcbtheorem[use counter from=Definition,crefname={theorem}{theorems}]%
	{Theorem}{Theorem}{theostyle}{theo}
\newtcbtheorem[use counter from=Definition]{theo}{Theorem}%
{
	theorem style=plain,
	enhanced,
	colframe=blue!50!black,
	colback=yellow!20!white,
	coltitle=red!50!black,
	fonttitle=\upshape\bfseries,
	fontupper=\itshape,
	drop fuzzy shadow=blue!50!black!50!white,
	boxrule=0.4pt}{theo}
\newtcbtheorem[use counter from=Definition]{DashedDefinition}{Definition}%
 {
 	enhanced,
 	frame empty,
 	interior empty,
 	colframe=darkspringgreen!50!white,
	coltitle=darkspringgreen!50!black,
	fonttitle=\bfseries,
	colbacktitle=darkspringgreen!15!white,
	borderline={0.5mm}{0mm}{darkspringgreen!15!white},
	borderline={0.5mm}{0mm}{darkspringgreen!50!white,dashed},
	attach boxed title to top center={yshift=-2mm},
	boxed title style={boxrule=0.4pt},
	varwidth boxed title}{theo}
\newtcblisting[auto counter,number within=section]{disexam}{
	skin=bicolor,
	colback=white!30!beaublue,
	colbacklower=white,
	colframe=black,
	before skip=\medskipamount,
	after skip=\medskipamount,
	fontlower=\footnotesize,
	listing options={style=tcblatex,texcsstyle=*\color{red!70!black}},}
\begin{document}
\begin{titlepage}
	\centering % Center everything on the title page
	\scshape % Use small caps for all text on the title page
	\vspace*{1.5\baselineskip} % White space at the top of the page
% ===================
%	Title Section 	
% ===================

	\rule{13cm}{1.6pt}\vspace*{-\baselineskip}\vspace*{2pt} % Thick horizontal rule
	\rule{13cm}{0.4pt} % Thin horizontal rule
	
		\vspace{0.75\baselineskip} % Whitespace above the title
% ========== Title ===============	
	{	\Huge Design and implementation of a soccer ball detection system \\ 
			\vspace{4mm}
		with multiple cameras \\	}
% ======================================
		\vspace{0.75\baselineskip} % Whitespace below the title
	\rule{13cm}{0.4pt}\vspace*{-\baselineskip}\vspace{3.2pt} % Thin horizontal rule
	\rule{13cm}{1.6pt} % Thick horizontal rule
	
		\vspace{1.75\baselineskip} % Whitespace after the title block
% =================
%	Information	
% =================
	{\large Lei Li, Tianfang Zhang, Zhongfeng Kang, Wenhan Zhang\\
		\vspace*{3.0\baselineskip} }
	
 li-lei@ustc.edu, sparkcarleton@gmail.com, kangzhf@gmail.com, wenhanzhang430@gmail.com \\
	\vfill
% If you come across any problems, see section \ref{sec:resources} for possible\\ \vspace{1mm}
% solutions or contact me at \url{armindubert19@gmail.com}\\ \vspace{1mm}
% Happy \LaTeX-ing!
\end{titlepage}
%%%%%%%%%%%%%%%%%%%%%%%%%%%%%%%%%%%%%%%%%%%%%%%%%%%%%%%%%%%

\begin{abstract}
    The detection of small and medium-sized objects in three dimensions has always
been a frontier exploration problem. This technology has a very wide application in
sports analysis, games, virtual reality, human animation and other fields. The
traditional three-dimensional small target detection technology has the disadvantages
of high cost, low precision and inconvenience, so it is difficult to apply in practice.
With the development of machine learning and deep learning, the technology of
computer vision algorithms is becoming more mature. Creating an immersive media
experience is considered to be a very important research work in sports.

The main work is to explore and solve the problem of football detection under
the 36 cameras, aiming at the research and implementation of the live broadcast
system of football matches. Using multi-cameras detects a target ball that is about
2 * 10$^-5$ of the original image and determines its position in three dimension. This
problem is relatively advanced in the AI live broadcast system. It’s full of challenges
due to the occlusion, motion, uneven illumination of the target object.

This paper designed and implemented football detection system under multiple
cameras for the detection and capture of targets in real-time matches. The main work
mainly consists of three parts. First, design and implement the football detector, using
the deep convolutional neural network framework to design the detector, which
compress and optimize the model to maintain good accuracy and real-time. Secondly,
design and implement single camera detection, aiming at the target in the monocular
camera in the lighting, motion and occlusion scenarios. Third, design and implement
multi-cameras detection, for a large number of video stream information, using an
adaptive video detection algorithm framework.the system used bundle adjustment to
obtain the three-dimensional position of the target, and the GPU to accelerates data
pre-processing and achieve accurate real-time capture of the target. By testing the
system, it shows that the system can accurately detect and capture the moving targets
in 3D.

In addition, the solution in this paper is reusable for large-scale competitions,like
basketball and soccer. The system framework can be well transplanted into other
similar engineering project systems. It has been put into the market.

\textbf{Key Workds : 3D, multi-cameras, Computer vision, Detection} 
\end{abstract}

\vfill

\tableofcontents
\vfill
% \small{\noindent \textbf{About This File} \vspace{-3mm}\\
% \noindent \rule{3.3cm}{0.5pt} \\
% This file was created for the benefit of all teachers and students wanting to use Latex for tests/exams/lessons/thesis/articles etc.\\
% The entirety of the contents within this file, and folder, are free for public use.}
\newpage
\newgeometry{
	left=29mm, 
	right=29mm, 
	top=20mm, 
	bottom=15mm}
%%%%%%%%%%%%%%%%%%%%%%%%%%%%%%%%%%%%%%%%%%%%%%%%%%%%%%%%%%%
\section{Introduction}
\vspace{10.5cm}
This chapter introduce the background and the development of the study, and illustrate the contribution and organlization of the paper.
% \newpage

	\subsection{Background and significance of the study}
        With the continuous development of information technology and the rise of Internet technology at home and abroad, as well as the improvement of computer computing power and the popularization of smartphones, people are getting more diversified and richer information.
        With the improvement of computer computing power and the popularization of smartphones, people are getting more diversified and richer information.
        The main focus here is on the processing of text, audio, image and video information.
        The main focus here is on the acquisition and processing of text, audio, image and video information. Images and videos are more visual and easier to represent, and they are constantly being developed by image computing technologies.
        Developments in computer technology. Driven by artificial intelligence, computer vision technology has been greatly developed in both academic and industrial fields.
        Especially with the popularity of deep learning, computer vision techniques have received more attention from researchers.
        Especially with the popularity of deep learning, computer vision techniques have been favored and promoted by more researchers.

        Small target detection is one of the most active research topics in the field of computer vision, pattern recognition and image processing.
        One of the most active research topics in the field of computer vision, pattern recognition and image processing centers on the use of computer vision techniques to classify, identify targets in images or videos, and
        The core of this research is the use of computer vision techniques to classify, recognize, and identify and describe the contents of targets in images or videos. In reality, the first step in many tasks that require video and images is to perform detection.
        Detection is the foundation of many computer vision tasks and is at the heart of the work, and today's more cutting-edge
        Today's more cutting-edge autonomous driving, AI image search, and augmented reality all exist for algorithms and tasks that require target detection.
        Meanwhile, live video streaming belongs to the category of intelligent video, which is naturally a very cutting-edge topic as well. In recent years, with deep
        neural networks have been vigorously developed in recent years, and with more than sufficient computing power, many techniques from academia
        can be applied and implemented in the commercial field to improve people's lifestyles, and many times a year
        In recent years, as deep neural networks have been vigorously developed in recent years, with more than enough computing power, many techniques from academia can be applied and implemented in the commercial sector to improve the way people live, and many times a year.

        In recent years, VR technology (Virtual Reality) and AR technology (Augmented Reality) have become more and more sophisticated with the help of AI technology.
        Augmented Reality (AR) has become more and more mature with the help of AI technology.
        Target detection solutions in video and images are expected to be developed and practically applied to products by researchers and entrepreneurs.
        Such as real-time target tracking in competitive sports, human behavior detection, and target object recognition in the security industry.
        Then the problem of small target detection in security field, sports and games VR will have a wide range of
        application scenarios. And in live sports, creating immersive media is what the relevant research departments and companies want
        to use data to create and deliver immersive content, i.e. VR, specifically True VR technology and
        This specifically refers to True VR technology and True view technology. For example, in competitive sports in soccer, once the soccer ball is tracked in real time
        Once the soccer ball is tracked in real time, the team can analyze the game while rendering a view of the relevant footage so that the viewer can see the soccer ball from multiple angles.
        This way, viewers can see more exciting moments from multiple angles.

        Motion target detection is the foundation for tasks such as target tracking, traffic monitoring, and behavior analysis, as well as
        is the basis for immersive experiences in sporting events. The immersive experience is created by computer technology to synthesize or
        The computer is then used to superimpose the virtual scenery on the scene, which is finally viewed through the screen.
        However, because the extraction of motion targets is susceptible to background, light changes, shadows, motion speed and other factors
        influence and cause failure, so how to better implement motion small target detection is of considerable importance. 
        
        % As shown in Fig \ref{fig:applicationView}
    % \begin{figure}
    %     \centering
    %     \includegraphics[width=1.0\textwidth]{Fig/fig1-applicationView.png}
    %     \caption{Selected applications in the field of target detection and tracking for image vision.}
    %     \label{fig:applicationView}
    % \end{figure}
        The practical context of this topic is in the field of sports, where multiple cameras are installed around the audience to define the lens
        3D space within the spectator stand to achieve immersive media (immersive media) and True view (real viewing), i.e.
        experience the action in the field of play, i.e., the ball is captured in real time, using all the cameras in the field, to achieve immersive viewing and view
        The camera is used for immersive viewing and angle switching. The tracking of the motion target in 3D space is to get the approximate position of the motion target in the real-time video image, and put the different cameras in different locations.
        The tracking of motion targets in 3D space is to get the approximate position of the motion target in the real-time video image, and to correspond the same target between different frames without using cameras.

	\subsection{Problems}
        As the image or video acquisition process is susceptible to environmental influences, such as lighting changes, local occlusion, target scale changes, etc., making the target appear a certain deformation, for detection will increase the difficulty and infeasibility. At the same time, the same target due to the distance of the camera or different scenes back to bring great variability, the same and different classes of various deformations and scenes of inconsistency often bring great challenges to the detection and identification of moving targets. The second is the impact of the size of the target on the detection accuracy, the smaller the target, the greater the impact on the overall detection effect, because small targets are susceptible to environmental noise, while the characteristics of small targets are not very obvious, so there will always be some methods on small target detection to solve related problems.

        The last is the speed and accuracy of the detection problem, the detection results are generally used in order to give tracking. In particular, the rise of intelligent video for a large amount of video image information, the use of the information how to carry out intelligent analysis, that is, simple classification, tracking and identification and complex judgment of the relevant events, these are built on the detection based on these aspects, while considering these aspects, the performance of the system is also of paramount importance, i.e., the accuracy and speed must reach the degree of feasibility, so that the research and analysis is groundbreaking, so due to the above-mentioned various problems, it also makes the field has been a continuous breakthrough and development.
        
	\subsection{Development of visual target detection tracking}
        Earlier, people have carried out research on motion target detection and tracking abroad as early as the 1960s Papageorgiou et al [2] proposed target detection in static images, which is of course a relevant basis for deep learning interpretation nowadays, to analyze images directly without any prior knowledge, models or motion segmentation. feature representation and used cascade classifier to achieve robust real-time detection of targets.Lowe [4] proposed scale invariant feature transform (SIFT) by obtaining gradient information near key points of an image to describe motion targets.Dalal et al [5] proposed histogram of gradients feature ( Felzenszwalb et al [6] combined HOG with Support vector machine (SVM) and proposed Deformable part model (DPM) [7] to solve the pedestrian detection problem in static images. part model (DPM) [7], which gradually became one of the most popular target detection models in recent years. This work was awarded as the precursor of the model in 2010, and also made the general direction for the later work, which is the basis for the later related work.

        In 2012, Krizhevsky's alexnet network [8] proposed at the Imagenet image competition won the championship of image recognition task and opened the prelude of deep learning in computer vision detection and recognition, and various networks and frameworks led by deep learning have been shaped into open source and landed products. Based on deep learning framework open source one after another, it brings more possibilities for the collision and convergence of ideas. However, based on deep learning with powerful
        The deep learning framework has a strong ability of self-learning, which can deeply explore the potential relationship between the implicit data. Among them, feature representation methods based on convolutional neural networks are particularly effective and have achieved remarkable detection results in recent years. At the same time, there are some issues that have not yet been decided, such as how to determine the number of layers of deep learning and the number of nodes in the hidden layer, and how to evaluate the advantages and disadvantages of the features learned by deep learning. Therefore, further research on deep learning-based
        based feature representation methods may yield groundbreaking results that will eventually contribute to the development of the field. Among the related researches, 3D vision-related papers and researches account for a large proportion, and 3D-based point clouds and 3D detection and reconstruction have been presented in the conference. In the conference, there is a greater breakthrough in 3D detection and reconstruction, not only in theory, but also in application, and later in the establishment of 3D cameras.
        It is believed that 3D vision will be developed very fast with the establishment of 3D cameras.

        Among them, target detection can already be used in security, industry, car-assisted driving and many other fields, such as the widely concerned about the field of car-assisted driving, target detection can achieve accurate detection of people around the body, vehicles, road signs, etc. information, real-time alarm, etc., and has achieved many important results. But at the same time, there are also many challenges There are many challenges. For example, the appearance of different forms of characteristics, complex and diverse background environment, dynamic changes between the pedestrian and the camera The scene changes, the system real-time and stability of strict requirements. In the field of security detection, it is possible to achieve security In the field of security detection, it is possible to achieve the target detection of security facilities, or assisted fatigue driving of cars, etc.

        Target detection and recognition is a very important research direction in computer vision A very important research direction in the field of computer vision, the development of the Internet and artificial intelligence technology, the existence of a large number of images and video data in human life The development of the Internet and artificial intelligence technology, the existence of a large number of images and video data in human life, which makes computer vision technology plays a greater and greater role in human life
        This makes computer vision technology play a greater role in human life, and research on computer vision is getting hotter and hotter. Target detection and recognition, as the cornerstone of computer vision The cornerstone of the computer vision field is also receiving more and more attention. Applications in real life are becoming more and more widespread, such as autonomous driving As a cornerstone of computer vision, research on target detection and recognition is also receiving more and more attention.

        Various information about the target, such as position, trajectory, and pose, is the basis for detection and tracking, as well as object-specific features. These extracted features are also precursors and foundations for the entire system to understand the object of interest. The extracted features are also precursors and foundations for the whole system. At the same time, dynamic and static scenes correspond to different modeling processes, for example, in static scenes, a static background modeling approach can be used. For example, in static scenes, static background modeling can basically meet the overall project requirements, while in some dynamic scenes, more dynamic background modeling is obviously needed. This obviously requires more dynamic background modeling to take into account the change of scenes and other situations [12], and the change of scenes undoubtedly brings more problems to the whole detection, one is how to filter the impact of new scenes, and the other is more than the loss of time and accuracy brought by filtering, and whether all scenes can be fully covered when understanding the definition of the product. One is to raise the standard of data rationality, and one is to challenge the robustness of the algorithm.

        Both image-based target detection and video-based target detection
        have developed rapidly recently, but the tracking and detection of
        motion targets often takes many factors into account, such as accuracy,
        speed, robustness, etc. Especially in doing work related to video
        understanding, manual a priori knowledge is very labor-intensive and
        material intensive, and learning from samples to get the feature
        information needed in deep learning is now the basis of all effective
        The basis of algorithms, supervised learning accounts for more weight
        in landing products, accordingly, whether unsupervised learning can be
        developed to the point where it can play a role in landing products is
        yet to be more developed, while more accurate feature representation is
        also the basis of the whole target detection and tracking project.

        Multi-camera based video analytics is now still based on the image
        aspect of the research, mainly or for the image for the target position,
        pose and other simple motion information for detection and recognition,
        the whole recognition process in the industry is mostly classification
        of the proceeding. Today's video analysis in a complex environment, how
        to describe the scene and what is done in the video is currently a more
        advanced aspect. And for what the target object is doing in three
        dimensions is naturally a very challenge problem. 

        The traditional machine learning algorithm can rely on the geometric
        features in the three-dimensional image, mathematical analysis and
        other operations can be effectively achieved, while the improvement of
        computing power for three-dimensional analysis is also a possibility of
        existence, but nowadays three-dimensional detection and tracking for the
        camera hardware to obtain image information is also an objective
        requirement, so the three-dimensional analysis if it can Therefore, if
        the three-dimensional analysis can be developed more, the hardware can
        be developed by leaps and bounds is the most important thing.

        For the analysis and modeling of motion targets of various complex
        scenes, there are some limitations in the existing advanced algorithms,
        and today's detection and tracking algorithms are still only for the
        analysis and modeling of specific scenes. At the same time, on the basis
        of ensuring real-time, the single-minded increase in computing power
        for the implementation of the entire product is not very reusable, then
        naturally for all kinds of complex scenes, the detection and tracking of
        motion targets must be in real-time and accuracy can be more improved,
        is the need to work on the algorithm and system level to improve, here the
        author according to their own understanding of the target detection and
        tracking Development trends for analysis.

        \begin{itemize}
            \item \textbf{The integration of scenes and targets.}
                Detection algorithms are endless, but algorithms for all kinds of sceneswill be lacking in terms of accuracy and real time regardless, then algorithms for specific target detection must be specified for all kinds of scenes. In this way, for the information in the scene to learn a priori knowledge, then for the entire should be extracted features can be extracted, will significantly reduce the interference of redundant information in complex scenes, and if the a priori knowledge is particularly pure, may cause poorer robustness. In short, this then will contribute to the
                overall robustness and carry out the fusion of scene information, which can improve the overall algorithm performance.
            \item \textbf{Multi-dimensional and multi-level information fusion.}
            The fusion of multidimensional and multilayered information is, to some extent, a key research focus of researchers in major research institutes nowadays. [13] Multidimensionality mainly focuses on the extraction and application of information in the time domain, frequency domain and space domain, so that all kinds of relevant and redundant information can be relevantly used in deep learning. The fusion of multidimensional and multi-layer information requires consideration of several aspects.
            \item \textbf{Hardware design and software design improvement.}
            As hardware facilities in acquiring images of target objects
            nowadays can only reach structured light or depth maps, where for such
            images, the limitations of imaging technology lead to limitations at
            the algorithm level. The effect of a single brush-up algorithm in the
            relevant data set is actually not effective in landing field products.
            Then for the feature map needed for the algorithm, if the imaging
            hardware design can be achieved, such as obtaining the depth point
            cloud features of the target object, then for the analysis and the role
            of such images will be very good to improve. So when the hardware is
            greatly improved, the software design aspect will be greatly improved.
        \end{itemize}
	\subsection{Discussion}
        Artificial intelligence is under the national policy, major research
        departments and enterprises are vigorously promoting the development of
        artificial intelligence in the industry. The technical aspects involved in
        artificial intelligence are mainly deep learning / machine learning, natural
        language processing, computer vision / image recognition, etc. And in the
        global landing technology, computer vision has been in the security,
        auto autopilot, education, AR / VR and other aspects of the
        breakthrough and development, AI / AR / VR three revolutionary
        technologies must be based on AI technology.

        Artificial intelligence technology is the study of making
        computers to simulate certain human thought processes and
        intelligent behavior (such as learning, reasoning, thinking, planning,
        etc.), mainly including the principles of computers to achieve
        intelligence, the manufacture of computers similar to the intelligence
        of the human brain, so that computers can achieve higher-level
        applications

        The core value of VR technology is to bring the audiovisual
        experience to a new level, bringing people a more realistic feeling.
        AR technology, augmented reality, is a technology that calculates
        the position and angle of the camera image in real time and adds the
        corresponding image, with the goal of applying the virtual world to the
        real world on the screen and interacting with it. There are two main
        types of approaches. The first one, synthesis by computer technology. By fusing the actual scene and the computer scene, and then modeling the
        reconstruction analysis for the actual scene.
        The second type, holographic projection. It is the projection of
        virtual scenes directly into reality through a projection device.
        The main research of this topic is the computer vision in the AI
        live broadcast of sports, for VR products, and for the VR product.
        AR products, etc. have references, so the selected topics are highly
        advanced and practical.
        
	\subsection{Contribution}
        \textbf{First, design and implementation of the soccer detector module.}
            \begin{itemize}
                \item Design of data pre-processing algorithms.
                \item Research on deep learning related algorithms.
                \item Design of a training image set annotation scheme.
                \item Identify the design of discriminative targets.
                \item The whole training and implementation of network model optimization.
            \end{itemize}
        \textbf{Second, the design and implementation of the single camera detection module.}
            \begin{itemize}
                \item For the data obtained from a single camera, the purpose is to
                obtain the position of the live capture ball in each frame, so as to
                determine and define the relevant events.
                \item For the study of image filtering algorithms, the purpose is to
                speed up the subsequent detection of the whole, the filtering
                algorithm is good or bad basically determines the loss of time of the whole project.
                \item The research for the detection after the target object is
                occluded and defocused mainly adds the detection of multi-scale
                templates and the model optimization of context (contextual information).
                \item Investigate and test the reusability and implementability of
                relevant algorithms for soccer games, basketball games and rugby
                games. The main focus is to test the speed and accuracy of the
                various modules in the overall project system and until the actual
                product level requirements are met.
            \end{itemize}   
        \textbf{Third, the design and implementation of multi-camera detection
        module.}
        \begin{itemize}
            \item The 2D positions of multiple cameras are used to reconstruct
            the exact position of the target in 3-dimensional space using beam
            flow difference method for multiple cameras.
            \item Verify single camera detection results by 3D position for more
            stable and correct system output.
            \item To increase the detection accuracy of the target by using the
            collaboration information between cameras for the high AP value that
            is not obtained due to the occlusion and vignetting of single camera.
            \item Due to the large amount of multi-camera for data access, image
            compression processing is required to achieve real-time accurate
            target detection, model compression and optimization of the research
            and implementation.
        \end{itemize}
        \subsection{Organization of this paper}
        This paper presents a detailed analysis of the problems of the
        system for multi-camera soccer ball detection, and proposes a series of
        solutions for the overall implementation. In combination with reference
        to domestic and international research work, a soccer ball detection
        system with multiple cameras is designed and implemented, which mainly
        does live soccer AI broadcasting, i.e., capturing soccer balls in real
        time during live matches. The organization and related arrangements of
        this paper are shown below.		

        Chapter 1: Introductory section. Firstly, the background of the
        research involved in the topic and the main work of this paper are introduced to give a relevant overview.

        Chapter 2: Introduction to the relevant principle techniques. This
        chapter describes the relevant technologies involved in the system,
        starting with an introduction to convolutional neural networks,
        followed by a brief overview of the principles of target detection in
        images, and finally a brief introduction to the principles of 3D
        vision.

        Chapter 3: Requirements analysis of the soccer ball detection
        system with multiple cameras. This chapter mainly analyzes the
        requirements of the soccer detection system with multiple cameras,
        which includes business requirements analysis, functional requirements
        analysis and non-functional requirements analysis.

        Chapter 4: Outline design of the soccer ball detection system with
        multiple cameras. The outline design mainly introduces the general
        processing flow, and then describes the design of each module one by
        one, i.e., the design of data pre-processing module, the design of
        detector module, the design of single camera target detection module,
        the design of multi-camera detection and its 3D optimization module,
        and the design of visualization module.

        Chapter 5: Detailed design of the soccer ball detection system with
        multiple cameras. This chapter is the detailed design of the system,
        which mainly focuses on the detailed development and design of the
        outline design from the previous chapter. For data pre-processing,
        target detector, the design and implementation of single camera detection and the framework of multi-camera based video detection system are elaborated and analyzed, and the related effect pictures are shown and analyzed and introduced.

        Chapter 6: Testing of the soccer detection system with multiple
        cameras. The chapter starts by presenting the configuration and the
        environment in which the soccer detection system with multiple cameras
        will be used, while unfolding the environmental parameters to be
        deployed by the system for the subsequent tests. Subsequently, this
        chapter tests the accuracy and speed of the detector, then tests the
        operation of the single-camera detection module and the multi-camera
        detection module, and makes the testing of the whole system fully
        addressed by showing the final related screenshots and test tables.

        Chapter 7: Summary and outlook. A brief summary, along with a
        descriptive analysis of the shortcomings, and an outlook on the AI live
        sports system in the context of future developments.

	\vspace{-1.5mm}
\newpage
%%%%%%%%%%%%%%%%%%%%%%%%%%%%%%%%%%%%%%%%%%%%%%%%%%%%%%%%%%%
%%%%%%%%%%%%%%%%%%%%%%%%%%%%%%%%%%%%%%%%%%%%%%%%%%%%%%%%%%%

\section{Related Work}
\vspace{10.5cm}
	This chapter introduces the main techniques and theories involved
in the design and implementation of a soccer detection system with
multiple cameras, and briefly describes the basic features of these
techniques.

% \newpage

    \subsection{Fundamentals of Convolutional Neural Networks for Detection}

    Before neural networks, if a large amount of feature engineering is
    required when performing detection and recognition, and the work of
    extracting relevant features is a relatively time-consuming and laborintensive task, so a lot of practical engineering cannot be landed and
    implemented. The convolutional neural network provides a certain degree
    of convenience. Theoretically, it is possible to define the relevant
    features and the number of features before inputting the neural
    network, and after determining the framework of the network, the
    relevant data is input into the network for feature extraction and
    classification, and then training, and the most important feature of
    the convolutional neural network is the local perceptual field and
    weight sharing[11] . The efficiency of neural networks has also made them
    widely used throughout the industry and achieved better results.

    \begin{figure}
        \centering
        \includegraphics[width=0.6\textwidth]{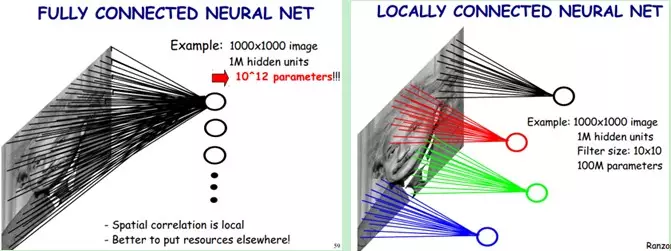}
        \caption{Convolutional neural network extracted features }
        \label{fig:2.1}
    \end{figure}
    As can be seen in Figure \ref{fig:2.1}, the neurons do not need to perceive the
    whole image, but only need to be integrated at the end, which is one of
    the aspects of neural networks that can be computationally efficient,
    especially in target detection is more obvious. After the local perception,
    the parameters obtained are still many, then the weight sharing can be more
    effective is the features are more obvious, i.e. one part of the image has
    the same characteristics on another part and can be considered as the
    same feature. By these two features, the model complexity can be greatly
    simplified and the parameters of the model reduced. In which, the whole
    neural network includes data layer, convolutional layer[12] etc. The data
    is obtained through the correlation layer to obtain the final result[13]
    ,and the results obtained through each layer are analyzed to get the
    final result such as the classification effect needed by the system.
    \begin{figure}
        \centering
        \includegraphics[width=0.6\textwidth]{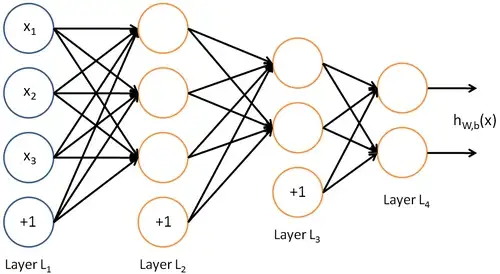}
        \caption{Convolutional neural network diagram.}
        \label{fig:2.2}
    \end{figure}
    The schematic diagram of a convolutional neural network shown in
    Figure \ref{fig:2.2} is the structure of the basic neural network. Convolutional
    Neural Network (CNN[13]) , is a feed-forward neural network in which artificial neurons can respond to surrounding units and can perform large image processing, Convolutional Neural Network includes convolutional layer and pooling[14].
    
    \subsection{Deep neural network-based target detection}
        \subsubsection{Overview of deep learning-based target detection algorithms}
        The broad framework of the target detection task is following:
        \begin{itemize}
            \item   The first step is to select the relevant candidate area, i.e.,
                    the area where the target object may be present. It should be noted
                    that this paper has some innovative points in selecting the relevant
                    candidate areas, which greatly improves the feasibility of the product
                    by filtering algorithm and judging the candidate areas.
            \item   Then for the classification of candidate regions, here the
                    classification, often need to map the score of each region to the
                    corresponding interval, which is now the basis of many research work to
                    design the design of the loss function, the design of the loss function
                    often determines the accuracy and feasibility of the entire algorithm.
            \item   Finally, the possibility judgment is made by classifying or
                    fitting and other parameter information to obtain the relevant
                    location information, and the results obtained by fitting are used to judge the similarity between regions and other characteristics, and to
                    adjust to obtain the most accurate target location.
        \end{itemize}       
        Based on deep learning at present can be divided into three
        categories, mainly is the use of deep neural network models, the use of
        CNN or RNN and other related networks to extract features, and then
        clustering of features and other processing, to obtain relatively
        intelligent results, the following three algorithmic models used by
        major research institutions now.

        \begin{itemize}
            \item   Based on the region proposal of[15 ], i.e., the extraction of
                    features in the target, the entire detection is directly obtained
                    by Results. Such as R-CNN, Fast-R-CNN, Faster-R-CNN [16].
            \item   Regression-based target detection and identification algorithms
                    such as YOLO[17 ], SSD[18 ]. It should be noted that the detection
                    algorithms covered in this paper are designed for self-contained use
                    based on this class.
            \item   Search-based, i.e., further learning is performed for certain
                    features by which the target is judged, e.g., visual attention-based
                    AttentionNet[19 ], reinforcement learning-based algorithms[20 ].
        \end{itemize}
        We have tested the main algorithms in the development of this system,
        and finally selected the YOLO algorithm model that fits well with this
        project as the basic detection framework for optimization and
        development.

        \subsubsection{YOLO Algorithm Principle}

        The system is designed based on the structure of the YOLOv3
        network, which is a fully convolutional network FCN [21] and can be
        suitable for different size images as input. The basic network
        framework design of YOLOv3 is shown in Figure 2.3. The network implements
        end-to-end target detection, so the detection is normalized to the
        cluster fitting problem and the process of target detection is unified
        into a single neural network.

        Nowadays, the authors of YOLO network design from YOLOv1 to
        YOLOv5 and then to the recently proposed YOLOv5 model of each network[22],
        constantly improve the speed of detection, as can be seen
        from the three generations of YOLO changes, for the target detection
        strategy, mainly from five points to make relevant improvements to the strategy,
        so as to adapt to the requirements of the whole system, one is to set
        the a priori box, that is The first is to set the a priori frame, i.e.,
        the region of interest, and extract the region of interest through
        simple feature correspondence; the second is to use convolution for
        scoring prediction to see if the object is contained; the third is to
        use residual networks, i.e., to fuse information through multiple
        layers to obtain more accurate features and more robustness; the fourth
        is to perform multi-scale prediction, which is especially obvious in
        detection and classification problems, corresponding to the size of the
        target in the acquired image will be adapted; the fifth is to compress
        from the model perspective. Fifth is compression from the model
        perspective, by removing redundant layers, or by performing relevant
        pruning operations to make the detected targets more accurate and
        reliable.

    \subsection{Basic principles of multi-camera target detection}
        \subsubsection{Mapping principle of 2D and 3D space}
\begin{figure}
            \centering
            \includegraphics[width=1.0\textwidth]{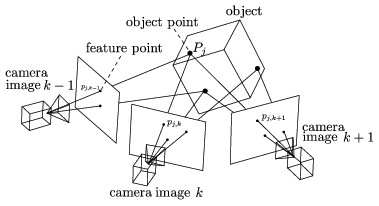}
            \caption{Multi-camera 3D spatial target detection schematic}
            \label{fig:2.4}
        \end{figure}

        Nowadays, there are two main types of methods to basically
        determine the target in 3D space, one is similar to the mathematical
        model of visual slam to obtain the target information in 3D space, i.e.,
        it can be considered as multiple orientation of the camera, according
        to the observation point and the inherent parameters of the camera, the
        3D structure of the scene can be calculated from the target obtained
        by the camera, using Bundle Adjustment (beam flow difference method) The
        principle of the algorithm is shown in Figure \ref{fig:2.4}. Second, based on the
        depth camera to obtain depth (depth) photos, generate the relevant point
        cloud information to obtain three-dimensional information[23].

        Again, the results of multi-camera target detection are based on the
        detection results of the single camera from each angle to get the target in
        three-dimensional space, i.e., the multi-camera detection is based on
        the single camera, and then calculated based on the relevant
        calibration parameters of the camera, and the three-dimensional target
        results are reconstructed from the results of the single camera.

        \begin{figure}
            \centering
            \includegraphics[width=0.6\textwidth]{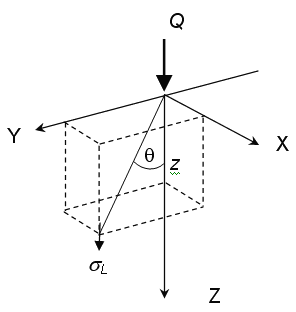}
            \caption{Default coordinate system for 3D images}
            \label{fig:2.5}
        \end{figure}
        
        For images taken under multiple cameras, since the whole system
        is using algorithms and implementations from OpenCV (an open source
        library for images), then it is necessary to model the acquired images,
        which are consistent with the coordinate system of OpenCV. The definition
        of the coordinate system in OpenCV[24] is shown in Figure \ref{fig:2.5}, and the
        coordinate system of the image is the coordinate system of the camera.

        Then, according to the parametric geometric characteristics of the camera model, the coordinates of a point in the 2D image as (u,v) and the position of the target point in the corresponding 3D space as (x,y,z) are interchanged in the form of equations \ref{eq:1} and \ref{eq:2} by obtaining information about the camera parameters, which are very effective for 3D reconstruction and back-calculation of multiple cameras into 2D [24].

        \begin{equation}
            \centering        
            s\left[\begin{array}{l}
            u \\
            v \\
            1
            \end{array}\right]=\left(\begin{array}{ccc}
            f_x & 0 & c_x \\
            0 & f_y & c_y \\
            0 & 0 & 1
            \end{array}\right)\left[\begin{array}{l}
            x \\
            y \\
            z
            \end{array}\right]
            \label{eq:1}
        \end{equation}

        \begin{equation}
        \left\{\begin{array}{l}
        x=\frac{\left(u-c_x\right) z}{f_x} \\
        y=\frac{\left(v-c_y\right) z}{f_y} \\
        z=\frac{\operatorname{depth}(u, v)}{s}
        \end{array}\right.
            \label{eq:2}
        \end{equation}
         So by the formula can calculate the position of each camera to the
        three-dimensional, then how to determine the same target point in
        multiple cameras to the three-dimensional position, the need to use
        the beam flow difference method, through the Sparse Bundle Adjustment
        (sparse beam flow difference method) [26] can effectively solve the
        problem of large sites, that is, for larger spatial values, and may be
        particularly large differences, inaccurate The detailed algorithm
        optimization and implementation is shown in the section Multi-Camera
        Detection in Chapter 5 Detailed Design.

    \subsubsection{Multi-camera 3D vision inspection}
\begin{figure}
            \centering
            \includegraphics[width=1.0\textwidth]{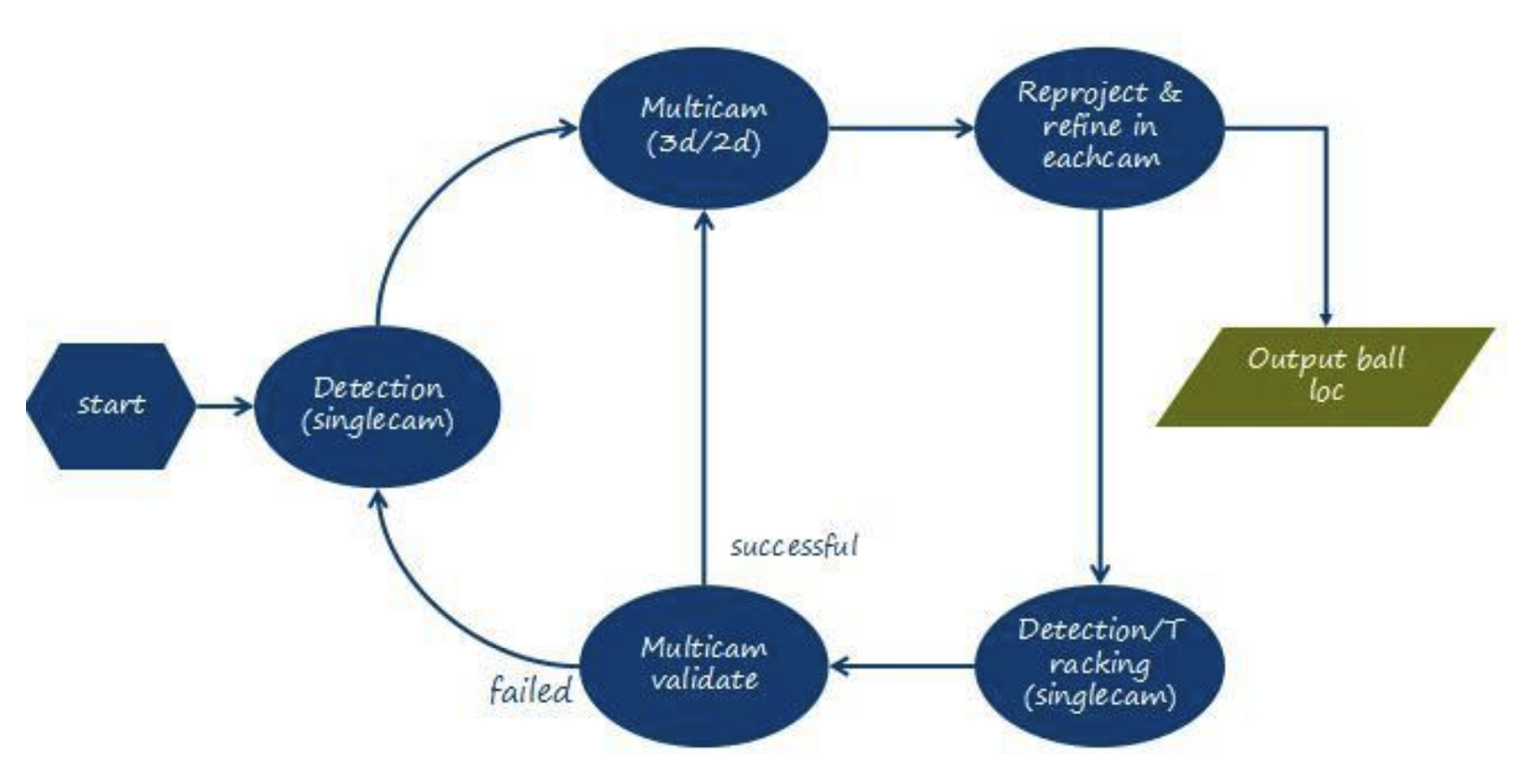}
            \caption{Overview of our pipeline.}
            \label{fig:2.6}
        \end{figure}

        Whether multi-camera 3D vision detection can be effectively
        achieved is based on the single camera, jointly with the introduction
        of the previous subsection for the relevant mapping, for the target
        detection of pictures and videos, the pictures uploaded to the server,
        the background program first find out the ROI (region of interest) in
        the pictures, the pictures are segmented[27] , for multiple ROI regions
        for end-to-end detection, if there is a ball in the picture, the The
        background program will mark the position of the ball and return the
        target coordinate value. The basic network framework of this system is
        shown in Figure \ref{fig:2.6}[28] , and the basic steps of multi-camera detection
        can be clarified from the flow chart of multi-camera target detection
        as follows.
    
        \begin{itemize}
            \item For each frame in each camera into the entire detection system.
            \item Detection is performed on demand, not for all images from all
                    cameras fed to the detector for detection.
            \item Detection is performed for the filtered region of interest and
                    only one target is identified.
            \item Determination of detection-based trackers.
            
        \end{itemize}

        Here are some explanation of terms related to Figure \ref{fig:2.6}.
        \begin{itemize}
            \item \textbf{3d/2d:} From 3D to 2D view.
            \item \textbf{Reproject \& refine in each cam}: Based on the multi-camera results, the
                    single camera with wrong detection or insufficient detection accuracy
                    is reconstructed to get more accurate results for the corresponding
                    single camera.
            \item \textbf{Detection/Tracking(single cam)} : For video streams, tracking can be
                    performed for the previous frame, where tracking is based on the ROI
                    detection model.
            \item \textbf{Multicam validate}: Evaluates the results of the trace to determine
                    if the test results are correct.
            \item \textbf{Output ball loc}: Outputs the location of the target ball for each camera
                    and the position of the target ball in 3D space
            
        \end{itemize}   

    \subsection{Model compression and optimization}
    The dominant approaches in the field of CNN model compression and acceleration can be divided into two categories.
    
    \textbf{Network design.} new network structure, for the entire parameter less,
        such as reducing the computation of larger convolutional layers or
        pruning and other operations, which is a more feasible method in the
        entire data model compression, with representative CNN structure:
        SqueezeNet, MobileNet, ShuffleNet, MobileNetV2, NASNe and so on. There is
        another method called Distillation distillation, which is to train a larger
        model with a larger network structure first, and a larger models tend to acquire more features and are relatively more accurate, but
        they tend to incur a loss in speed, so in using a large model to guide a
        small model, the first allows the small model to perform a fast fit,
        and the second may increase robustness, assisted training has been
        heavily tried and tested in the compression of CNN models[29].

    \textbf{Quantitative Compression.} By directly compressing the
        trained model, e.g., quantifying the weights and/or activations of a CNN
        from 32-bits floating point to a low bit representation[30] ,
        representative methods are Ternary weight networks (TWN), Binary
        Neural Networks (BNN), XNOR-net, and Deep Compression. In compressing the
        model, the author uses Pruning pruning to turn some of the weights into 0 and
        skip the computation of[31] , which achieves better results although it
        may destroy the parallelism of the network to some extent. But this is
        generally for larger models, and such a method Filter/Channel Pruning [32],
        is effective only for large models with large redundancy, and has
        little effect on small models that are naturally compact, and often
        in experiments will try pruning the large model, or the small model.

    \subsection{Summary}
    This chapter provides a brief introduction to the convolutional
    neural network technique, target detection technique and 3D vision
    modeling process used in the system, while it briefly introduces the
    theory of model optimization methods and the general process framework
    for multi-camera detection. With these introductions, the reader is
    paved with the foundation for the subsequent chapters.
    
\newpage

%%%%%%%%%%%%%%%%%%%%%%%%%%%%%%%%%%%%%%%%%%%%%%%%%%%%%%%%%%%
\section{Requirement Analysis of Soccer Detection System with Multiple Cameras}
\vspace{10.5cm}
In the whole iterative development process of the project
system, requirement analysis is a crucial part, as long as the whole
system is clearly defined, the indicators are clearly defined, and
there is an overall basic composition for all aspects of the system
such as functional, non-functional and business requirements. First of
all the system must meet the load under the single camera allows to
achieve the detection accuracy and speed indicators, in order to ensure
that the single camera in the detection has the accuracy allowed in the case
of multi-camera target detection algorithm and the construction of the system.
This chapter first analyzes and discusses the business requirements, and then
unfolds and analyzes the functional and non-functional aspects in various
aspects.
	\subsection{Business requirements for a soccer detection system with multiple cameras} 
		Because in the functional and non-functional requirements are
        built on the author's company's product operations line proposed,
        then in the functional and non-functional requirements analysis before,
        must be an in-depth understanding of the author's company to do the product,
        for what needs target groups, so that after getting a high-level understanding
        of the service provider or user for the entire system product, after the risk
        can be controlled within, such as product Risk, time risk, etc., so that we can
        better grasp the overall research and development process.

        Due to the rise of sports production, then for the entire live sports is
        also a very important and necessary demand, and the use of today's
        effective algorithms for fusion processing, but also a major company
        can seize the market a major factor, VR technology and AR technology
        in the deep learning algorithm spawned by the more mature, about the
        video, image for the target detection of the solution for the majority of researchers and entrepreneurs All hope that the long development of landing
        to the product, such as real-time tracking of targets in competitive
        sports, human behavior detection, security industry in the
        determination of target objects. Which for soccer, basketball, rugby
        and other events is also the majority of people's favorite, especially
        the soccer feast World Cup, once you can do the visual aspects of the
        impact, then for the audience, the experience will be more exciting.

        Creating that immersive media, as advanced business, using data to
        create immersive content, i.e. True VR and True View is especially
        important for businesses and research departments. Once you can
        capture the types of moments that viewers want to watch during a game,
        then that business need is fulfilled.

        In the context of the above requirements, it will certainly be
        poorly implemented with manual cost, and it is difficult to get realtime processing of huge video information through manual. At the same time
        the system can be processed for any sport, especially for soccer,
        basketball, rugby, which are characterized by many participants and
        difficult to detect behavior, can be well and effectively implemented.
        After the multi-camera detection of the target, then the surrounding
        sub-targets, such as player information can also be obtained, then in
        the analysis of player information, for the player's movement
        trajectory and conditions can give relevant data for analysis. At the
        same time, the real-time background processing of the relevant footage
        is also the result of three-dimensional space soccer capture, threedimensional results can also be returned in real time to the need for a
        single camera under the scene and figure, and the 3D reconstruction of the wonderful shots is also based on high level target detection and tracking.
        
	\subsection{Functional requirements}
    \subsubsection{Overall demand}
		When determining the functional requirements of the entire system, it
        is necessary to determine the functions and content that can be improved
        by each relevant module, so that the entire product line can be
        considered and started when the entire system is built. In this way,
        when designing each module, we can have a relevant dependency on the
        inclusion of the relationship.

        This system is composed of a system server-side program and a clientside program. And the main thing is to do is the real-time calculation and
        processing and other operations on the server side, while the client side
        is to output the relevant video frame stream and the results obtained from
        the server side to the client side.

        \begin{figure}
            \centering
            \includegraphics[width=1.0\textwidth]{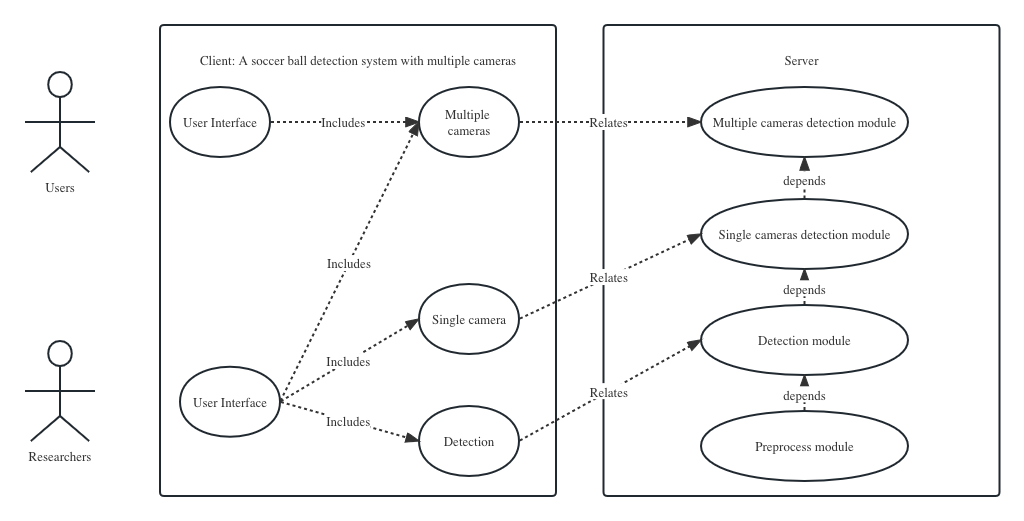}
            \caption{Use case diagram of a soccer detection system with multiple cameras }
            \label{fig:3.1}
        \end{figure}

        Figure \ref{fig:3.1} shows the use case diagram of a multi-camera soccer detection system. From the use case diagram of the multi-camera soccer detection system in Figure \ref{fig:3.1}, it can be seen that the system ultimately faces two types of users, one is the audience, the user group that the final live streaming system will face, and the other is the developer of the player group, because, the whole live streaming system is designed based on the target ball and player development, and the multi-camera detection module can be seen to be based on a single camera at this point.

        The actual application scenarios of this project, such as the La Liga soccer game, Olympic basketball game and NFL football game in which the author's company is involved, this thesis is detailed with soccer as the target, so take this soccer game as an example, a large soccer stadium surrounded by spectator stands and tall closed buildings, so 36 monocular cameras are arranged around to capture the game in the field and the sidelines each camera is Each camera is fixed and can only see part of the stadium and the sidelines. Then, the server-side program receives the video streams from the 36 cameras, preprocesses them, detects the target of each camera, gets the target position of each camera, and then determines the exact position of the target object in 3D space by the related algorithm. Finally, we output its position information in 3D space and the 2D positions of the cameras that can see the target object.

        \subsubsection{Modeling requirements for camera environments}

        The model built by the camera needs to be accurate, because this system has certain requirements for the accuracy of the three-dimensional space established by the camera, the subsequent need to get the location information from two-dimensional to three-dimensional space, then the camera model established needs to be complete location information, that is, the corresponding matrix information, the matrix information utilized can be seen in Chapter 2 related algorithm description, if the camera modeling is not accurate, then for the later multi-camera detection is difficult to be very accurate. At the same time, the location of the camera can not be too far from the target, take this project as an example, the smallest target is about 10pixel * 10pixel, and the size of the image pixel obtained by the camera is 5120pixel * 2760pixel, about $2 * 10^-5$ of the original image than the target ball.
        For the target sphere, if the distance is very far, the framework of the algorithm and the model will be difficult to maintain the accuracy. Therefore, if the system can function properly, it needs a certain light intensity and a well-established three-dimensional frame model of the camera, so that the subsequent acquisition of the image can be processed properly by each subsequent module.

        \subsubsection{Data pre-processing requirements}
        This requirement is based on the load, due to the large amount of image data acquired by 36 cameras, and also the large number of image pixels acquired by each camera, which will lead to a large amount of subsequent calculations, thus making it difficult to obtain real-time results and affecting the response time of the whole system, etc. Then by first compressing the video images acquired by the cameras to an executable range, the original images of each frame are then input to the system. In the subsequent model compression process, not only the corresponding layers are trimmed and model compressed from the algorithm level, but also the data related to the image video compression aspects are pre-processed. After entering the system, the entire image is filtered in order to speed up the corresponding time, so that the post-processed image has more distinct features and makes the image more recognizable. Also, although the system is robust to light, if the light is very dark, i.e., when the camera collects images against the light, the images are darker, and for this situation, transformations such as image brightness will be performed. From the above, it can be seen that the data pre-processing of the image is basically for the whole process, and the good or bad data processing directly affects the effect of the system.

        \subsubsection{Two-dimensional single camera target object detection requirements}
        In this system, the detection and tracking of the soccer is the most important aspect, and all the subsequent work is based on the detection. A good adaptive algorithm is actually derived from several algorithms that effectively improve the accuracy while guaranteeing real-time performance in the associated filtering algorithm. And at the same time the purpose of stand-alone detection is to detect moving targets in the background of different complex angles in the video stream. The detection-based tracking algorithm is the prototype of many current frameworks, so the good or bad target detection will directly affect the performance of advanced tasks such as target tracking, identifying dynamic behavior, and discerning the events of related exciting shots. This requirement is a basic one, but if the target object is obscured in a camera for a long time, here the author considers a large area obscured for a long time, i.e., there is no target object in that camera, then a multi-camera collaborative detection is needed so that the two-dimensional position of the obscured target object in that camera can be calculated with the help of the results from other visible cameras.

\begin{figure}
            \centering
            \includegraphics[width=1.0\textwidth]{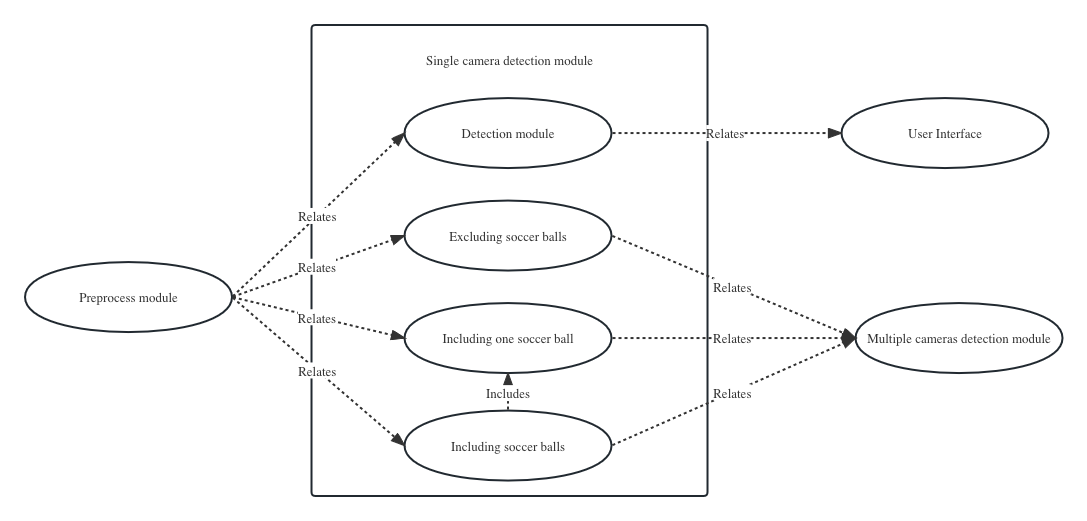}
            \caption{Use case diagram for the detection module of a target object with a 2D single camera}
            \label{fig:3.2}
        \end{figure}

        The target object detection module of a single camera can be used for unit testing of a single camera, and in actual use, the detection of multiple cameras is scored to get the visible sphere range of the relevant cameras, to determine which cameras contain the target object, and if no target object is obtained through that camera, to see if it is a missed detection, and if not, to terminate the detection module of that camera, so that the overall computation will be reduced, and also will reduce the latency. If there is a target in the image of a single camera, the judgment is made to see if it is a false detection, and then the relevant detection and tracking framework is used to make the judgment so that it reaches the target of the product design. Figure \ref{fig:3.2} shows the use case diagram of the target detection module for a 2D single camera.

        \subsubsection{Detection requirements of target objects with multiple cameras}
        Multi-camera target detection is commonly encountered in interactive virtual worlds, live sports and video surveillance, and is similar to the visual slamming method, but does not take into account the delay in camera motion. Two-dimensional estimation is the first step of a single-camera detection module, then a multi-camera detection module is to achieve the estimation of target objects in three-dimensional space, so it is required to determine the area that each camera can determine and thus which cameras have target objects.

        The requirement for multi-camera detection is to obtain a three-dimensional target object through a modeling process with multiple cameras based on single-camera detection, and then determine the two-dimensional location of each camera based on the three dimensions. After obtaining that position, the relevant area is then placed into the detector to obtain more accurate position information. For example, for an object in the air, it may be out of bounds if it is within the area of the camera, but for 3D space, multi-camera detection is required to determine its 3D position. In this system, the cooperative detection of multiple cameras is the most important and the last important module to be tested.

        \subsubsection{The requirement for target detection in video}
        The target detection in video is the last step of the multi-camera soccer detection processed by the authors and is a very important functional requirement. The target detection module of the video is built on top of the detection module of the image, which is based on a trained detector. Then the establishment of an adaptive unified framework to achieve the video multi-camera module of time propagation and cross-scale refinement will need to consider many factors of the whole system, and three-dimensional video detection is a lot of algorithms to do related work, can be based on the interrelationship in the time series to optimize the results of other cameras, while based on the video sequence, the accuracy of the entire video detection can be further by three-dimensional optimization algorithm optimization.

    \subsection{Non-functional requirements}
        The non-functional requirements of the system are the corresponding requirement elements to ensure the performance of the system, robustness of the system, ease of use, scalability requirements, etc. The non-functional requirements for the implementation of the soccer ball under the camera detection system designed in this paper are as follows.

        \subsubsection{Robustness}
        The system should have good robustness. It can operate effectively in multiple court environments, whether it is raining or under uneven lighting conditions, i.e., it should detect in multiple match videos, and the requirements for multiple match videos are different venues, different weather conditions, different light intensities, etc., to obtain good results.

        \subsubsection{Accuracy}
        The live sports system must get the requirement in accuracy before it can be installed and used for live broadcast. For example, in this system, multiple cameras in switching lenses, to have the picture that the audience needs, in multiple screens can be very accurate access, and for the target object in the screen must be high precision to get the location, so that for goals, long passes and other exciting shots can be effectively captured to achieve good results. This is also an important delivery requirement from the author's project team.

        \subsubsection{Real-time}
        The system needs to have real-time, real-time as a necessary indicator to enhance user experience must be considered in the requirements, so the system must be able to ensure that about 40ms per frame in the whole system as a system indicator, from the algorithm detector and computing power of both aspects to get an effective guarantee, need to meet the real-time requirements of the actual environment. Here, we mainly need to reduce the computation time of the detector by compressing the model, and build a multi-card operating environment for 36 cameras to work together. Building a multi-card cloud processing environment is the work of other group members in the authors' company and is not discussed below.

        \subsubsection{Portability}
        The portability here is mainly that the system can be effectively adapted to another system platform by exchanging one platform for another.

        \subsubsection{Reusability}
        The algorithm is reusable. The system is built to operate mainly for the three types of games in the pre-defined requirements, namely soccer games, basketball games and rugby games. The common denominator of these three games is ball sports, therefore, the natural target objects of the system are soccer, basketball and rugby. Therefore, the algorithm design, both in the training phase of the model and in the testing phase of the loaded model, should unify the relevant interfaces to make the algorithm reusable.
        The system framework is reusable. Since there are 36 cameras per site, the overall system framework of the cameras does not need to be changed, only the detectors need to be modified for training purposes. So the system needs to be well reusable.

        \subsubsection{Practicality}
        The project for which this system was designed is an industrial line of immersive media that the author's company is preparing to create. This live sports system not only needs to be viewed live in the cloud, but also needs to be effectively implemented with embedded VR glasses. So with all the server-side programs of the system running in the background, only accurate results and image transfer to mobile devices seemed crucial. Finally, it is necessary to intelligently determine the exciting shots, such as shots, passes, penalty kicks and other events through the detected soccer ball motion trajectory, and then render images for the relevant events. And at the same time, we obtain real-time information about the ball's position, perform relevant tactical analysis on the relevant data, and make relevant comments on the whole game, which are also related to the follow-up work.

        \subsection{Summary}
        This chapter analyzes the requirements of a multi-camera soccer detection system, including business requirements analysis, functional requirements analysis and non-functional requirements analysis.
\newpage
%%%%%%%%%%%%%%%%%%%%%%%%%%%%%%%%%%%%%%%%%%%%%%%%%%%%%%%%%%%
\section{Outline design of a soccer ball detection system with multiple cameras}

\vspace{10.5cm}
With the development of video detection theory, intelligent video analysis has been among the more cutting-edge research topics, and semantic segmentation of detection has been very successful, yet obtaining relevant semantic information based on video detection is still a very challenging task. Specifically, the throughput of video streams has a significant cost for both load and computation, while low latency is required for practical use, so frameworks developed for overall video semantic segmentation are crucial.

This chapter provides a general overview of the algorithmic framework of the designed multi-camera soccer ball detection system. The soccer ball detection system with multiple cameras designed in this paper contains three main parts: two-dimensional single-camera detection, three-dimensional multi-camera detection and video stream-based detection framework. The overall structure, functional and non-functional structure of the system is outlined to provide a good basic direction and ideas for the subsequent chapters. The system is summarized in three areas: first, the detection of targets in multi-camera video; second, the implementation of the soccer algorithm; and third, the implementation of a live streaming system based on video stream detection.

	\subsection{General framework design of the system}
		The design of a multi-camera based soccer ball detection system is to create a visualization of capturing targets for live ball sports. The main purpose of the system is to stream the video captured by multiple cameras to the overall system and then return the captured video to the client. The overall framework is to acquire the match video streams through 36 cameras, transfer the acquired video image load to each server to calculate the detection results, and then transfer the results and related images to the scheduling server, one to the image rendering server in order to present them to the general public through relevant smart devices, and the other to the player detection development server in order to obtain the target structure and thus carry out subsequent related work . The whole core module is the computational process of sub-module after the server receives the video, processes each camera in parallel, and then executes single-camera detection, multi-camera collaboration, multi-camera 3D determination, multi-camera to single-camera optimization, and finally selects the relevant strategy in turn.

\begin{figure}
        \centering
        \includegraphics[width=0.9\textwidth]{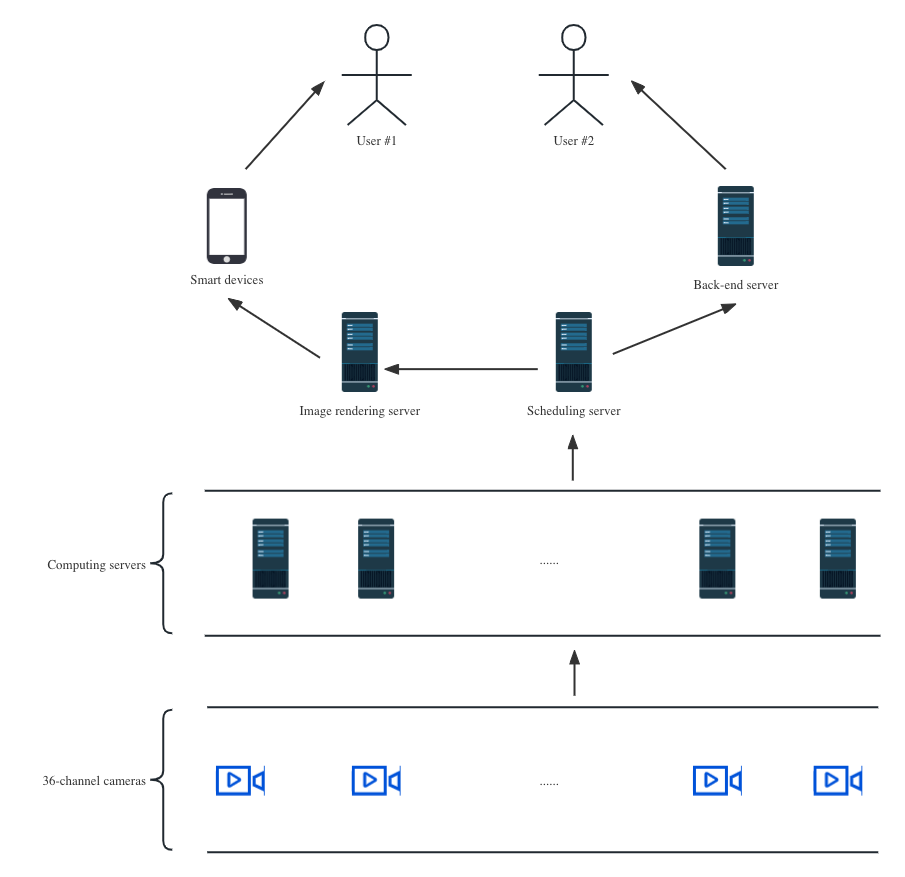}
        \caption{Overall system architecture}
        \label{fig:4.1}
    \end{figure}

      The whole architecture is shown in Figure \ref{fig:4.1}. The main work of this system lies in the design of the server-side backend algorithm, then the precise implementation of video and image processing is the work that needs to be done in the whole architecture. There are two main types of users, one is the viewer user who only needs to be able to get the motion capture of the total target object in the whole game screen directly. The server side executes the video pre-processing module, key frame processing module, target detection module, video key frame capture module and optimized 3D and 2D modules in a parallel order.

    Another class of users is the player detection developer, the overall architecture of the system will eventually output the processed raw image and the target position of the ball, the target position of the ball is the position in 3D space and the 2D position of each camera, for the player detection developer, to detect the player holding the ball according to the player motion detection algorithm, the developer has to define a clear interface and needs to be relevant to the camera 3D modeler requirements to be described. Eventually, the system needs to output two results, one is to transfer the target capture video images in 2D and 3D space to the graphics image rendering server to render the whole motion process and highlight shots, and the other is to output the target capture position information in 2D and 3D space to the server backend for the player detection developer.

    Figure 4.2 shows the module structure of the backend algorithm for soccer video detection with multiple cameras. The whole system is mainly for image and video processing, and the modules are mainly image pre-processing module, target detection module, 3D target detection module, etc.

    \begin{figure}
        \centering
        \includegraphics[width=1.0\textwidth]{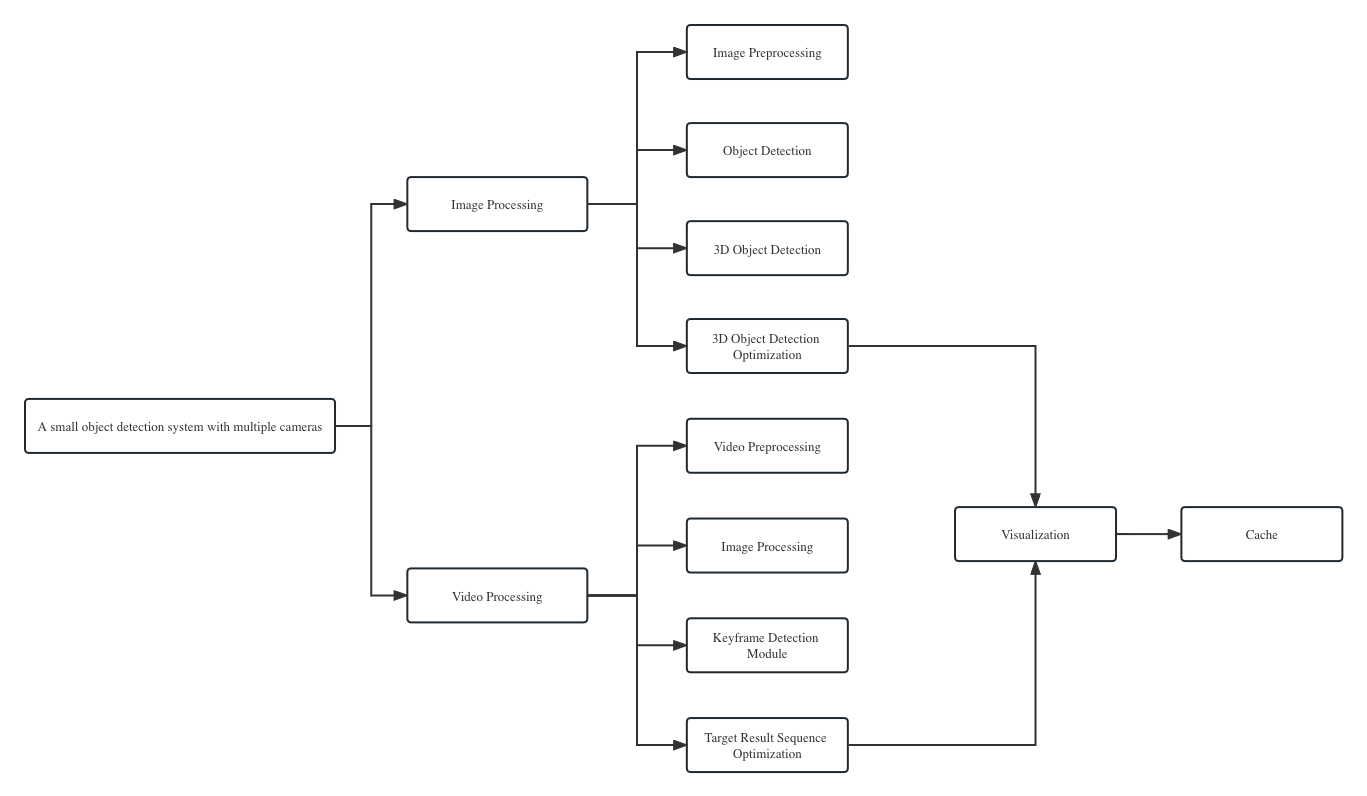}
        \caption{Caption}
        \label{fig:my_label}
    \end{figure}

	\subsection{Module Design}
		\subsubsection{Overall processing flow}
			Before implementing the whole system, designing relevant modules based on the overall system is also a key step in this engineering project, both for better division of labor and for efficiency. Clarifying the schedule and definition of each module is especially important for the subsequent design of a logically clear and functionally complete system. Each module is clear and complete, for if it involves changes in requirements iteration, after the interdependencies between the modules are clearly defined, it is also possible to quickly correspond and complete the update iteration of the algorithm code quickly.

           This system mainly consists of matching the server side of the system with the relevant clients. For the input in the form of pictures, relevant tests can be performed, firstly to verify whether the results are the same across platforms, and secondly for more effective on-site analysis, etc. That is, you can judge whether there is a target according to the picture, if not, then return the null value, and if there is a target, then judge how many targets there are, because there will be related false detections, at this time will be based on the results of other cameras for a comprehensive judgment, and the detector will have a scoring mechanism to make it get accurate results. For the input in video form and then in image form, the authors in this paper have designed an algorithm module for integrating detection, time span and cross-scale refinement so that the results of the relevant sequence are output in the video stream for the rendering of the relevant image. Of particular note in this system is the need to obtain information about the foreground, regardless of whether the input is an image or a video, where the foreground in this system is defined as a blank stadium, i.e. without redundant information such as players and balls.

            \begin{figure}
                \centering
                \includegraphics[width=1.0\textwidth]{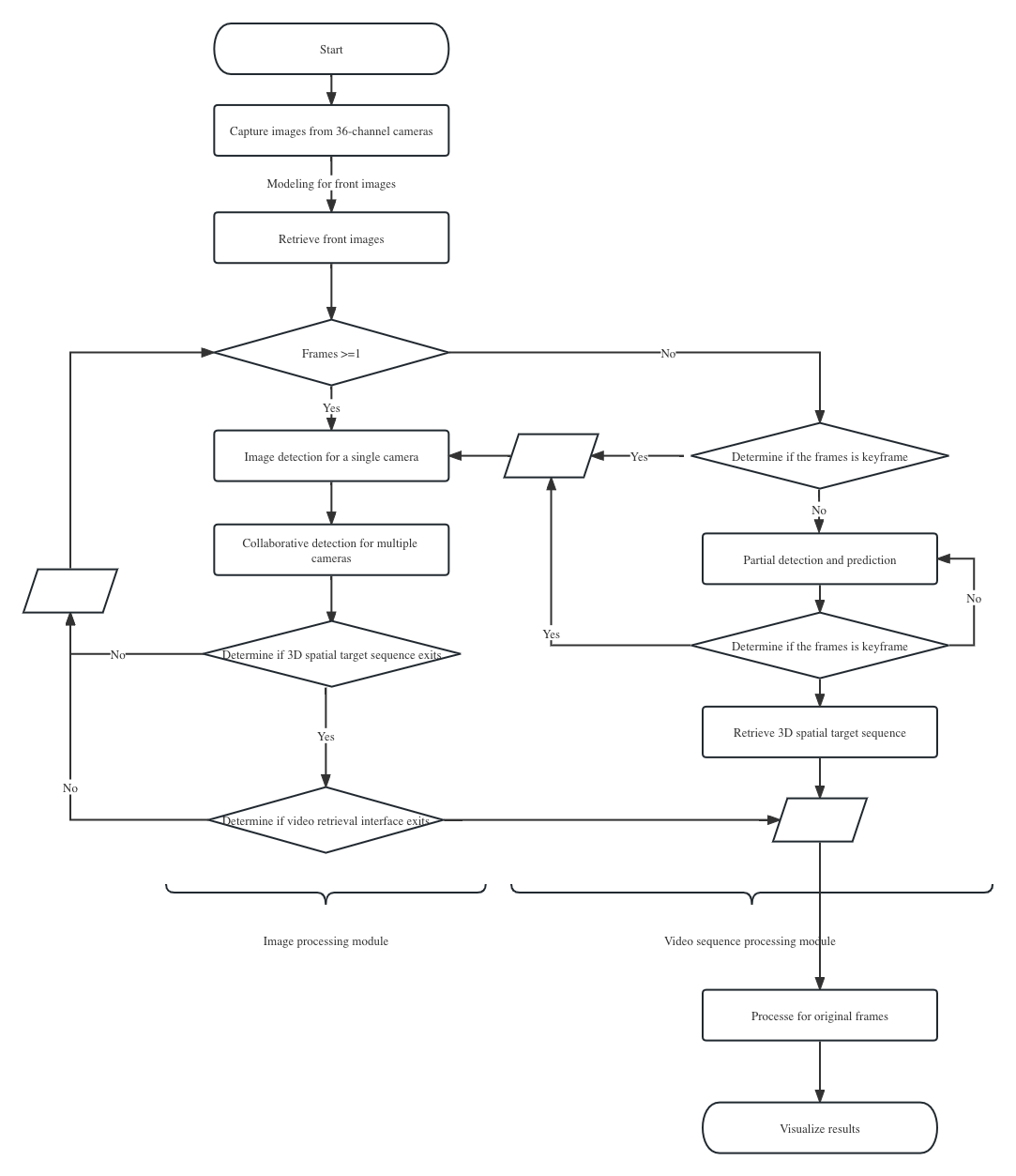}
                \caption{System overall logical module activity relationship diagram}
                \label{fig:4.3}
            \end{figure}

            The overall logical module activity relationship diagram of the system is shown in Figure \ref{fig:4.3}. It can be clearly seen that the server will have a larger load when receiving videos or pictures, especially when receiving videos, the whole video needs to be serialized, so the time series algorithm module has a great improvement for the detection of videos, if it can make the target more natural and accurate in the excess between the 3D space. On the other hand, when there is short-term occlusion in the video, the post-processing algorithm can correct it to some extent. And to visualize it is necessary to call the relevant image compression module to compress the image and then send it to the whole frame for detection, which is not only for the computational power to be reduced while maintaining the accuracy, but also for the speed to be improved, and finally the result is input to the original image and presented to the user. Because of the high accuracy of the video, then a caching mechanism is needed for a period of recording, and finally the backend program visualizes the results into a video image and then renders the rendered image.
           
		\subsubsection{Design of data pre-processing module}
\begin{figure}
            \centering
            \includegraphics[width=1.0\textwidth]{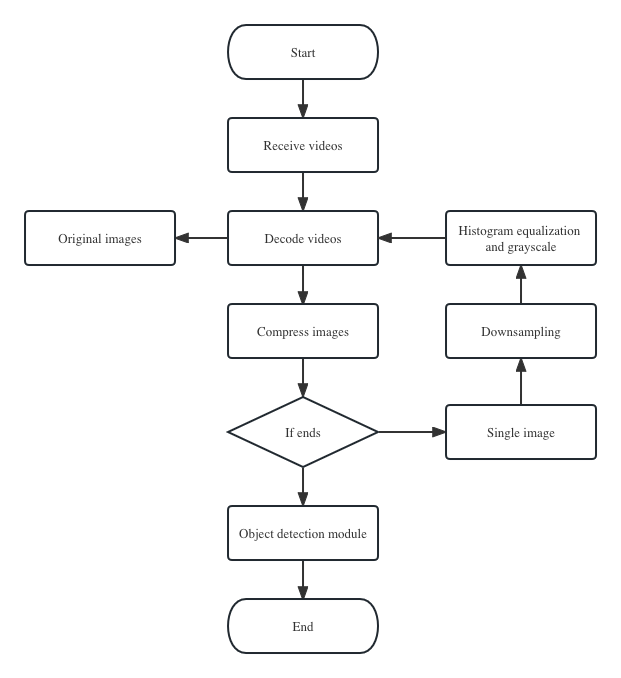}
            \caption{Data pre-processing module activity diagram}
            \label{fig:4.4}
        \end{figure}

		A well-designed data pre-processing module provides the basis for the effective operation of the whole system. As can be seen from Figure \ref{fig:4.4}data pre-processing module, video decoding will store the raw image in the relevant server, which is uncompressed. For the compression of high-definition video images is to reduce more load, because the reduced load can often reduce costs, then for data pre-processing needs to emphasize the video to picture, 36-way camera to obtain the high-definition video stream, the author in the design of the system, often for a single frame image analysis and processing, then the video to how much accuracy of the picture for the entire detection and system construction is an important indicator. Later, when the detection structure is modeled and visualized in series, it will bring a better experience to the user, but the data module is often accompanied by a caching mechanism, the definition of the relevant cache data is determined by the subsequent load and the subsequent demand for product services. This paper does not go into details.

        As the load requirements are relatively high, the subsequent convolutional neural network algorithm requires the input data to have as little noise as possible, so the video sequence from the camera needs to be compressed, and as this is the pre-processing of the image data, the image module is mainly for developers and service providers, and the module design is also carried out here. After decoding and compressing the transmitted video, the images are sorted and for individual images can also be detected and tested. This subsection unifies the interfaces of preprocessing of images and preprocessing of videos with different judgments, so that high response and high accuracy can speed up the calculation of the system and make the data processing of each module faster.

        \subsubsection{Design of the detector module}
        This module is a standalone module that is called frequently throughout the system and is the most important module in the system. This module needs to be constantly optimized and compressed, and the other modules have to be changed accordingly. The detector module is mainly a model trained by a deep neural network, and the images to be detected are sent to this detector to obtain the relevant results and scores. This module is divided into three main tasks, one is the processing of training and test data, which is to have a relatively pure data distribution so that a good model can be trained, the second is to compress the model, for possible speed and accuracy problems in the system, for overall pruning and dimensional compression, and the third is to load the model module, for the whole system, the training model needs to be loaded in, to predict the relevant results. The training and algorithm development process of the detector module will be described in detail in the subsequent sections.

        \subsubsection{Design of single camera target detection module}

        Target detection and capture is the first module after data processing and is one of the core modules of this paper. It mainly uses foreground modeling to first segment the pure data of the whole target field with the corresponding image, hand over the region that may contain balls to the detector for detection, and then return the detection result. This result may have multiple results containing targets or false detections, and a good accuracy needs to be maintained here to ensure that the subsequent detection module can effectively improve the accuracy of false and missed detections to a deliverable level in the case of multiple cameras.

        The stand-alone target detection module is based on a good detector which is designed based on YOLO, divides the image into multiple regions of interest, sends the regions of interest to the detector, gets the predicted values and then processes them further to filter out low confidence and overlapping parsed bounding boxes so that each region of interest of the image corresponds to the parsed bounding box with the highest confidence. The objective is to filter out the low confidence and overlapping parse bounding boxes so that each region of interest of the image corresponds to the parse bounding box with the highest confidence, and set a threshold to eliminate the low scoring regions of interest. The final image may have no target results or one target result and multiple target results, and the corresponding camera numbers and target results need to be stored.

        It is important to note here that before the single camera target detection module is implemented, the court outline of the corresponding camera must be segmented, where the relevant key points of the court need to be obtained manually, and the coordinates of the key points correspond to the camera number and the size of the court, which can calculate the camera's playing area. This also applies to other cameras.

        \begin{figure}
            \centering
            \includegraphics[width=0.4\textwidth]{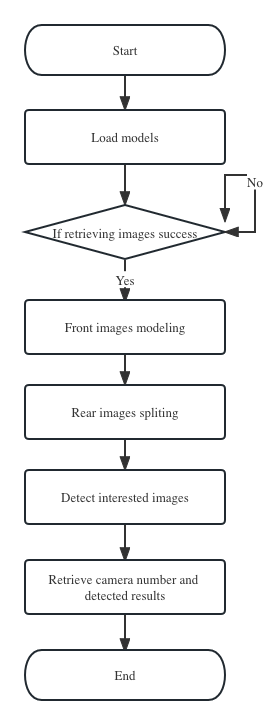}
            \caption{Single camera target detection activity diagram}
            \label{fig:4.5}
        \end{figure}
        Figure \ref{fig:4.5} shows the activity diagram for single camera target detection. The module is executed by first acquiring the image, and after the data preprocessing module, the image sent is a single frame image, and if the image is acquired incorrectly, the image is reloaded, as the course image is available before the match. The relevant foreground image is obtained by a segmentation algorithm based on hybrid Gaussian modeling of foreground and background, and the subsequent detection module, all subtracting the foreground image, so that a simple segmentation of the hindground can be achieved and the relevant threshold for image segmentation is determined. After subtracting the background from the current image to obtain the region that may contain the target sphere, the segmentation of the hind scene is established, and then the region is sent to the detector for determination. Without hindfield segmentation, searching the whole image according to the sliding window is not a small burden on computation and load, etc., and does not meet the requirement of real-time. And the good or bad filtering algorithm will directly affect the whole speed and accuracy, which is now the problem that many algorithms need to solve in end-to-end. At this time, the target area that may appear after the post scene segmentation is sent to the detector for relevant scoring, and finally the number of cameras and the scoring of all boxes are output, and the relevant data are output for multi-camera co-processing.

        \subsection{Detection of multiple cameras}
        The product based on this system is the capture of three-dimensional space target object, and all the current three-dimensional space technology is mainly depth camera to obtain the determination of the three-dimensional point cloud of the target object, and the second is multi-angle for the three-dimensional reconstruction of three-dimensional objects. The system is based on multiple monocular cameras, which can obtain 360 degrees of the target object without dead angle, so it can effectively solve the problem of occlusion in the collaborative detection of multiple cameras. The system is based on multiple monocular cameras to obtain each projection angle, and after obtaining the target value, the 3D position is obtained based on the beam flow difference method, and then the 3D position is back-calculated to the 2D position, and since the detector can be designed to be lightweight, then for the back-calculated position, another detector calculation can be performed to optimize the 2D position. Currently, the convolutional neural network-based algorithms outperform traditional image processing algorithms in terms of engineering implementation and algorithm optimization due to the increase in computational power.

        \begin{figure}
            \centering
            \includegraphics[width=1.0\textwidth]{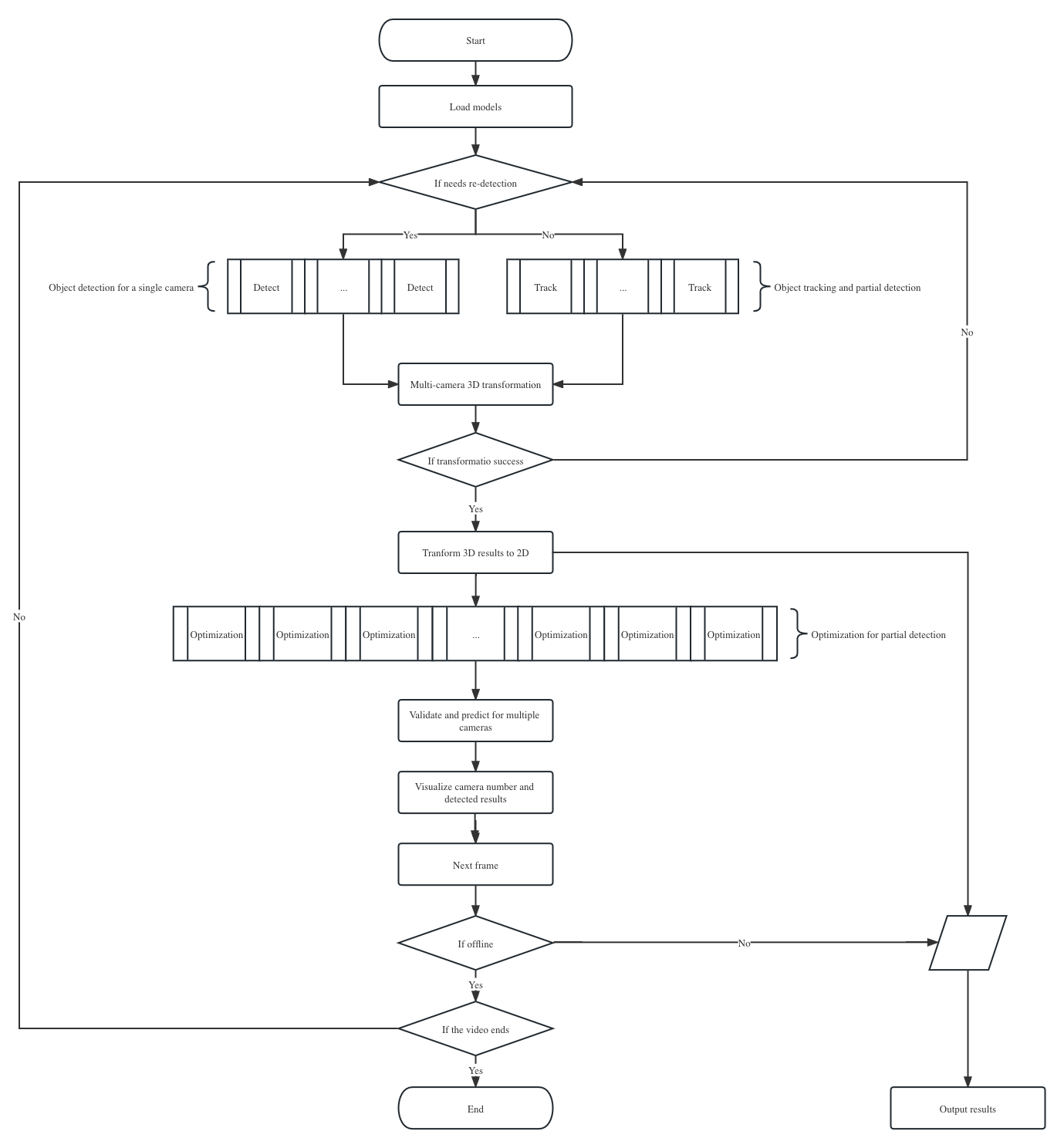}
            \caption{Detection of multiple cameras and its 3D optimization module}
            \label{fig:4.6}
        \end{figure}
        Figure \ref{fig:4.6} shows the activity diagram for the detection of multiple cameras and their 3D optimization module, as well as the activity diagram for the whole framework. As shown in the figure, the basic flow is as follows.

        \begin{itemize}
            \item  Load the model and initialize the relevant parameter settings. The system will execute the module of multi-camera detection after receiving the image, first load the model and initialize the detector, which is also the overall entrance to the system, the detector will be loaded into the program when the system accepts the need to perform work, because the detector is needed at any time during the work. The detector is an independent module in the whole system, because there are mainly two aspects at the moment of calling the module, one is the need to always call the module when detecting in a single machine, by calling the module to obtain the position of the target object under a single machine; the second is in the multi-computer target tracking, the need to call the results of the previous frame on the basis of the current frame attachment, so the module is good or bad determines the effectiveness of the whole system.
            \item  36 cameras were assigned for target detection. Based on the decoded video stream acquired by the 36 cameras and fed to the detector module of each single camera, the first step is to determine whether full image re-detection is required based on whether the target position of the previous frame in that camera was acquired, and if not, then global detection is performed.
            \item  The method of setting the relevant strategy is to determine whether full image detection or equal interval detection of full image detection after detection failure, because full image detection often symbolizes the need to detect more regions of interest, which can be very consuming in terms of computation and time, so that the detection model-based tracking is a guarantee of real-time performance. The determination of global target detection or local detection is analyzed.
            \item   The whole detection module of the single machine is called if it is detected or not, and if not, tracking is performed, here tracking is based on local detection as the detection speed of the detector has reached 10 milliseconds. The authors also designed modules for the traditional SVM classifier and Kalman filter algorithm, but detection-based tracking plays a more accurate role in achieving tracking.
            \item   Multi-camera 3D reconstruction. Through the detection results of multiple cameras in the same frame, multi-camera 3D reconstruction is performed, i.e., the unique position of the target object in 3D is judged, and if the reconstruction fails, the next frame is detected, and here the reconstruction failure may often be that multiple angles of the target ball are obscured, and if the reconstruction is successful, the reconstructed 3D position is then back-calculated into the 2D image.

            \item   The correct 3D coordinate values are predicted. The algorithm of the whole system is to find out the value of the ball predicted by every two cameras in the visible camera, and if it is correct, the distance value calculated between no two is small and almost the same, and if one or several values are far from the position of the ball predicted by the other cameras, the camera is considered to have a false detection. If one or several values are far from the position of the ball predicted by other cameras, the camera is considered to have a false detection, and the 3D coordinates of the camera are eliminated and the beam difference method is calculated with the correct result, which will be more accurate.

            \item The results of 3D reconstruction are optimized for 2D. As the results of three-dimensional reconstruction back-calculated into two-dimensional is more accurate results for stand-alone detection, because this is the result obtained by collaborative processing, can be because of the optimization of the results of masking, missed detection, false detection and other cases, so for the local detection of two-dimensional image near the location, so that the false detection and missed detection will play an optimization role.

            \item Finally it is determined whether the result is ready for output. If the 3D reconstruction fails or the individual camera detection fails, i.e. the frame is not visible in all cameras, a frame is obtained and drawn in the area of the previous frame, then for the video stream is further loaded and if there is a valid output for the frame image, the visible camera number, the target 3D coordinate value of the visible camera and the target 2D coordinate value are output for the whole system for the subsequent video processing Rendering of target values.
            \item The viewer's interface and the researcher's interface are set aside to store the positions of all the balls in the video stream for subsequent video understanding. For example, this system will set aside 1.5 seconds to determine the stimulus event, then the exact trajectory value of the ball is especially important if there is a stimulus event.
        \end{itemize}

        \subsubsection{Design of visualization module}
        The visualization module is mainly divided into the visualization of the detection result picture and the visualization of the video sequence. The visualization module mainly has the effect of visualizing the reconstruction of the three-dimensional spatial course, first through visualization and targeted modifications until both offline and online can be done in real time.

        \subsection{Non-functional design}

        This system can be used properly mainly to match the accuracy and real-time, then this system needs to work on two aspects, and these two aspects are placed on an equal footing, and finally need to reach a balance, both to ensure accuracy in the case of look a accept, so that the system can be effective and real-time running time. 
        
        First is the issue of accuracy, this system is based on various types of neural network algorithms, using the YOLO algorithm based on YOLO's own design of dark net-based network migration to the open source framework caffe [in caffe, it is easier to increase the definition of relevant layers, increase the definition of relevant need layers, such as the design of data enhancement and loss layers, relevant pruning of the network [34], adding more contextual information to fully extract the features of each layer. Also, problems such as scattering, multiple angles of the target, and occlusion are effectively improved in the network design phase. Again, the data, in order to be accurate, then the data one is to be clean and accurate, the second is that the data for the environment to be more, while adding contextual information, the author's design is to obtain the output of the target, candidate frames and confidence and other information, after obtaining the output information, the relevant thresholds to determine whether the target ball, and then a single camera to obtain an accuracy, and then rely on multiple cameras for collaborative judgment. By the method based on the beam flow difference, here firstly, the smoothness of the target trajectory is ensured, secondly, most of the visible cameras' targets are correctly detected targets, here mainly for the collaborative judgment of error detection, and thirdly, by calculating the 3D of the target, and then projecting the 3D to the 2D of the visible camera, so that the target is on the local detector and get more accurate position information for for further verification.
        Again, the design is real-time. Live broadcast system, of course, needs to ensure real-time, in the design of the system, first to determine whether the accuracy of the ground can reach, can meet the requirements of the live broadcast, and then to meet the requirements of accuracy, then in maintaining the accuracy at the same time, but also to enhance the overall system of real-time, to the extent that it can be watched. Then the most time-consuming modules of this system are: the detection module of the detector, the detection of multiple cameras, and its 3D optimization module. Although this system certainly uses GPU parallel acceleration processing, but if the model is very large, loading images, etc. requires more computational load, then the ability to reduce the load on the computational server while maintaining speed is undoubtedly something to consider in the planning process of the product. 
        
        Finally, for the optimization of the detector, each call to the detector takes 10ms, then the final run will be the whole system running on 6 GPUs, making the whole system run time control within 200ms, which includes the judgment of the relevant events, greatly improving the efficiency of the whole system operation. In this paper, we only elaborate the optimization of multi-camera detection and detector, and do not explain how to run GPUs in parallel.

        \subsection{Summary}
        The system outline design in this chapter mainly introduces the overall processing flow of the system, and then introduces the design of each module one by one, i.e., the design of data pre-processing module, the design of detector module, the design of single-machine target detection module, the design of multi-machine detection and its 3D optimization module, and the design of visualization module. By introducing the overall activity diagram and the activity diagram of each module, the overall architecture and the interconnection of each module can be clearly shown, which lays a good foundation for how the whole system can operate effectively and the subsequent detailed design.
\newpage
%%%%%%%%%%%%%%%%%%%%%%%%%%%%%%%%%%%%%%%%%%%%%%%%%%%%%%%%%%%
\section{Detailed design of a soccer ball detection system with multiple cameras}	
\vspace{10.5cm}	
This chapter builds on the outline design of the multi-camera soccer ball detection system in the previous chapter. The design and implementation of the entire algorithm and framework is described in detail from the code point of view on the one hand, and the work done on multi-camera soccer detection is described in detail on the other hand, mainly by analyzing and describing the corresponding functions of each module in the entire system. At the same time, based on the design approach, reusable algorithms and system functional framework are defined, and finally a complete and feasible multi-camera soccer ball detection system solution is designed.

This chapter introduces the process of implementing detection algorithms in computer vision by familiarizing the problem to be solved and designing adaptive modules according to the characteristics of detection algorithms. At the same time, the system development adopts object-oriented analysis and design methods, the underlying code implementation are implemented in C++ code, and the open source vision library OpenCV is used as an auxiliary development tool for the system, one is easy to maintain and transplant the overall system framework, and the other can be compiled into static and dynamic link libraries, and the Python language is used in the upper layer of the framework, one is easy and fast to change the framework design, and the other is The developed system can be designed and tested several times to ensure the rapidity, adaptability, reusability and robustness of the system.

	\subsection{Overall system design and implementation}
		The difficulty of this system has three aspects, under the input of 36 video stream information, for the processing of numerous video image information, one is whether the detection of soccer is accurate and has the corresponding speed, the second is whether a single camera can accurately obtain the location of the target object, and the third is whether multiple cameras can collaborate to calculate the target object in 3D space. After obtaining the video through multiple cameras, all subsequent designs of this system are based on the video, then obtaining the foreground image through hybrid Gaussian modeling, obtaining the game area radiated by the camera through manual labeling of key points, then performing the cut of the rear view image, obtaining the area that may contain the target ball through a simple pixel threshold judgment, and then performing the various ways of single camera call detectors for detection . The final result is then visualized by iterative optimization through multiple camera calculations to obtain the final result.

\begin{figure}
            \centering
            \includegraphics[width=0.8\textwidth]{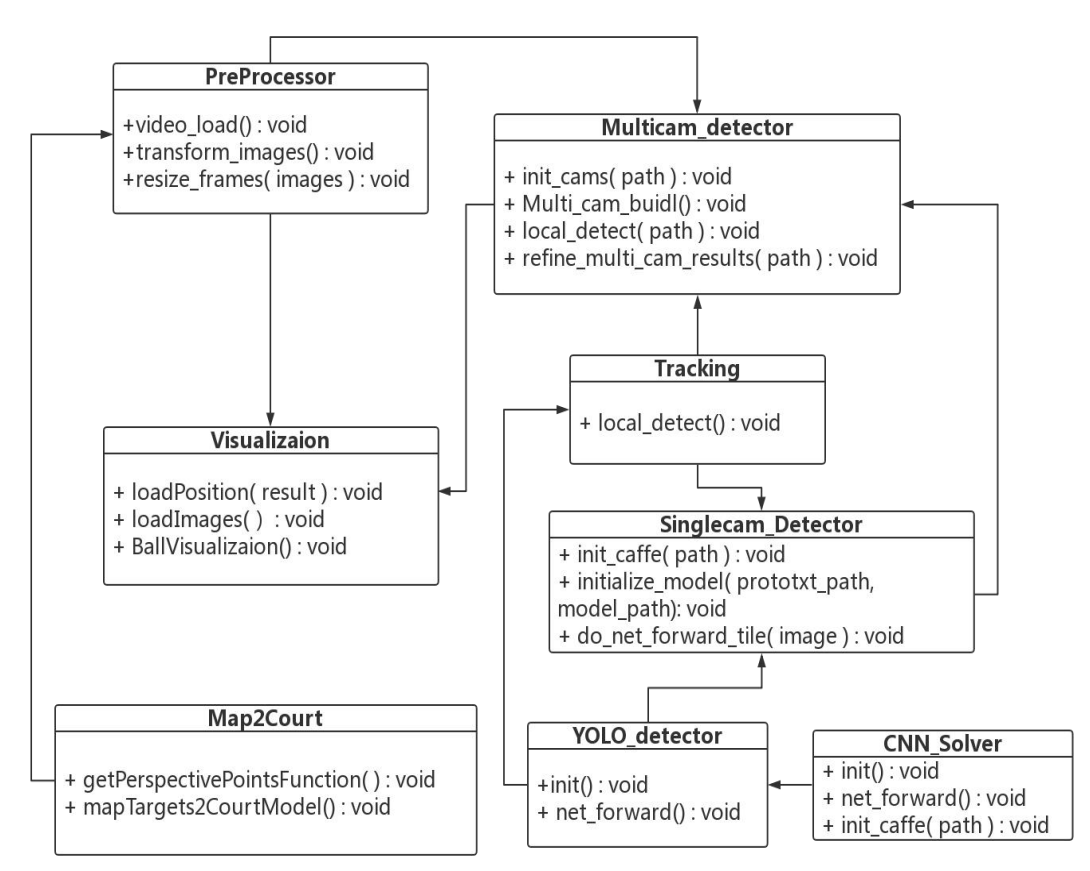}
            \caption{Relationship diagram of each module of the system}
            \label{fig:5.1}
        \end{figure}

        The core work of this system is performed on the server side. The advantage of server-based design and implementation is that only the input and output of each module needs to be focused on. The disadvantage is that it is not easy to perform relevant debugging through visualization, but it is possible to store the results, send them to the server for image visualization and image rendering to see if the overall effect meets the criteria, and then evaluate the relevant metrics from the data level. The system is divided into a data pre-processing module, a detector module, a single-machine target detection module, a multi-machine detection and its 3D optimization module, and a visualization module. The relationship class diagram between the modules of the whole system is shown in Figure \ref{fig:5.1}.

        Here are some illustration of the Figure \ref{fig:5.1}.
        \begin{itemize}
            \item Pre-processing module is the pre-processing class of the system, its work is mainly to load the video information transmitted by multiple cameras, video decoding, video decoding is mainly based on the open source tool ffmpg, according to the demand for relevant compression and resizing and other operations, this is in order to change the image channel in the subsequent implementation process can reduce the amount of relevant calculations, while the maximum frame rate of the video for storage, wait until the whole system for each camera capture results, the uncompressed image for video processing, visualization to the cloud for users to watch.
            \item  The visualization module is the final output module, which mainly does the loading of the target position calculated by the system, and the PreProcessor module loads the image into it, and then performs the related rendering. Here, we need to pay special attention to the fact that the module has an important method to judge and render the related events, such as the intelligent recognition and judgment of long passes, penalty kicks, corner kicks and shots, etc. Since the author is not involved in the development of this module, it is not proposed in this module.
            \item Map2Court module is similar to the offline calculation module, the main work is to take 36 cameras in the radiated area, divide the court, and then get the 3D spatial information of the court through the mapping relationship. Mainly by marking six key points to get the key coordinate points of each area of the court, through a simple Hough transformation, outline the edge of the court, etc..
            \item The tracking module in this system is based on the detection of tracking. When doing the tracking module, I tried relevant tracking algorithms such as ROLO algorithm, SVM classifier algorithm, Kalman filter algorithm, etc., but due to the presence of a lot of occlusion information and blurred information, none of them can be tracked effectively, so the tracking module is finally used to build a detection-based tracking model. The base class here is the YOLO detector class, which is in the video, doing intelligent video, there is a lot of work to do due to the continuous information of time, and this is the module that can be enhanced in the future, as long as the tracking module reacts fast enough, then the less time consuming, which has great economic and product benefits for enhancing at the algorithm level.
            \item CNN Solver module is the base class of the detector module, through which mainly defines the algorithm framework model that has been SSD and speedcnn, used to load the trained model for analysis and processing, the system in this paper finally uses the detector module based on the YOLO algorithm model framework. Subsequent work can also be tried and experimented on the sub-base, and constantly update the iterative product.
            \item YOLO detector class is the detector module, is based on the CNN Solver class, here the author in the design, YOLO detector can maintain good accuracy, so the other two classes in the system's module relationship diagram is not reflected, and the subsequent two detector base class, in the design of the detector. The detectors of other algorithmic frameworks were tested Because it is possible that different algorithms will have different effects for different scenarios, the relevant interfaces were also left after the module was written, which may be useful for other sports games, so here is a special note.
            \item Singlecam Detector class first because I ported the network structure and parameters involved in YOLO to caffe and added the corresponding layer structure in caffe, so first initialize caffe, then load the model and lock the cut out region to YOLO detector.
            This interface allows loading the trained model into the GPU quickly and performing multiple forward propagation. This class is therefore constantly optimized, as it is constantly called while utilizing the detector module.

            \item Multicam detector class keeps calling Singlecam Detector class and YOLO detector
            class, for the camera's visible target is divided into two cases, one is for the detected target in the vicinity of the local segmentation sent to the detector for judgment, which is more effective in this system, because the more tracking, then the less time consumed, can reduce response time and achieve real-time processing. Second is the re-detection of the whole map, this method will generally be in the stage when the video just comes out and the stage when the ball is not detected, and this stage is mainly for the Singlecam Detector class to do all the operations and implementations, mainly to prevent detection errors, i.e. long time errors.
            
        \end{itemize}

        \textbf{All the above modules are simple introductory modules, and the specific details of the core implementation of each class will be described in detail in subsequent sections.}

    \subsection{Design and implementation of data pre-processing}
    For the data preprocessing module, the author introduces the data processing modules involved in the preprocessing module, visualization module and Map2Court module in Figure 5.2.
    \begin{figure}
        \centering
        \includegraphics[width=0.9\textwidth]{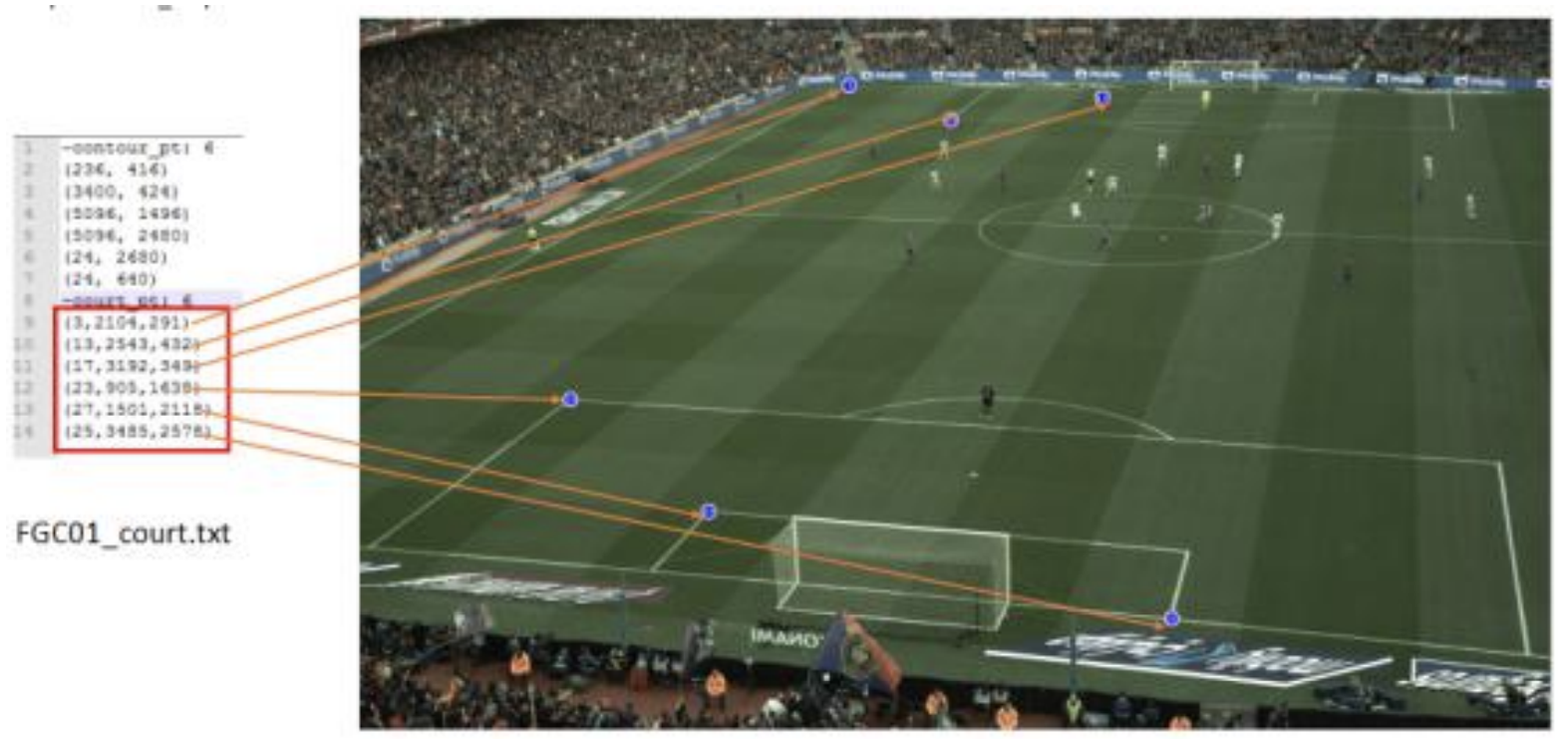}
        \caption{Key points of the court's map}
        \label{fig:5.2}
    \end{figure}

    As shown in Figure 5.2, by determining the key points of the court, the court lines and court sides are determined by the key points and the Hoff algorithm, and the court of the game is segmented out and sent to the next module for calculation. At the same time, the corresponding 3D coordinate values are obtained by obtaining information about the 3D position matrix of multiple cameras. Here it is necessary to explain why it is necessary to determine the key points of the court, and its main functions are as follows.

\begin{figure}
        \centering
        \includegraphics[width=0.3\textwidth]{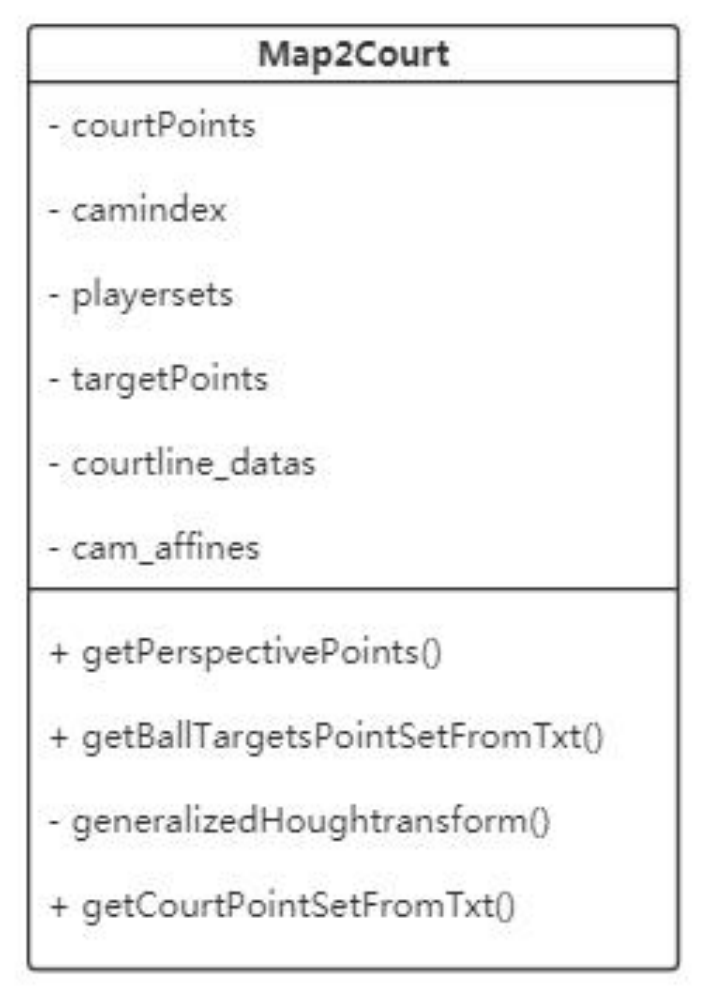}
        \caption{class diagram of the Map2Court class}
        \label{fig:5.3}
    \end{figure}

    \begin{itemize}
        \item Determine if the ball is out of bounds. The information in the diagram allows you to see the boundary of the entire court. The boundary is then determined and once the ball is judged to be out of bounds, no detection is made and the related shot can be converted to another location.
        \item  Culling the stadiums. After obtaining the key points, it is necessary to analyze only the pitch and then eliminate those redundant information such as spectator information, other player information of the pitch, etc. so that the detection of the whole system will be more accurate.
        \item Determine the relevant events. After getting information about the various types of sidelines in the stadium, such as goal line, midfield line, serving point, etc., then after getting information about the location of the ball, the definition of exciting shots such as shots, serves, corner kicks, long passes, etc. and whether the player crossed the line can be performed.
        \item Determining the position of the players. As a result of getting information about the whole stadium, when determining soccer goals, the author's research group has for players to analyze, and this part of information is very important for analyzing players.
    \end{itemize}
    Figure \ref{fig:5.3} shows the class diagram of the Map2Court class.
    
    Then we need to get the function description of Map2Court class.
    \begin{itemize}
        \item \textbf{courtPoints : }The key points of the access summation modeling are the key points obtained from subsequent calculations.
        \item \textbf{Cam index : } Numbering parameters of the camera.
        \item \textbf{playersets : } Player number parameters.
        \item \textbf{targetPoints : } Calculation of the obtained pitch key points.
        \item \textbf{courtline datas : } Stadium sideline description information.
    \end{itemize}

    From the above information, it can be effectively known that whether the module can obtain good court information is crucial for the subsequent acquisition of either target ball positioning capture or player information, because once the cutting is not accurate, there may be unclear judgment of out-of-bounds balls and inaccurate judgment of player information, so the module must be accurately positioned.

    \subsubsection{PreProcessor Module} \label{FFmpegRendering}
    This module is mainly for the subsequent convolutional neural network can judge the process, but also if the video stream information pre-processing process, one is for the video stream information decoding and encoding operations, using Nvidia hardware for FFmpeg hardware acceleration, the 36-way camera video information to get decoded and encoded. Secondly, in order to make the data input to the system network clean enough for faster data transfer between modules, and in order to make the texture information more prominent, the flip and resize interface will be added to the preprocessor, and based on this, downsampling and histogram equalization processing will be performed.

    \begin{figure}
        \centering
        \includegraphics[width=0.8\textwidth]{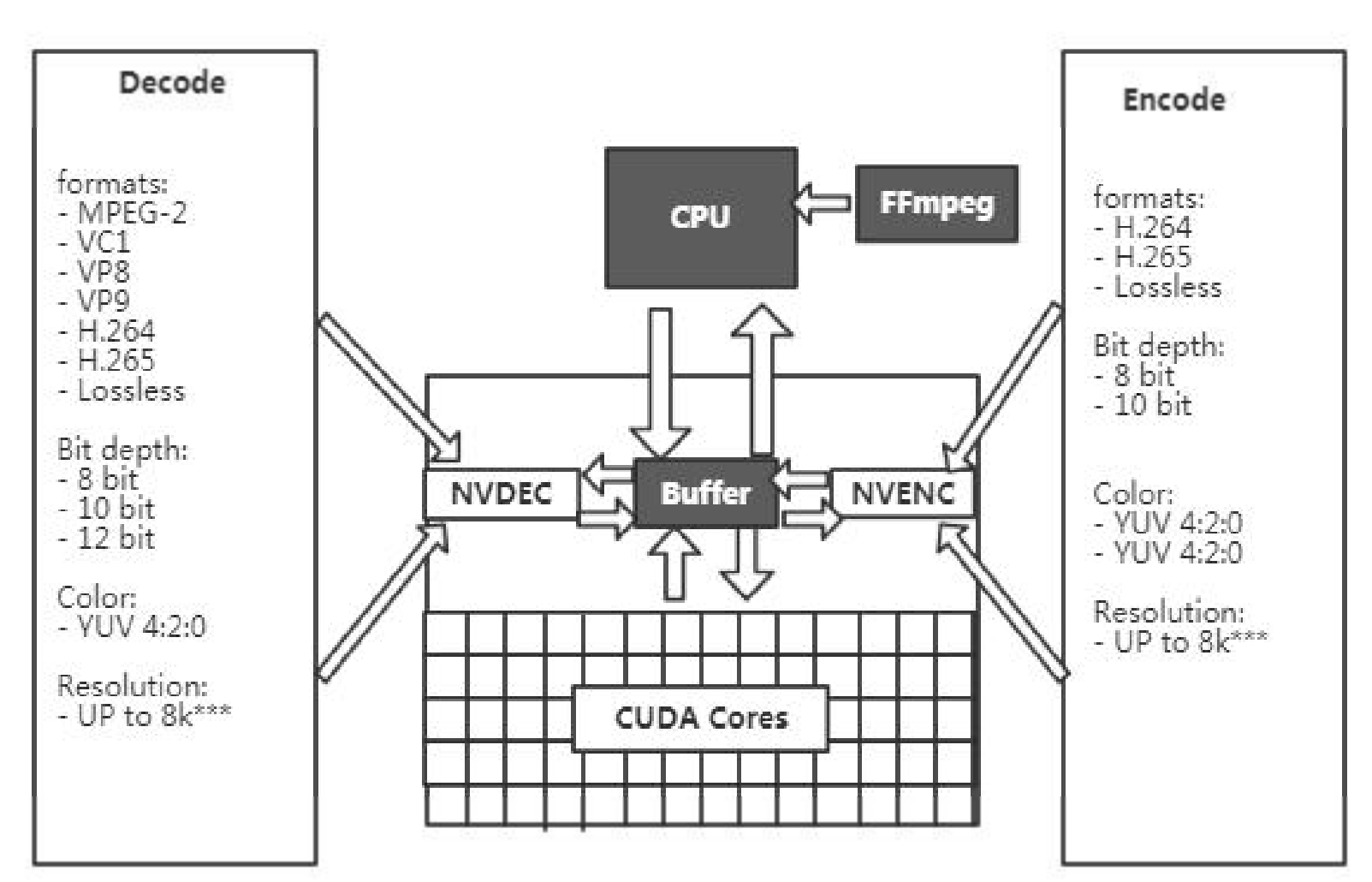}
        \caption{FFmpeg using Nvidia hardware systems}
        \label{fig:5.4}
    \end{figure}
    Because of the large amount of video information data, so using cpu for codec cannot effectively meet the subsequent system development process, so using Nvidia's GPU for FFmpeg hardware acceleration can well meet this system process, by calling the high-level interface Video Codec SDK in CUDA to call GPU resources, while the high-level interface has good The high-level interface has good packaging, source code and documentation support. As shown in Figure \ref{fig:5.4}, the Nvidai GPU can effectively use the CUDA core for encoding and decoding. FFmpeg's project also inherits the Video Codec SDK, and FFmpeg's configuration of this decoding encoder basically meets the requirements of this system. It should be noted here that when the video stream information is transmitted to the system, FFmpeg will decode twice at the same time, once to decompress the low resolution and send it to the detection module of multiple cameras in the system, and the other time to decompress the high resolution and get the relevant results of the system for rendering and visualization, so the use of GPU greatly improves the speed of the whole system, which also creates the conditions and foundation for real-time live streaming created the conditions and foundation. This also creates the conditions and foundation for real-time live streaming.

    \begin{figure}
        \centering
        \includegraphics[width=0.8\textwidth]{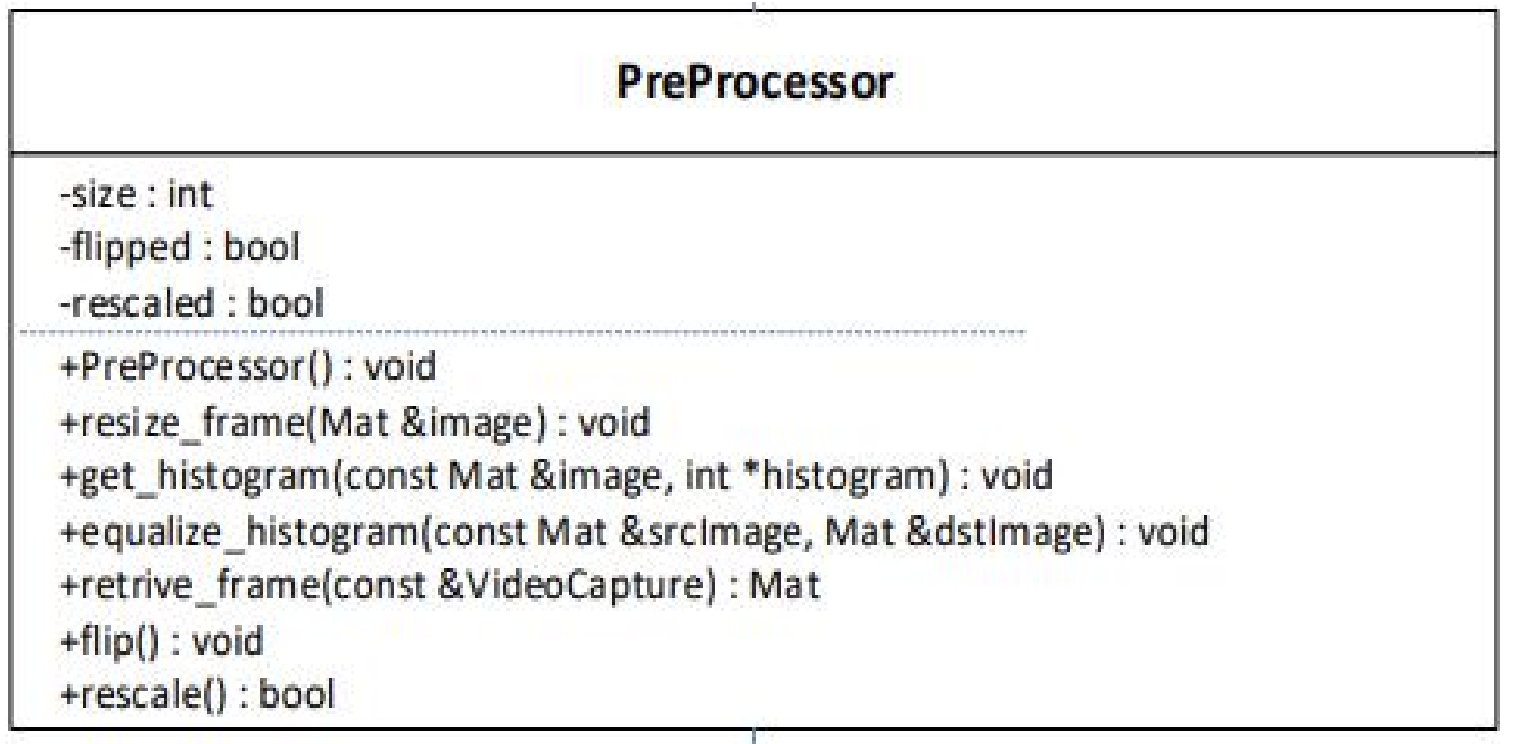}
        \caption{Class diagram of the preprocessing module}
        \label{fig:5.5}
    \end{figure}

    Figure 5.5 shows the further operation of the decompressed video image in the data processing module, whose core functions are resize, get\_histogram, equalize\_histogram and restore to the original image. The image resizing operation is determined by the input of the subsequent detection model, and the system uses an end-to-end detection algorithm based on the YOLO model to detect frame sizes from 417pixel*417pixel to 320pixel*320pixel and finally to 160pixel*pixel, where the corresponding speed from 20ms to 15ms accuracy is basically maintained at 96\%, which will be explained in detail subsequently.

    The module also provides an implementation of the background difference method, where the background is modeled by a Gaussian mixture and the background is updated at regular intervals.
    The background is modeled by a Gaussian mixture and the background is re-updated at regular intervals. This is done to prevent the effects of light and other influences during the race.
    This operation is done to prevent the influence of light during the race so that a large amount of information can be filtered by the background differencing method, and it is obtained by subtracting, binarizing, post-processing and then obtaining the result.

    When testing the filtering algorithm, due to the simplicity and efficiency of the background phase difference method, the author changes the background every 200 frames or so when performing the background subtraction method in order to overcome the constant change of the background during the game. In order to overcome the constant change of the background during the game, the author changes the background every 200 frames or so when performing background subtraction. This design is not only to ensure the robustness of the system, but also to reduce the detection time of the whole image due to the excessive region of interest after background subtraction. This design not only ensures the robustness of the system, but also reduces the detection time of the whole image due to too many regions of interest after background reduction.

    \subsubsection{Visualization Module}

    \begin{figure}
        \centering
        \includegraphics[width=0.8\textwidth]{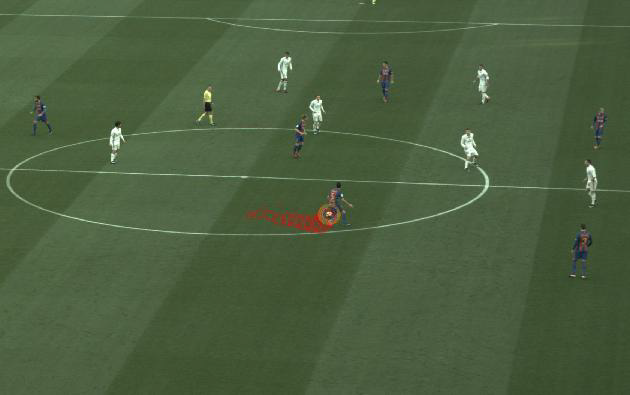}
        \caption{Visualization map of the ball target.}
        \label{fig:5.6}
    \end{figure}
\begin{figure}
        \centering
        \includegraphics[width=0.6\textwidth]{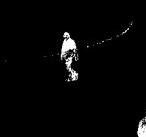}
        \caption{The background segmentation map.}
        \label{fig:5.7}
        \end{figure}

        video encoding library libavcode, but due to the special nature of this system, there will be a large amount of input and output video data streams. As described in Section \ref{FFmpegRendering}, the hardware acceleration of FFmpeg can be used to decode the low-resolution video streams to be fed into the detection module of this system. The next step is to decode the high-resolution images for subsequent live streaming. The high-resolution images are used for subsequent rendering and visualization during the live broadcast. Then for the visualization of the detection module, an associated interface is provided to save the image processing for each stage of the visualization, and finally the role of this module is to render and capture at the decoded high resolution, Figure \ref{fig:5.6} shows the visualization of a frame of the target ball capture, and Figure \ref{fig:5.7} shows the rear view cut of the player on the court.

    \subsubsection{Data structure}

\begin{figure}
        \centering
        \includegraphics[width=0.5\textwidth]{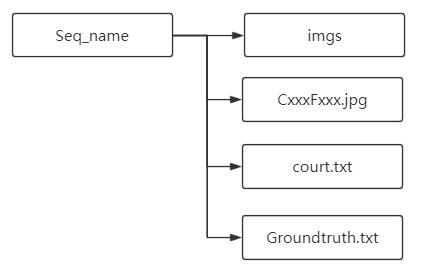}
        \caption{Data Structure of the original source.}
        \label{fig:5.9}
        \end{figure}

    Since this system has a large amount of video information and court information, a well-defined data structure can effectively implement the functions of the modules of this system, specifying the form of incoming and accessible data, as shown in the data structure diagram in Figure \ref{fig:5.9}.
    then, in the data structure form, there are some definition of them:
    \begin{itemize}
        \item Seq\_name: indicates the number of each camera.
        \item Imgs: folder must have access to the relevant images, where the images are read and accessed with serial information read and accessed with serial information.
        \item CxxxFxxx.jpg: Access is to the background image of this camera.
        \item court.txt: Access to key points in the course.
        \item Groundtruth.txt: The set of accurately marked target points contained in the sequence.
        
    \end{itemize}

    Specifically, for \textbf{Imgs Folder}:
    \begin{itemize}
        \item The background image is stored at the top of the data folder, if this file is not available, you need to do a background build If this file is not available, you need to do background modeling to get a background image.
        \item Formatted as jpg
        \item The name format is CxxxFxxxx.jpg, where Cxxx: means the      camera number is xxx (3-digit integer)
            Fxxxx: means that the frame number is xxxx (4 digits)
            Example: C018F0001.jpg
        
    \end{itemize}

    \begin{figure}
        \centering
        \includegraphics[width=0.3\textwidth]{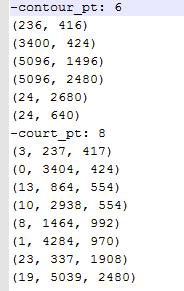}
        \caption{Diagram of court.}
        \label{fig:5.10}
    \end{figure}
    \textbf{Court File}, as shown in Figure \ref{fig:5.1}.The content consists of two parts. 

    \begin{itemize}
        \item Court contour points:
            \begin{itemize}
                \item contour\_pt: <num points>
                \item (x\_1, y\_1)
                \item (x\_2, y\_2)
                \item ...
                \item (x\_n, y\_n)
    
            \end{itemize}
        \item 
                Court line points:
                \begin{itemize}
                    \item -court\_pt: <num points>
                    \item (pi\_idx, x\_1, y\_1)
                    \item (pi\_idx, x\_2, y\_2)
                    \item ...
                    \item (pi\_idx, x\_n, y\_n)
                \end{itemize}
        \item If the file is missing, the pitch cannot be split out.
    \end{itemize}       
    \begin{figure}
        \centering
        \includegraphics[width=0.25\textwidth]{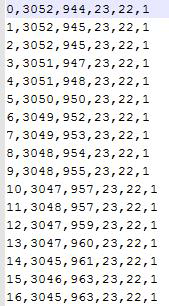}
        \caption{Diagram of court.}
        \label{fig:5.11}
    \end{figure}
    \textbf{Groundtruth File}, as shown in Figure \ref{fig:5.11}.
    \begin{itemize}
        \item Named groundtruth.txt
            The data are in the form: frm\_no, x, y, w, h, vis
            where, frm\_no: frame number, starting from 0
        \item x, y, w, h: the frame of the target object, the coordinates of the upper left corner and the width and height of the frame
            Vis: 0: target is not visible
            1: target is visible
    \end{itemize}

\subsection{Design and implementation of target detector}
    \subsubsection{Basic model framework for target detectors}
    The system is designed and implemented using a convolutional neural network structure, mainly with reference to the YOLO detector network prototype for optimization and modification, adapted to the purpose of this system for soccer detection. The system is based on the YOLOv2 network. The basic network structure of YOLOv2 and its parameters are shown in Table 5.2. The network structure darknet19 is shown in Table 5.2.

    From the table, we can see that the YOLOv2 network designed based on Darknet-19 can basically satisfy the effect of feature extraction and target detection. And the basic principle of the network designed based on this system is to adjust the network structure. On the basis of this, optimization is performed. The network structure mainly refers to the last layer of the convolutional layer, which extracts the information. The network structure mainly connects the last layer of the convolutional layer that extracts information to the penultimate 3rd, 5th and 7th layers, which has greatly improved the detection accuracy of small targets.

    %\ ignore the table5.2

    \subsubsection{Optimization of the detector}
    \begin{itemize}
        \item \textbf{Network setup.} The base network for YOLO is Googlenet, and is based on the network in network principle.
        \item \textbf{Optimize the mechanism of Anchor Boxes.} YOLO v2 defines different aspect ratios and scales of each size, i.e., a central judgment of different scales and aspect ratios of boxes. since the size ratio of the ball is relatively fixed, the aspect ratio setting of Anchor Boxer is reduced to 3 kinds here, so that its multi-scale forced to fit to that kind of scale, and the experiment proves that it can reduce the corresponding time and does not lose accuracy. It is important to emphasize that this system uses both industry detection methods together, i.e., RPN-based detection and Anchor-based detection, where RPN-based means that the region of interest is segmented during image preprocessing, and the pixel size comparison is used to very cleverly cut out images where targets may exist.
        \item \textbf{Dimensional clustering and direct location prediction.} Using K-means clustering method class training bounding boxes can automatically find better boxes width-height dimension. In the predicted position, the width and height of the Anchor and the corresponding distance are forcibly constrained. After constraining the proportion of one, the non-target objects can be eliminated, and the second is the end-to-end implementation of the fitting and detection process.
        \item \textbf{Detection of fine-grained feature soccer. } The network acquires the features of the previous layer of features and combines them with the final output features, in order to optimize the detection of fine-grained soccer balls, adding more fine-grained features to combine the features of the penultimate layer as well. Also, the 7x7 grid in the last layer is set to 9x9 grid, which is an effective improvement for fine-grained targets. The following is the construction of the Deep Residual net. We base resnet, which is a popular framework module structure. It can be seen that the whole transfer layer can connect different layers, and the feature maps of different resolutions can be connected once, so that after the connection, the features in different channels can extract better fine-grained features by expanding the feature maps, which finally makes a 1\% improvement.
        \item \textbf{Add context information.} According to the characteristics of convolutional neural network extracted features, the design reduces false and missed detections for substantial occlusion and motion defocusing. As the trained data are processed as shown in Figure \ref{fig:5.13}.

        \begin{figure}
            \centering
            \includegraphics[width=0.5\textwidth]{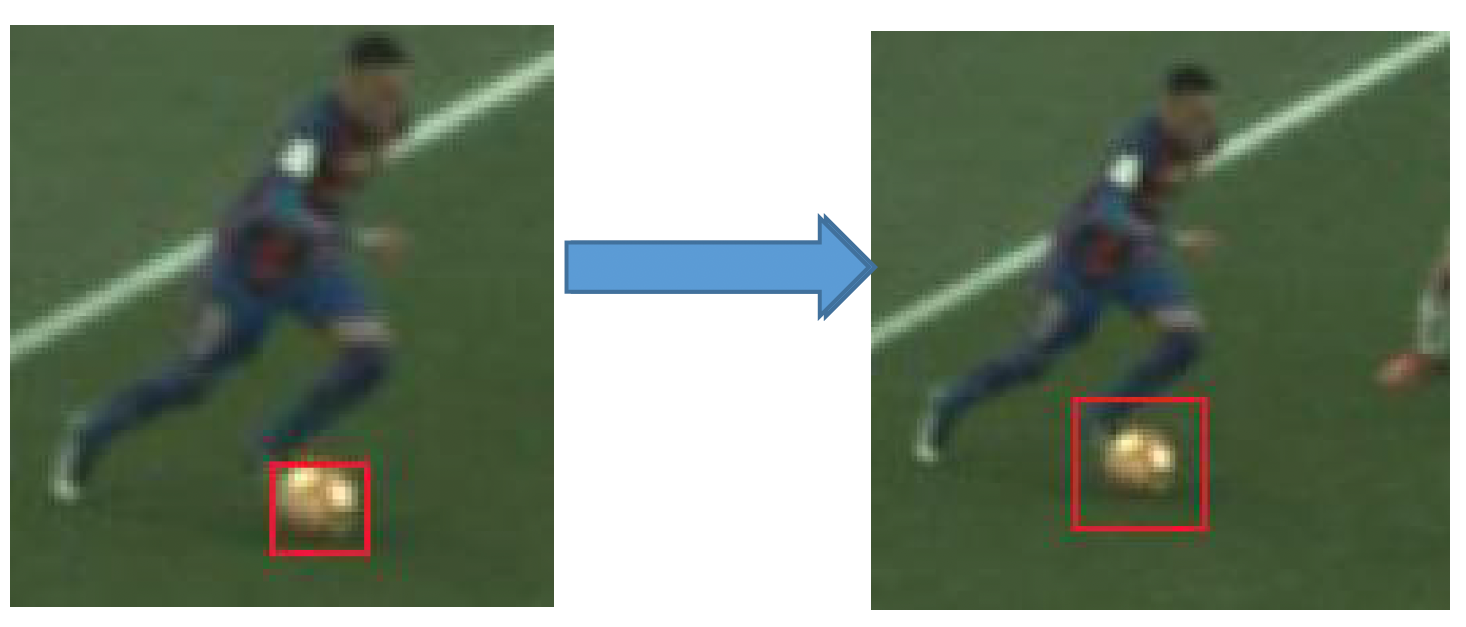}
            \caption{Adding contextual information to soccer}
            \label{fig:5.13}
        \end{figure}

        \item \textbf{Design Linked loss function.} Deep convolutional neural networks are supervised to converge by loss function, which is a crucial step in the whole network training process. The design of loss function is also a hot research topic, and the loss functions of different projects are designed to supervise the convergence of different targets during training. The loss function of this system is.

        \begin{equation}
            \mathrm{L}(x, c, l, g) = \frac{1}{N}\left(L_{\text{conf}}(x, c) + \alpha L_{\text {loc}}(x, l, g)\right)
        \end{equation}
        
        where L denotes the overall loss function, x denotes the sign bit tensor of whether the default box and the true target location box are consistent at all feature map locations, c denotes the offset of the predicted target location with respect to the default box, l denotes the predicted box, and g denotes the true target location. Here, it is mainly composed of the localization loss function and the classification loss function. Among them, YOLO itself detects the loss function of each frame, and readers can refer to YOLO related papers.
    \end{itemize}

    \subsubsection{Porting the caffe framework and adding related layers}
    Since the network design is trained and tested on darknet, it is not highly reusable for future improvement and porting of other algorithms, so what we have done is to port the network design to caffe and add the layer of data augmentation fuzzy. Due to the limited space, the steps and algorithm flow for adding the corresponding layers are not expanded.

    \subsubsection{Model training and model compression}
    This subsection is the most important module for this system to be able to run efficiently, because the system can run efficiently, the detector must be to run with a small load and run fast, for the operation of the images obtained from so many cameras. The following describes the entire training process of the model and the compression process of the model.

    \begin{itemize}
        \item \textbf{model training.}
            \begin{itemize}
                \item Collation of sample data. The training data need to be extracted from each camera for labeling, i.e., the training groundtruth data. The main data are distant targets, occluded targets, blurred targets and normal targets.
                \item Image size decision. Image size decision.
                Through the previous data preprocessing module, an image of 5120pixel x 3072pixel will be obtained. In order to reduce the load, the resize is designed to be half of the image, i.e. 2560pixel x 1536pixel, at this time the size of the target ball in the court is about 10pixel x 10pixel to 5pixel x 5pixel. So the models are defined from 417 size input mods to 320 size input mods and finally to 160 size input mods.

                $$
                \begin{array}{|l|l|}
                \hline \text { size of image } & \text { size of detector } \\
                \hline 5120 \text { pixel x 3072pixel } & 320 \text { pixel x 320pixel } \\
                \hline 2560 \text { pixel x 1536pixel } & 160 \text { pixel x 160pixel } \\
                \hline
                \end{array}
                $$

            \end{itemize}

        \item \textbf{Training data pre-processing.} The data preprocessing for training is to convert the original sample image into the desired form and size.

        \begin{figure}
            \centering
            \includegraphics[width=0.8\textwidth]{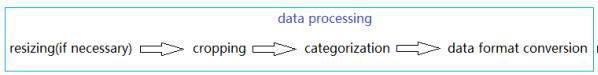}
            \caption{Training data pre-processin.}
            \label{fig:5.14}
        \end{figure}

        \begin{figure}
            \centering
            \includegraphics[width=1.0\textwidth]{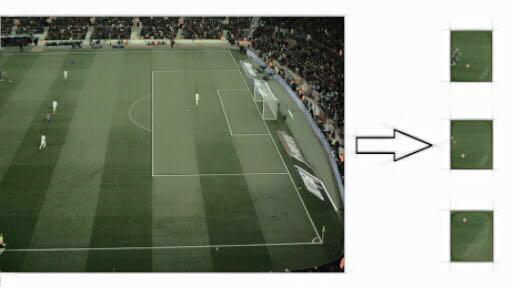}
            \caption{Crop augmentation operation.}
            \label{fig:5.15}
        \end{figure}

    \item \textbf{The operation of slicing data.}
    As shown in Figure \ref{fig:5.15}, the position of the ball in the whole image is sliced into 320 pixel x 320pixel or 160pixel x 160pixel to fit the size of the model. It is important to mention that the size of the training data often determines the final The size of the training data often determines the size of the final system input and output, i.e., the model corresponding to 320 pixel x 320 pixel tends to work well in the original image. The model corresponding to 320 pixel x 320 pixel tends to work well in the original image, while the model corresponding to 160 pixel x 160 pixel works better for the image after resize. The model corresponding to 160 pixel x 160 pixel works better for the resized image.

        \begin{figure}
            \centering
            \includegraphics[width=0.8\textwidth]{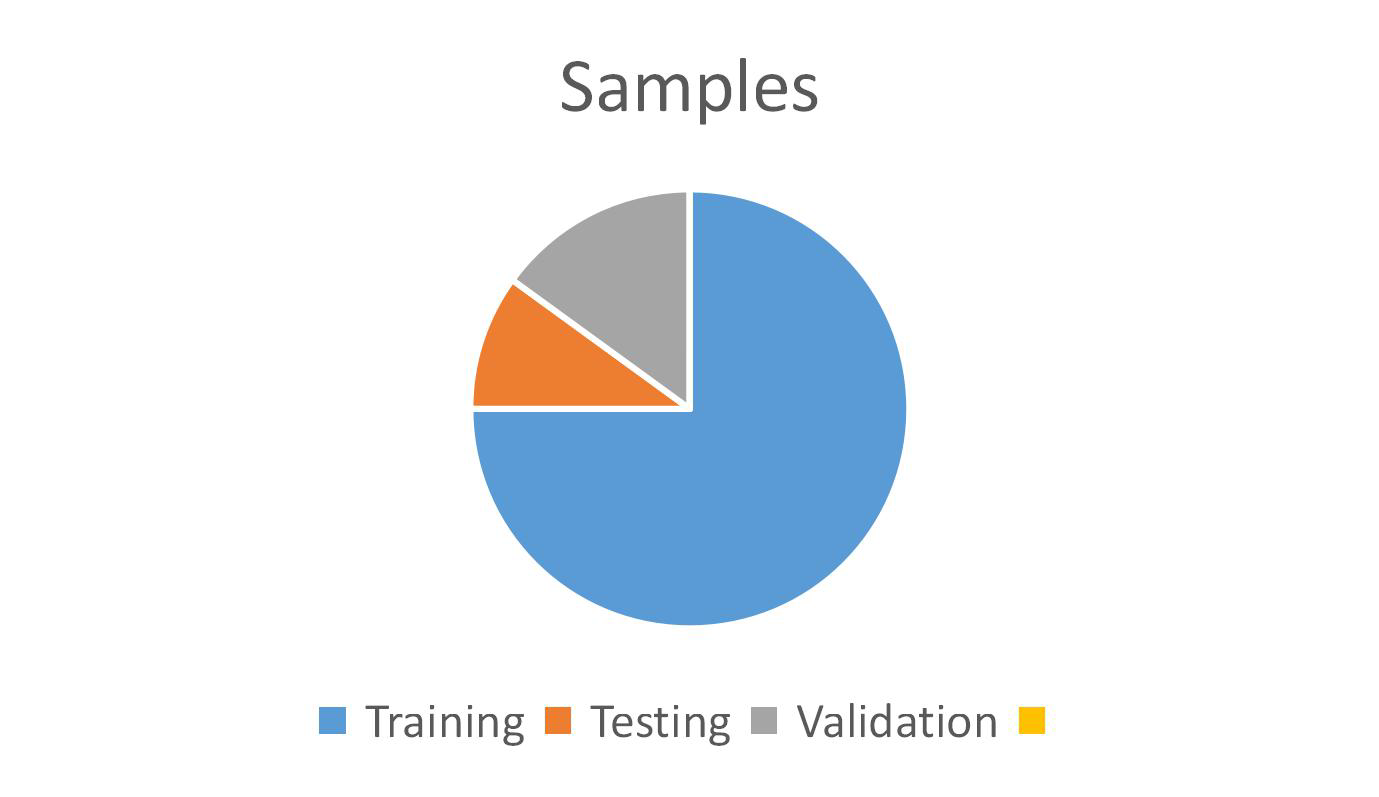}
            \caption{Data Classification Chart.}
            \label{fig:5.16}
        \end{figure}

    \item \textbf{Data distribution for training, testing and validation.}
    The sample is divided into three parts, where the training data is mainly for training out a model, the test is for to calculate the accuracy and the associated score values, and the validation set is to prepare later for whether the test is consistent, and the data The scale of the data is shown in Figure \ref{fig:5.16}.
    \end{itemize}
    
    \subsubsection{Design and implementation of single camera detection module.}

        \begin{figure}
            \centering
            \includegraphics[width=0.9\textwidth]{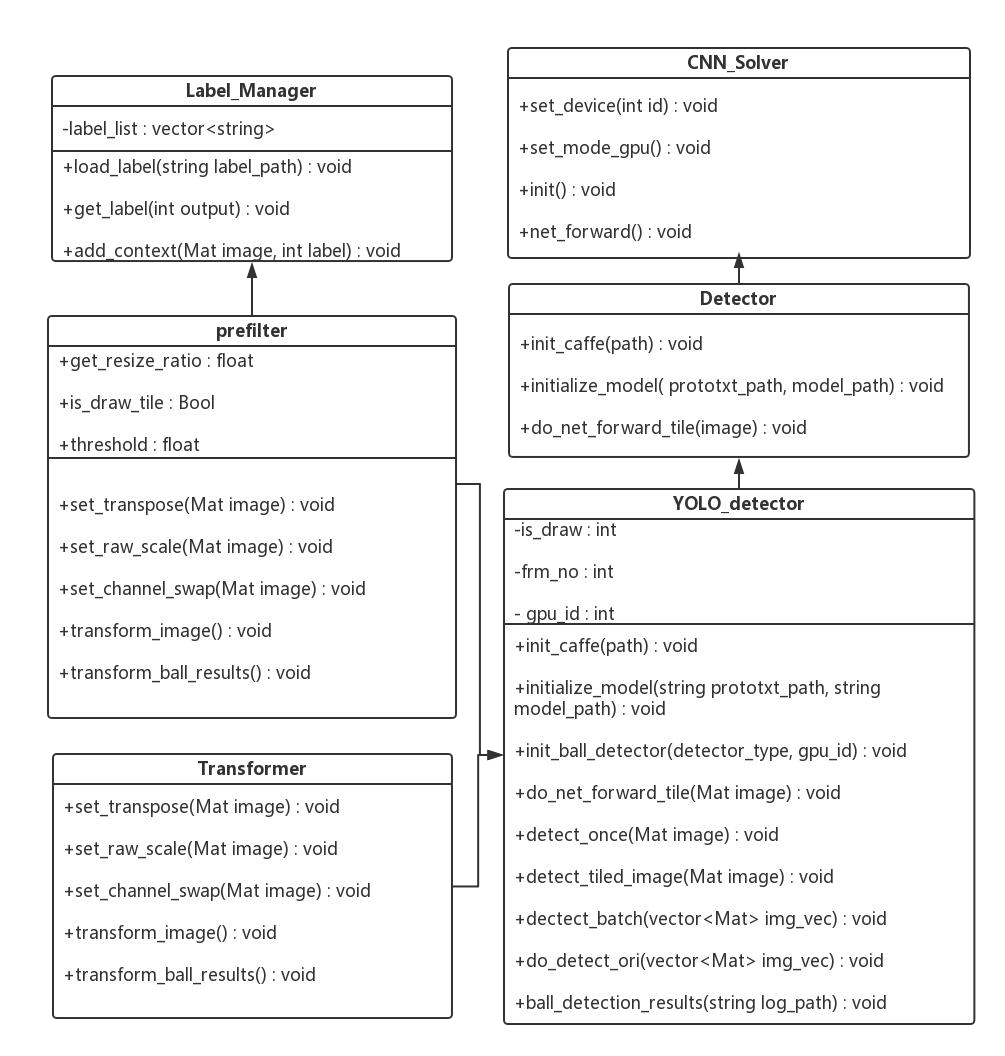}
            \caption{Single camera module class diagram.}
            \label{fig:5.17}
        \end{figure}
    
    The implementation structure of the single camera detection module can be seen in Figure \ref{fig:5.17}, which is the class diagram of single camera detection, and it can be seen from the diagram that the final implementation of this module is the YOLO\_detector class, which inherits from the Detector class.
    The CNN\_Solver class implements the basic operation interface of the convolutional neural network.
    The CNN\_Solver class implements the basic operation interface of the convolutional neural network, which can load the model and perform some basic operations such as convolution and pooling and nonlinearity, while the implementation of do\_net\_forward in the Detector class can output the detection results, i.e.
    The implementation of do\_net\_forward in the Detector class can output the detection result, which is a forward propagation.

    Among them, the prefilter module and transformer module are both implementations of the input detector. The prefilter class implements a simple background think subtraction to get a grayscale image, and a simple threshold judgment to frame the area of interest, where the parameter get\_resize\_ratio is to decide whether to perform the resize operation, and is\_draw\_tile indicates whether to call YOLO\_detector to frame the target location, which can be well visualized for subsequent error analysis and testing, and can be targeted for error detection. Optimization and improvement at the data and algorithm level.

        \begin{figure}
            \centering
            \includegraphics[width=1.0\textwidth]{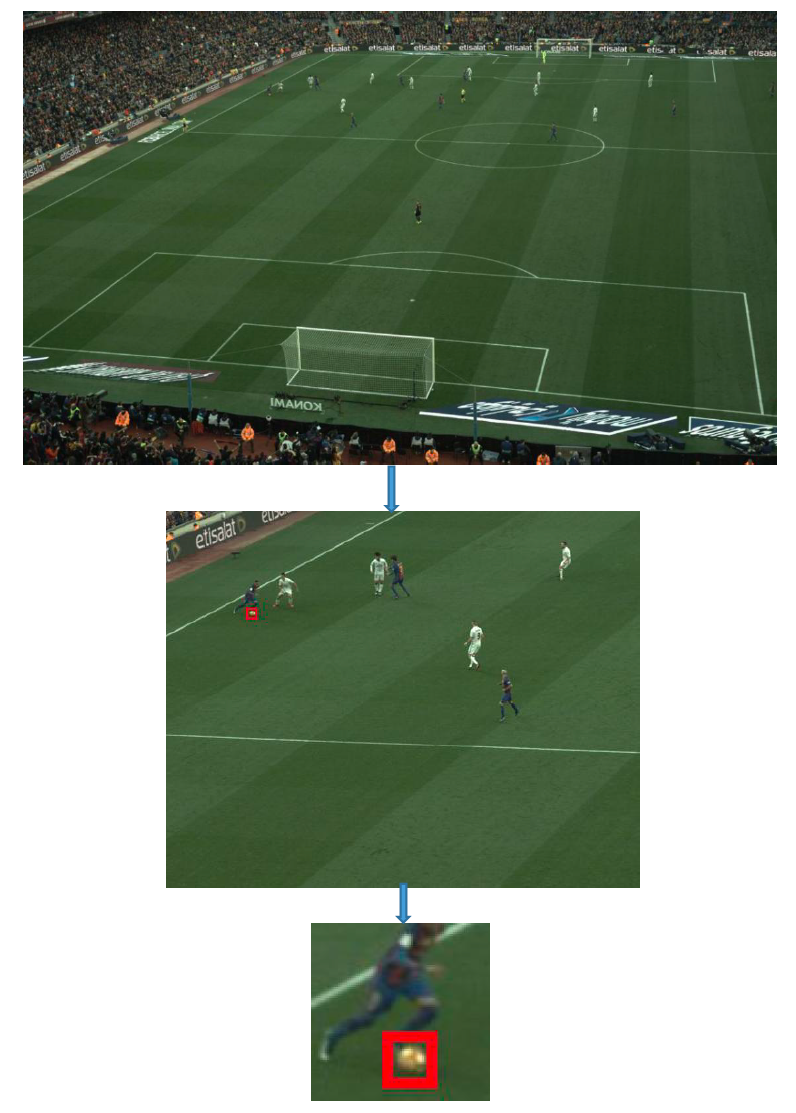}
            \caption{Single camera test results.}
            \label{fig:5.18}
        \end{figure}
    Figure \ref{fig:5.18} shows the results of the detection with the YOLO detector in a single camera. It can be seen that the target occupies less of the original image, but it can be effectively detected, and the experiment proves that the resize into a small image and the corresponding small detector can also effectively detect the correct target. From Figure 5.18, it can be seen that the whole single camera detection sphere is a step-by-step process, which first identifies the region of interest in the whole figure and then sends it to the detector for judgment.

    \subsection{Design and implementation of multi-camera detection module.}
    This subsection describes how to perform the construction of multi-camera-based 3D vision, the implementation of video-based detection, and the development and implementation of an overall multi-camera detection framework.

    \subsubsection{Description of the layout of multiplex cameras in space.}
    
        \begin{figure}
            \centering
            \includegraphics[width=1.0\textwidth]{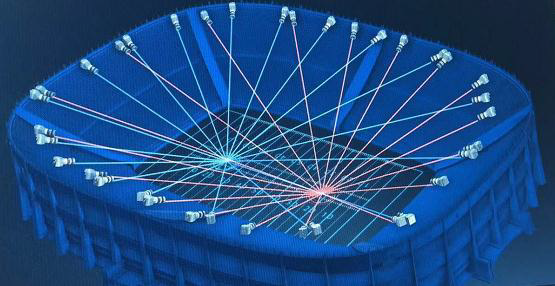}
            \caption{36 cameras layout in space.}
            \label{fig:5.19}
        \end{figure}

    From Figure \ref{fig:5.19}, we can see that the 36 cameras are fixed in space on the buildings around the court, and each camera The area that each camera can radiate is fixed, and the camera itself is also fixed, so in getting the ball in the three-dimensional space Therefore, the position of the ball in the 3D space corresponds to a fixed position of each camera. At the same time, a certain area in the 36-way camera will have This is also based on the consideration of multi-camera detection algorithm, so this 36-way camera in The author's company has high position requirements and accuracy requirements when setting camera positions. Since this system can be reused for basketball and rugby games, so the whole 36-camera setup is unified, which also provides the possibility for subsequent This provides the possibility and convenience for subsequent algorithm and system migration. Based on the system's ability to obtain the ball's position in real time, the 3D rendering can be done. After acquiring the ball position in real time, the rendering in 3D is also based on the 36-way camera, and the rendering of actions and events in 3D space is also based on the results of this system. The rendering of actions and events in 3D space is also based on the results of this system, and I will not repeat the rendering work in this paper.

    \subsubsection{Construction of multi-camera based 3D vision.}
    In Chapter 2, the principle description of the 3D visual construction is introduced, using the bundle Adjustment (BA) algorithm prototype, which requires relevant modifications and optimizations here, namely the generalized Sparse Bundle Adjustment (SBA) implementation. Since the radiating areas of the camera overlap, but also have their own camera areas that are not visible to the camera, the reprojection error can be effectively minimized using this algorithm, but in this system, the voting mechanism is required, i.e., there is a possibility that the detection of 2D images is wrong, then for sending to the BA algorithm In this system, however, a voting mechanism is required, i.e., there is a possibility that the detection of the 2D image is wrong, and then for sending to the BA algorithm, a judgment and rejection is required.

        \begin{figure}
            \centering
            \includegraphics[width=0.5\textwidth]{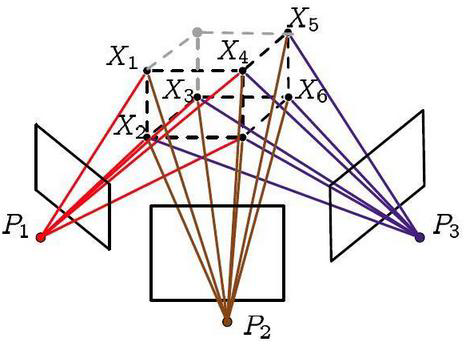}
            \caption{BA Algorithm Model.}
            \label{fig:5.20}
        \end{figure}

    As shown in Figure\ref{fig:5.20}, the BA algorithm is an optimized model, where the points where multiple beams cross are the points in 3D, and the observation Pi is the projection of the same spatial point P, the projection \^P of P and the observation P\_i have a certain distance between them. This is the reprojection error. The system cannot adjust the position of the camera to make the distance smaller, so the system only consider how to minimize the error at each point, where the criterion is to use multiple correctly detected results to reconstruct The criterion is to reconstruct the result with the smallest error using multiple correct detections.

        \begin{figure}
            \centering
            \includegraphics[width=1.0\textwidth]{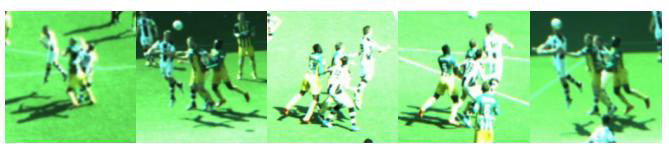}
            \caption{Ball detection in multiple cameras for the same target scene.}
            \label{fig:5.21}
        \end{figure}

    As shown in Figure \ref{fig:5.21}, the scene is the detection of multiple visible cameras for the same scene. As can be seen from the figure it is clear that the target sphere is different under each camera, and it is this that gives the 3D basis.

    Considering the present system, there may be invisible points, and here to accommodate the sparsity of the whole large area of the system Suppose that for the object with n=4 points, there are three angles of the above-mentioned pictures, i.e. m=3, then we can set the viewing The observation coordinates X and the parameters P of the camera are, respectively

    \begin{equation}
    \begin{aligned}
    & X=\left(x_{11}^T, x_{12}^T, x_{13}{ }^T, x_{21}{ }^T, x_{22}{ }^T, x_{23}{ }^T, x_{31}^T, x_{32}{ }^T, x_{33}{ }^T, x_{41}^T, x_{42}{ }^T, x_{43}{ }^T\right)^T \\
    & P=\left(a_1^T, a_2^T, a_3^T, b_1^T, b_2^T, b_3^T, b_4^T\right)^T
    \end{aligned}
    \end{equation}

    There is no intersection between the parameters in each camera and the points in 3D for the time being, so the bias derivatives of the 2D images belonging to the current camera in coordinates are all 0, and the bias derivatives of the current 3D points and the other 2D images in coordinates are also 0. The bias derivatives of the current 3D point and other 2D images are also 0, i.e..:

    \begin{equation}
        \begin{aligned}
            & A_{i j} =\frac{\delta x_{i j}^{\prime}}{\delta a_k}=0, \forall j \neq k \\
            & B_{l j} =\frac{\delta x_{j j}^{\prime}}{\delta b_k}=0, \forall i \neq k
        \end{aligned}
    \end{equation}

    Then it is the LM algorithm (Levenberg-Marquardt) that is used to find the iteration step of the above equation.

    \begin{equation}
    \delta^T=\left(\delta_a^T, \delta_b^T\right)^T=\left(\delta_{a 1}^T, \delta_{a 2}^T, \delta_{a 3}{ }^T, \delta_{b 1}^T, \delta_{b 2}{ }^T, \delta_{b 3}{ }^T, \delta_{b 4}{ }^T\right)^T
    \end{equation}

    And the process of finding partial derivatives at this point is also the problem of solving Jacobi matrices.

    \begin{equation}
    \frac{\partial \mathbf{X}}{\partial \mathbf{P}}=\left(\begin{array}{ccccccc}
    \mathbf{A}_{11} & \mathbf{0} & \mathbf{0} & \mathbf{B}_{11} & \mathbf{0} & \mathbf{0} & \mathbf{0} \\
    \mathbf{0} & \mathbf{A}_{12} & \mathbf{0} & \mathbf{B}_{12} & \mathbf{0} & \mathbf{0} & \mathbf{0} \\
    \mathbf{0} & \mathbf{0} & \mathbf{A}_{13} & \mathbf{B}_{13} & \mathbf{0} & \mathbf{0} & \mathbf{0} \\
    \mathbf{A}_{21} & \mathbf{0} & \mathbf{0} & \mathbf{0} & \mathbf{B}_{21} & \mathbf{0} & \mathbf{0} \\
    \mathbf{0} & \mathbf{A}_{22} & \mathbf{0} & \mathbf{0} & \mathbf{B}_{22} & \mathbf{0} & \mathbf{0} \\
    \mathbf{0} & \mathbf{0} & \mathbf{A}_{23} & \mathbf{0} & \mathbf{B}_{23} & \mathbf{0} & \mathbf{0} \\
    \mathbf{A}_{31} & \mathbf{0} & \mathbf{0} & \mathbf{0} & \mathbf{0} & \mathbf{B}_{31} & \mathbf{0} \\
    \mathbf{0} & \mathbf{A}_{32} & \mathbf{0} & \mathbf{0} & \mathbf{0} & \mathbf{B}_{32} & \mathbf{0} \\
    \mathbf{0} & \mathbf{0} & \mathbf{A}_{33} & \mathbf{0} & \mathbf{0} & \mathbf{B}_{33} & \mathbf{0} \\
    \mathbf{A}_{41} & \mathbf{0} & \mathbf{0} & \mathbf{0} & \mathbf{0} & \mathbf{0} & \mathbf{B}_{41} \\
    \mathbf{0} & \mathbf{A}_{42} & \mathbf{0} & \mathbf{0} & \mathbf{0} & \mathbf{0} & \mathbf{B}_{42} \\
    \mathbf{0} & \mathbf{0} & \mathbf{A}_{43} & \mathbf{0} & \mathbf{0} & \mathbf{0} & \mathbf{B}_{43}
    \end{array}\right)
    \end{equation}

        The covariance matrix of the matrix X can be observed in the diagonal structure.

        \begin{equation}
        \Sigma_{\mathbf{X}}=\operatorname{diag}\left(\Sigma_{\mathbf{x}_{11}}, \Sigma_{\mathbf{x}_{12}}, \Sigma_{\mathbf{x}_{13}}, \Sigma_{\mathbf{x}_{21}}, \Sigma_{\mathbf{x}_{22}}, \Sigma_{\mathbf{x}_{23}}, \Sigma_{\mathbf{x}_{31}}, \Sigma_{\mathbf{x}_{32}}, \Sigma_{\mathbf{x}_{33}}, \Sigma_{\mathbf{x}_{41}}, \Sigma_{\mathbf{x}_{42}}, \Sigma_{\mathbf{x}_{43}}\right)
        \end{equation}

\begin{figure}
            \centering
            \includegraphics[width=0.5\textwidth]{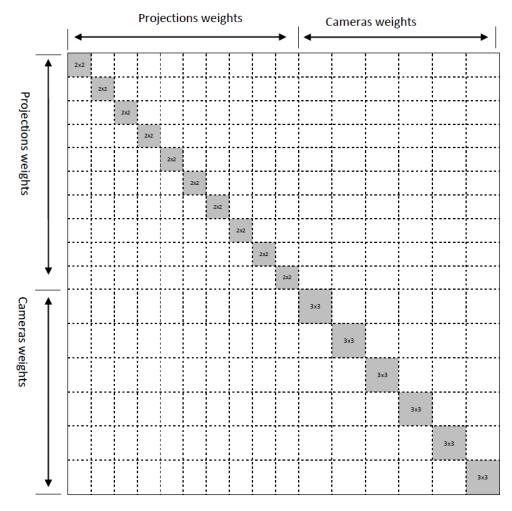}
            \caption{BA Weighting Matrix.}
            \label{fig:5.22}
        \end{figure}

\begin{figure}
            \centering
            \includegraphics[width=0.5\textwidth]{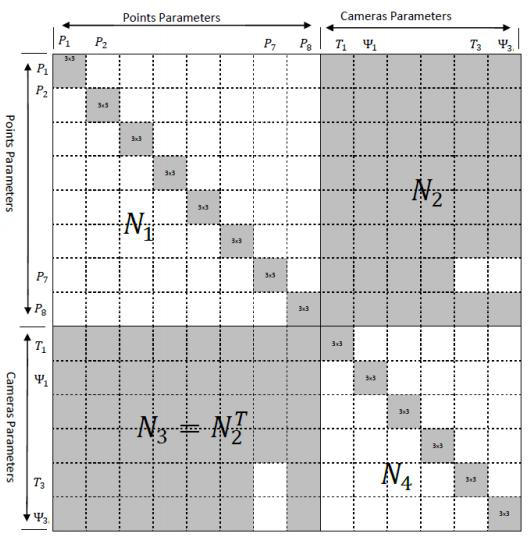}
            \caption{BA Hessian matrix.}
            \label{fig:5.23}
        \end{figure}

    At this time on this basis, then how to solve the whole matrix in the complexity of the calculation it, not only to consider BA algorithm in the weight matrix, then how to calculate it for the points that go wrong? In this way, it is used to calculate This is done by subtracting the calculated points in three dimensions from each other and removing the wrong points, i.e., from Figure \ref{fig:5.22} to Figure \ref{fig:5.23}. The covariance and Jacobi matrices are brought into the right and left sides of the LM algorithm to remove the wrong points. Structure diagram.

    Finally, the result obtained is calculated by the formula:
    \begin{equation}
        \delta x_2 =\left( \mathbf{N_4} - {\mathbf{N_2^TN_1^{-1}N_2}} \right)^{-1} \left( g_2 - \mathbf{N_2^{T}N_1^{-1}g_1}\right)
    \end{equation}

    \subsubsection{Implementation of video-based detection.}
\begin{figure}
            \centering
            \includegraphics[width=0.8\textwidth]{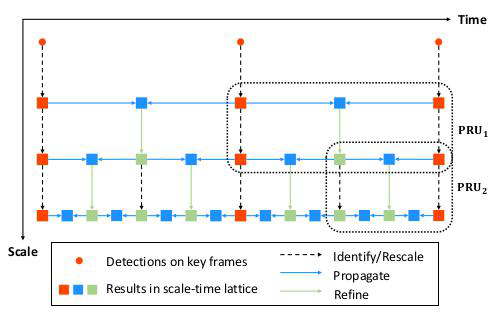}
            \caption{Video detection algorithm.}
            \label{fig:5.24}
        \end{figure}

    This subsection is mainly to introduce how this system is based on video to achieve detection, by designing a good video detection mechanism, the purpose of which is to be fast, i.e., the target position can be captured in real time. As shown in Figure \ref{fig:5.24}, it can be seen that the selection of key frames is video detection is a relatively common means in intelligent video recognition, where the axes are designed for the temporal and spatial dimensions, the horizontal axis is the temporal dimension, i.e., the sequence frames in the video stream, the vertical axis is the spatial dimension, and the Propagation and Refinement Unit (PRU) [35] modules are the modules that are tracked after the accurate results are considered to be obtained.PRU1 and PRU2 are the two modules of the framework where PRU2 is the smallest module, it can be seen that the detection of the whole video is made by obtaining key frames, the whole image which takes a long time, and then by predicting the position between key frames and key frames, and then targeting a The detection of a small area, that is, the operation of optimizing the target position (refine), is also the industry's more mature to This is also the more mature implementation of the tracking algorithm with detection as the model.

    \subsubsection{Overall detection framework for multiple cameras.}

    Since the overall framework of the multi-camera [36] is built on the previous modules, this subsection focuses on the overall multi-camera detection framework and the implementation of each camera target detection in this framework.
    
        \begin{figure}
            \centering
            \includegraphics[width=1.0\textwidth]{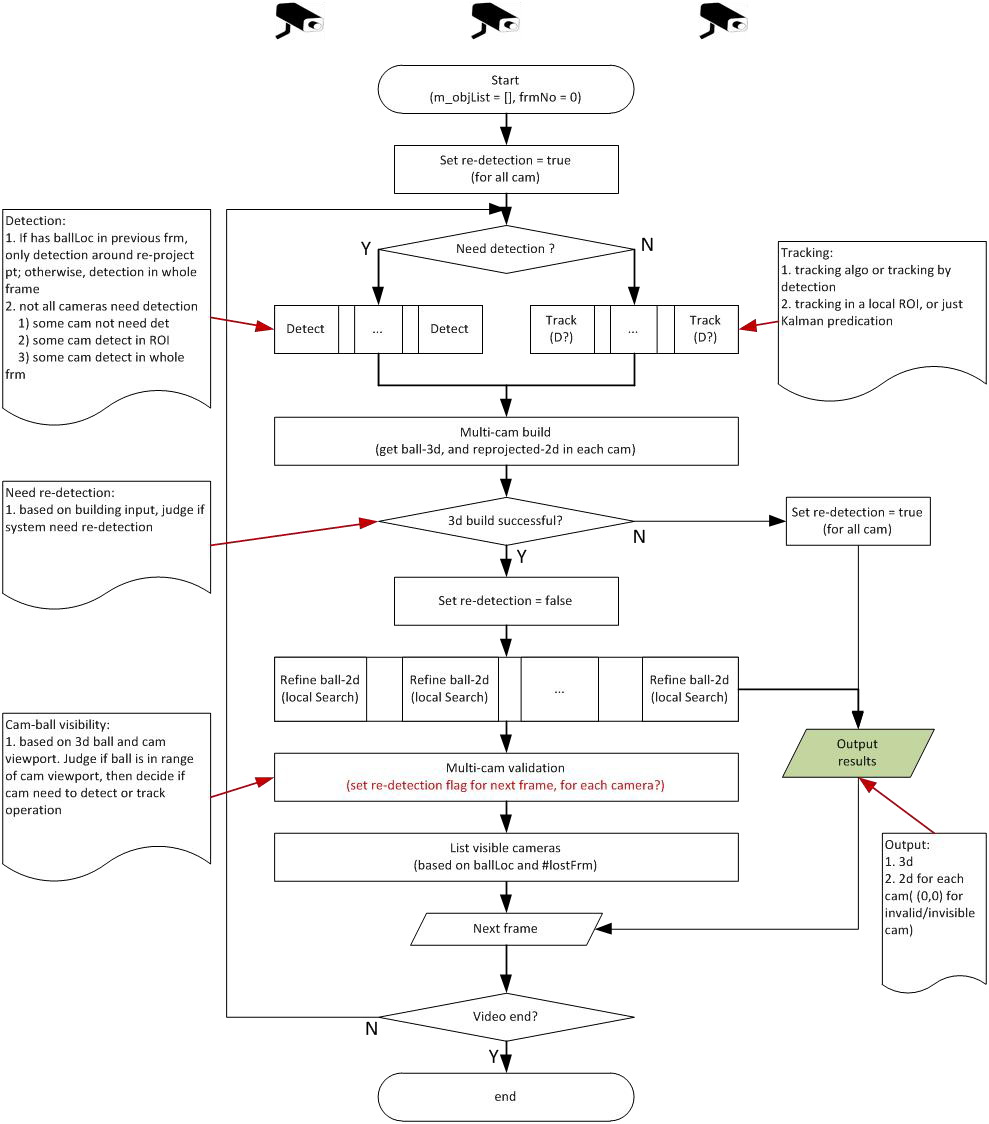}
            \caption{Multi-camera target detection algorithm flow chart.}
            \label{fig:5.25}
        \end{figure}

    As shown in Figure 5.25, the flow chart of multi-camera detection is shown. From the figure, it can be seen that for target detection of images and videos, the images are uploaded to the server, and the background program first finds the ROI in the images, segments the images, and performs end-to-end detection for multiple ROI regions [37], and if there is a ball in the image, the background program marks the location of the ball and returns the target coordinate value.

    From Figure \ref{fig:5.25}, we can know that
    Detection (Detection): Detection in some cameras only, detecting the whole image or ROI.
    Tracking: tracking or tracking by detection (in ROI), or based on the detection Tracking model.
    Validation: Depending on whether re-detection is required.

    The General steps are following:
    \begin{itemize}
        \item Whether the target ball is detected in the field of view with multiple cameras.
        \item The classification of detection by the model, whether multiple balls are detected, is scored, resulting in the highest rated considered is the target.
        \item By tracking the trajectory of the ball and doing smoothing to determine if it is a smooth curve [37], eliminating more of false detections.
        \item Visible camera determination.
            \begin{itemize}
                \item List the visible cameras according to the position of the ball.
                \item If the number of undetected frames lostFrm (the number of frames where the ball was not detected) <= 2, then ROI is given for detection.
                \item If 2 <lostFrm <= 5, then ROI is given for detection.
                \item If 5 <lostFrm, give ROI for detection.
            \end{itemize}    
        \item When the video or image sequence ends, the target location is output in real time.
    \end{itemize}

    \subsection{Summary of this chapter}
    This chapter is the detailed design of the system, which is mainly the detailed development and design of the outline design in the previous chapter. design. Data pre-processing, design and implementation of target detector, single camera detection and multi-camera based This chapter is a detailed description and analysis of the data pre-processing, target detector design and implementation, single camera detection and multi-camera based video detection, and also shows the related effect diagrams, and analyzes and introduces The results are shown and analyzed and introduced.
    
\newpage
%%%%%%%%%%%%%%%%%%%%%%%%%%%%%%%%%%%%%%%%%%%%%%%%%%%%%%%%%%%
\section{Experiments}	

\vspace{10.5cm}
This system is developed and designed to perform relevant tests when each module is ready to run. This system The system is tested for accuracy and speed of each module to see if each module can work effectively. The system is tested for the speed and accuracy of the detector in multiple scenes, the speed and accuracy of each individual camera in 36 channels, and the speed and accuracy of the entire multi-camera system. The main purpose is to update the speed and accuracy of the detector in multiple scenes, the speed and accuracy of each individual camera in 36 channels, and the speed and accuracy of the entire multi-camera system for soccer detection. Iteration is done to make the system work efficiently.

\subsection{System implementation and test environment}

The development and porting of the entire multi-camera soccer detection system was carried out under different platforms. The development and porting of the entire multi-camera soccer detection system was carried out under different platforms, and the system was developed and implemented and tested in the following environment.
Hardware environment is 2.70GHz Intel E5-2680 processor, 64GB memory, GeForce GTX 1080 graphics card, 8GB video memory. The corresponding software environment, operating system, linux and windows can be used.

\subsection{Setting of system test parameters}
Building a live system that can be applied for use in large sports competition conditions usually requires consideration of many factors, among which the work of setting system testing parameters is a design criterion to ensure that the system can operate effectively. As detailed in the previous chapter, some of the parameters used for the developed multi-camera soccer detection system under is an a priori set value, and then through more data testing, more scenes of progressive comparison testing, to The setting of the test parameters can be completed effectively by testing more data and comparing more scenarios. The parameters and description of the system set in this paper are shown in Table \ref{tab:1}.

    \begin{table}
    \centering
    \resizebox{0.9\textwidth}{!}{
        % \begin{tabular}{|l|l|l|}
        \begin{tabular}{|l|l|p{10cm}|}

        \hline Name & Value & Detail \\
        \hline DET\_CONF & $0.9$ & Threshold of detection \\
        \hline IOU\_THD & $0.2 / 1 \mathrm{e}-6$ & Target overlap area threshold \\
        \hline RESIZE & 160 & Image resizing \\
        \hline Visible & $1 / 0$ & Whether the target is visible to the camera, 1 Visible, 0 Not visible \\
        \hline Distance & $50 \mathrm{~cm}$ & Predicted target distance from the real target \\
        \hline Alpha & $0.5$ & Regularization loss factor\\
        \hline Method\_1 & & Every image is detected without tracking \\
        \hline Method\_2 & & The first frame is detected, the threshold value for detection is set, and the detection is waited until it is needed. detection, and the remaining images are tracked \\
        \hline Method\_3 & & First frame detection, set the threshold value for detection, set 30 frames as an interval, and the remaining images are tracked\\
        \hline
        \end{tabular}
    }
        \caption{Description and setting of system-related parameters}
        \label{tab:1}
    
    \end{table}

\subsection{System testing and optimization}
    \subsubsection{System test content overview}
        According to the standard process of software development, this system has been designed modularly, and the whole system development process are elaborated in detail, but whether the whole system can operate effectively requires relevant testing. In this chapter The testing methods used are unit testing and system testing.
        
        The whole testing process can be considered as inputting more competition data, simulating more complex scenes, testing the accuracy and speed of the model, testing the single camera detection unit, testing the multi-camera detection unit to the final The final integration test still tests the speed and accuracy of the whole system, and when the whole testing process reaches the product's index, then the test process will be completed. The final integration test is still to test the speed and accuracy of the whole system, and when the whole test process reaches the product index, the whole system is able to run well.

    \subsubsection{Test evaluation metrics and output definitions}

    The research objectives of this topic require the analysis of relevant detection indicators, Evaluation (evaluation) is a necessary Evaluation is a necessary task, and the evaluation indexes are often the following: Accuracy, Precision, Recall.
    Table 1 shows the explanation of the evaluation criteria, in which the results are analyzed for each camera. Here, the images captured by the main cameras are not the same, mainly the difference between near and far, and the difference between backlight and backlight. The difference is tested for multiple scenes in order to effectively judge the criteria of the detector [38].

    \begin{table}
        \centering
        \begin{tabular}{||c|c|c||}
        \hline
              class     & relevant, positive category  &  Irrelevant, negative category \\
        \hline
              Detected & TP (Correct and awarded as correct) & FP (Incorrectly judged as correct) \\
              Non Detected & FN (incorrectly judged as correct) & TN (incorrectly judged as incorrect) \\
         \hline

        \end{tabular}
        \caption{Evaluation Criteria}
        \label{tab:2}
    \end{table}

    Then, 
        $$
        Precision = \frac{TP}{TP + FP}
        $$

        $$
        Recall = \frac{TP}{TP + FN}
        $$
    and AP (Average Precision), is the area enclosed by the coordinate system of Precision and Recall. The area of The goal of the whole project is to increase the AP value of Single camera and Multiple camera.
    The objective of the whole project is to increase the AP value of Single camera and Multiple camera.

    Among them, the actual scenario, which does not require particularly accurate detection localization, solidly requires the presentation of the target detection and evaluation A parameter in the system, IOU (intersection-over-union, detection evaluation function) [39], can be understood as follows. The intersection of DetectionResult (detection result) and GroundTrurh (accurate set) than their union, i.e.

    $$
    IOU = \frac{DetectionResult \cap GroundTruth}{DetectionResult \cup GroundTruth}
    $$

\begin{figure}
            \centering
            \includegraphics[width=0.4\textwidth]{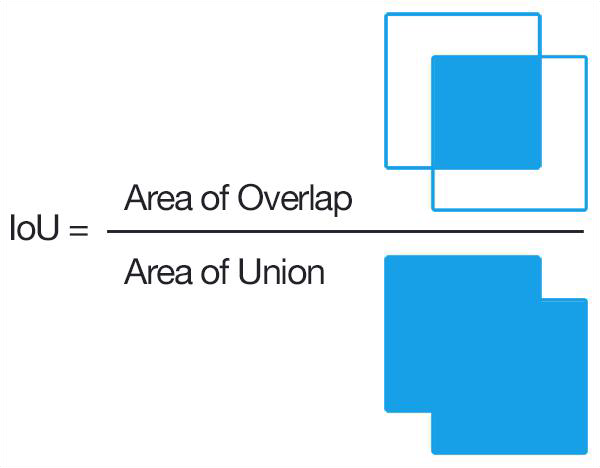}
            \caption{IOU definition.}
            \label{fig:6.1}
        \end{figure}

    Its definition is shown in Figure \ref{fig:6.1}, which is the definition of IOU, from the definition of IOU we can know that the target object The detection accuracy of the target object can be reflected from the IOU or distance, that is, the box of the detected target and the actual box of the target are If there is a small error between the detected frame and the marked frame, then the detection accuracy is accurate. The accuracy of detection is accurate. Meanwhile, the larger the value of IOU will be, the better.

    By setting the above parameters, the ultimate goal of the whole project is to set the AP value of Single camera and AP of Multiple camera under the IOU. The AP value of single camera and multiple camera is increased under the setting of IOU. The AP value of the single camera and the multiple camera will be increased under the IOU setting, and finally the AP value of the target ball will be detected within 50cm of the location. The AP value is 95\%. As shown in Table \ref{tab:3}.

    \begin{table}
        \centering
        
        \begin{tabular}{|c|c|c|c|}
        \hline
              AP     &  IOU>=0.2  &  IOU>=0.6 & Dist=50cm \\
        \hline
              Single camera & 0.7 & 0.75 & 0.85 \\
              Multiple cameras & 0.85 & 0.9 & 0.98 \\
         \hline
        \end{tabular}
        
        \caption{Target of Research}
        \label{tab:3}
    \end{table}

% \begin{table}[h!]

% \begin{tabular}{|c|l|c|c|}
% \hline Class & Type & Detail & Note \\
% \hline 
% \multirow{3}{*}{3D} 
% & (frmNo, $\mathrm{x}, \mathrm{y}, \mathrm{z})$ & frame\_no,3d position & In the world coordinate system \\
% & Normal value & $\mathrm{X}:[-325,325] , \mathrm{Y}:[0,200] , \mathrm{Z}:[-510,510]$& \\
% & (frmNo,-1000, $-1000,-1000)$ & Failure of 3d ball loc & \\
% \hline

% \hline \multirow{3}{*}{ 2D } & (frmNo, $\mathrm{x},  \mathrm{y}, \mathrm{z}, \mathrm{w}, \mathrm{h}, \mathrm{conf})$ & frame_no,3d position &  \\
% \cline { 2-5 } & Normal value & $\mathrm{X}:[0,imgWidth] , \mathrm{Y}:[0,imgHeight] , \mathrm{w,H}: ~25/scaleFactor, conf:[0,1]$ & In the 5k image, the The target diameter size is about is 25pixel \\
% \cline {2-5} & (frmNo, 0, 0, 0, 0, -1) & 3D backcalculation failed, no 2D output & & \\
% \cline{2-5} & (frmNo, x, y, w, h, conf=0)& 3D succeeds, but 2D detection fails & Commonly found in obscured
% situations \\
% \hline
% \end{tabular}
% \caption{Description and setting of system-related parameters}
% \label{tab:1}
% \end{table}

\begin{figure}
            \centering
            \includegraphics[width=1.0\textwidth]{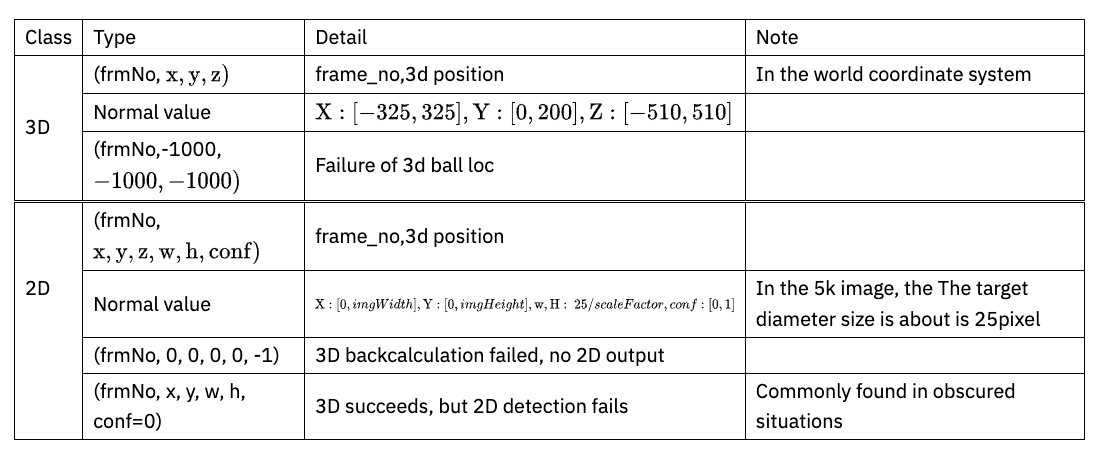}
            \caption{Target ball output result format.}
            \label{table:6.2}
        \end{figure}

The above indicators are calculated on the basis of the output, and Table \ref{table:6.2} shows the definition of the output data of the ball test. definition.

\subsubsection{Comparison test of related detection algorithms}
In this paper, before designing the system, the baseline test work is conducted on the popular and effective algorithms nowadays, here the system is mainly based on ssd algorithm, fast-rcnn algorithm and YOLOv2 algorithm.
The adaptive testing of the system is based on ssd, fast-rcnn and YOLOv2 algorithms. In order to determine the algorithmic framework of the whole system, the same training data set and the same test data set were used to evaluate the three algorithms, where the size of the image fed to the detector is 320*320. The algorithm framework of YOLOv2 is optimized and improved based on it. In which, the table ballNum indicates the number of detected balls, gtNum indicates the data with balls in groundtruth, and hitNum indicates the number of detected balls in gtNum of hits.

\begin{table}[h!]
\centering

\begin{tabular}{|l|l|l|l|l|l|}

\hline methods & & & & iou $=0.2$ & \\

\hline & ballNum & gtNum & hitNum & precision & recall \\
\hline YOLOv2 & 878 & 924 & 810 & $0.86$ & $0.92$ \\
\hline ssd & 858 & 924 & 804 & $0.84$ & $0.91$ \\
\hline Faster-rcnn & 899 & 924 & 756 & $0.81$ & $0.84$ \\
\hline
\end{tabular}
    \caption{Algorithm accuracy evaluation}
    \label{tab:6.4}
\end{table}

In Table \ref{tab:6.5}, the time taken for a graph and a sequence (here, 450 frames is a sequence) is evaluated. time evaluation.
\begin{table}
\centering

\begin{tabular}{|c|l|l|l|}

\hline methods & YOLOv2 & ssd & Faster-rcnn \\
\hline Time-1-frame &  $200 \mathrm{~ms}$ & $210 \mathrm{~ms}$ &  $230 \mathrm{~ms}$ \\
\hline Time-1-squence &  $160000 \mathrm{~ms}$ & $160150 \mathrm{~ms}$ &  $161000 \mathrm{~ms}$ \\
\hline
\end{tabular}
    \caption{Algorithm speed evaluation}
    \label{tab:6.5}
\end{table}

Therefore, through this subsection, the system uses the YOLOv2-based algorithm framework to build the soccer ball detector. detector. On the other hand, we can see that if the system only uses the detector for detection, i.e., it detects every frame, and sends many partial images to the detector each time. If the system only uses the detector, i.e., detects each frame, many partial images will be sent to the detector each time, in this case, the whole time cannot reach Therefore, it is necessary to test the time consumption of detection and tracking in the whole video detection framework.

\subsubsection{Model accuracy and speed testing}
The tests in this section must be combined with filtering algorithms, and the goodness of the Pre-filter Methods (filtering algorithms) [40] affects the goodness of the detector. The testing and evaluation of the Pre-filter Methods are shown in Table \ref{tab:6.5}. and evaluation are shown in Table \ref{tab:6.5}. Among them, after the introduction of the filtering algorithm, then the accuracy of the detector predicts the overall effectiveness of the whole The detector size is constantly compressed in the ball detection, which also effectively reduces the time. The time is reduced.

As shown in Table \ref{tab:6.5}, the overall effect is closely related to the filtering algorithm, and the good or bad filtering algorithm directly affects the speed and accuracy of the whole system, and the diversity of scenarios also makes an effective and good speed The diversity of scenes also makes an effective and speedy filtering algorithm the basis for the operation of the whole system.

%%% TODO to change to the latex template
% \begin{table}
%     \centering
%     % \resizebox{0.9\textwidth}{!}{
% \begin{tabular}{l|p{2cm}|l|l|l|l}
%     \hline Strategy & details & strength & weakness & note &\\
%     \hline $$fg = frm(t)-bg$$ & The current frame difference is directly minus the background image & simple and fast & Since bg and frm have deviates from the frm, resulting in the subtraction of line, there will be two lines, increasing the number of The number of tested tiles & & \\
%     \hline $$fg = frm(t)-frm(t-1)$$ & Subtract the current frame from the previous frame & Simple, fast, and can remove the pitch line impact & May make the ball lost lost, if the ball does not move & & \\
%     \hline  $$fg =frm(t)-frm(t-1)+tile_in_frm(t-1)$$  & The current frame is subtracted from the previous frame and the previous frame to get the result & Simple, fast and can to remove pitch lines, etc. and can maintain a higher accuracy rate & Dependent on the previous frame Results & If chronological detection/tracking then , there is no problem & \\
%     \hline $$fg = frm(t)- frm_line(t)$$ & Current frame minus out Image of the sidefield line & & & Align the fieldline marked on bg fieldline, adjust the to the frm, then subtract, requiring Fast corresponding &
% \end{tabular}

%     \caption{Caption}
%     \label{tab:6.5}
% \end{table}

        \begin{figure*}
            \centering
            \includegraphics[width=1.0\textwidth]{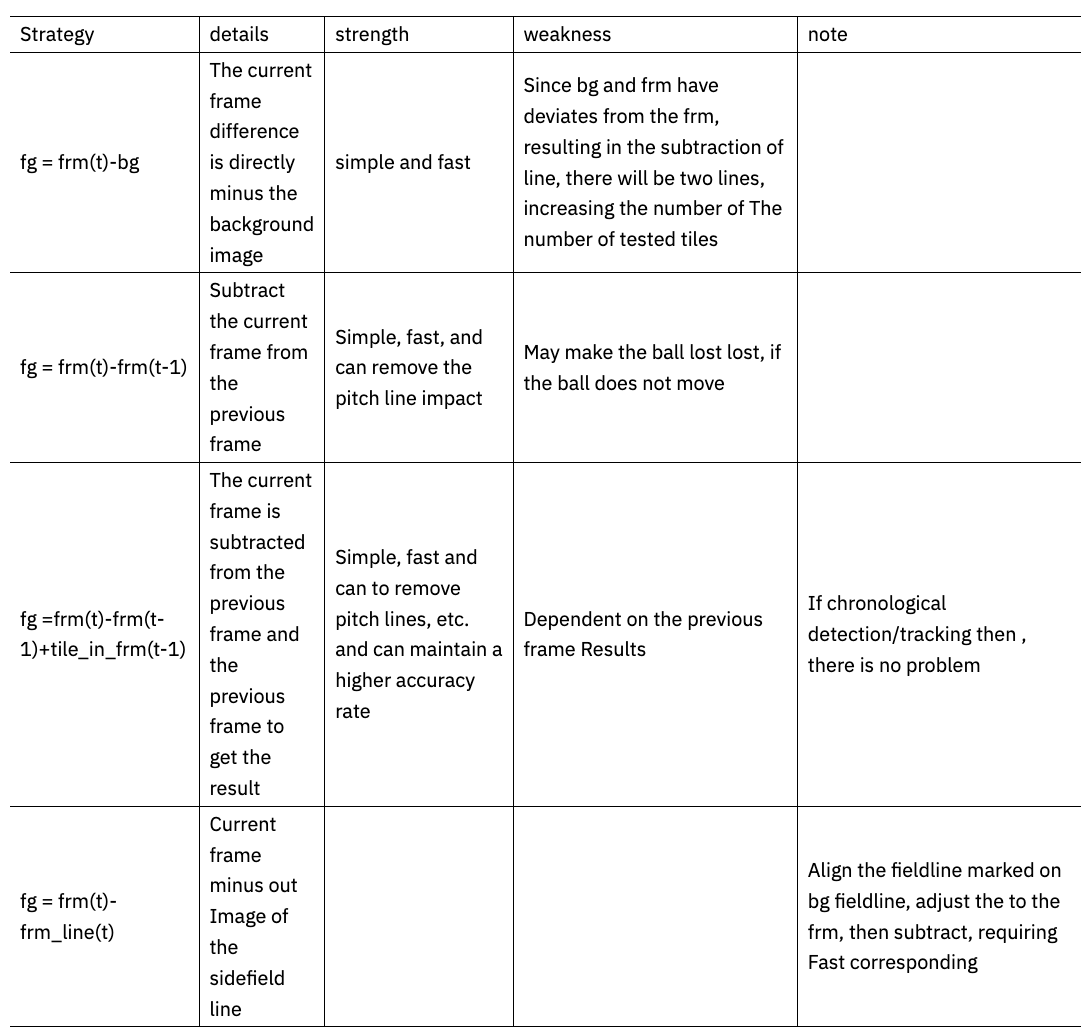}
            \caption{Pre-filter methods evaluation.}
            \label{table:6.5}
        \end{figure*}

\begin{figure*}[ht!]
            \centering
            \includegraphics[width=0.8\textwidth]{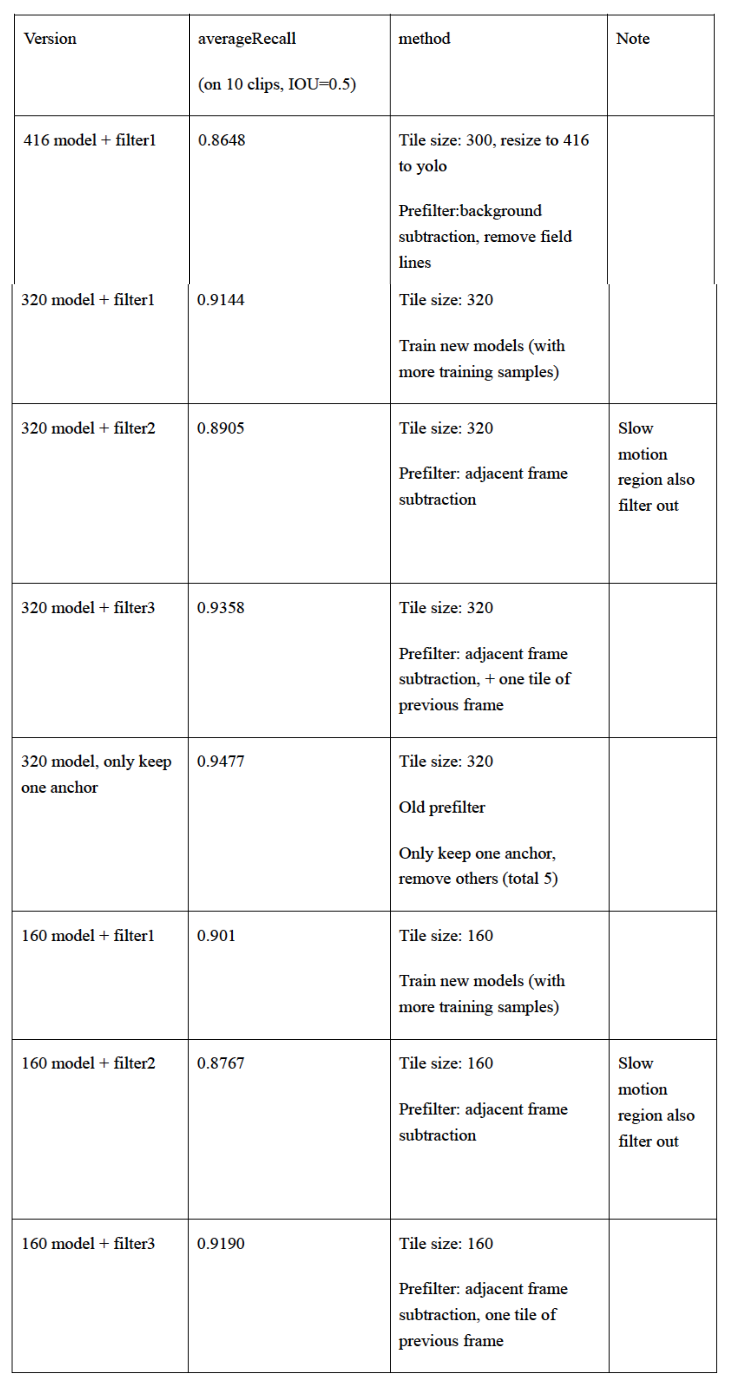}
            \caption{Evaluation of the value of recall for target ball detection.}
            \label{table:6.6}
        \end{figure*}

Table \ref{table:6.6} shows the value of recall for target ball detection, which is also an important criterion to determine the accuracy of ball detection. The index of recall determines whether the soccer is lost in the live broadcast, so it can be seen that recall is important for the overall system performance. performance is very important

Table \ref{table:6.6} shows that the 160 model does not lose much accuracy and can meet the basic requirements for subsequent multi-camera co-detection, so the next step is to test how well it performs in terms of speed. The overall optimization of the model is a direct compression in the model size, from 416model to 320model to 160model, for targeted velocity and measurement accuracy, gradually improve the process. The optimization of the speed aspect of the ball detection model shows that 160model finally accelerates about ten times, and the superior performance of 160model for soccer ball detection through the detailed design in the previous chapter can be very well adapted to this system.

        \begin{figure*}
            \centering
            \includegraphics[width=0.8\textwidth]{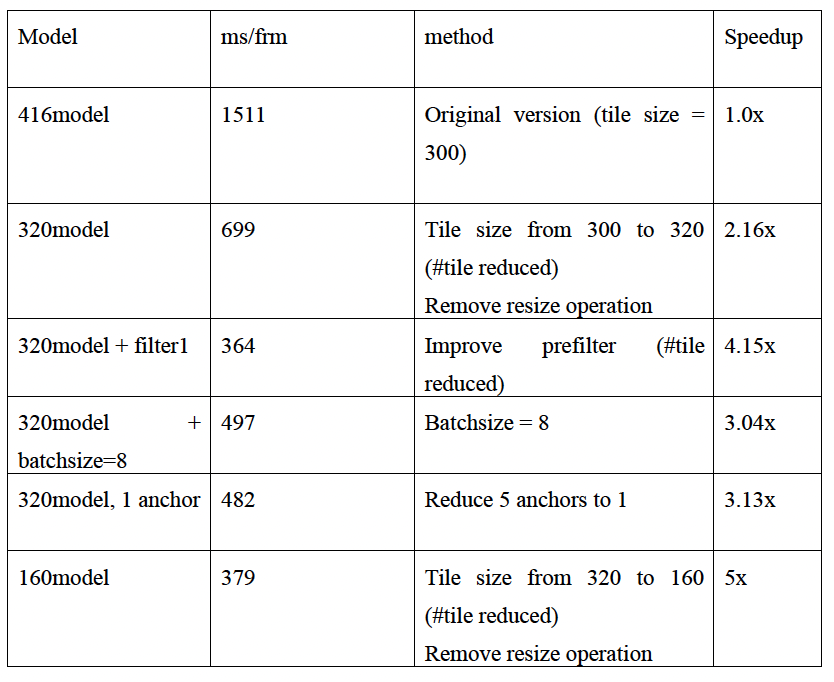}
            \caption{Optimization of the ball detection model in terms of speed review.}
            \label{table:6.7}
        \end{figure*}

\subsubsection{Test of single camera detection}
\begin{figure*}[ht!]
            \centering
            \includegraphics[width=1.0\textwidth]{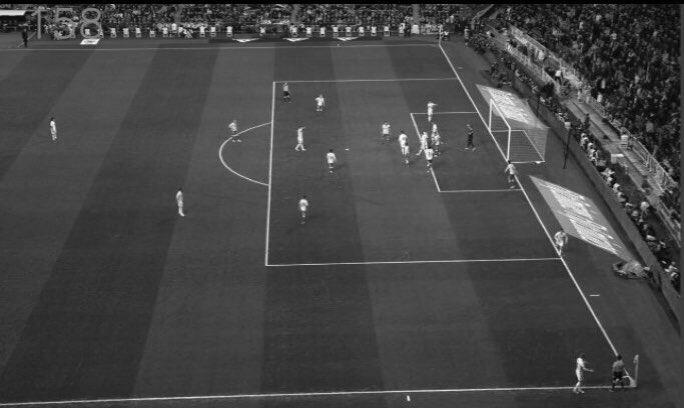}
            \caption{Single camera test results.}
            \label{fig:6.3}
        \end{figure*}

        From the unit tests in the system, it is clear that this subsection focuses on the performance aspects of single camera detection. The values ofprecision and recall are mainly investigated here. First, for the visualization of camera 08 As shown in Figure 6.3, the top left corner of the figure shows frame 289 of the camera, and the trajectory with the tail in the figure is the ball The effect of target detection is shown in Figure \ref{fig:6.3}.

        The detection of single camera needs to be judged by multiple cameras with different angles and different scenes. From \ref{tab:6.8} to \ref{tab:6.13} are the detection results of single camera, and the detection results also reflect the performance detection of the whole single camera. In this case, each camera has 450 frames per clip, and the values of precision, recall, iou, and dist=50 are pointed out in the previous evaluation index.

        \begin{table}
            \centering
                \begin{tabular}{|l|l|l|l|l|l|}
                \hline camera & & & & & iuu $=0.2$ \\
                \hline 3 & ballNum & gtNum & hitNum & precision & recall \\
                \hline FGC3 & 220 & 295 & 187 & $0.85$ & $0.6339$ \\
                \hline FGC5 & 154 & 155 & 154 & 1 & $0.99355$ \\
                \hline FGC7 & 157 & 154 & 150 & $0.95541$ & $0.97403$ \\
                \hline FGC12 & 130 & 131 & 129 & $0.99231$ & $0.98473$ \\
                \hline FGC14 & 240 & 198 & 193 & $0.80417$ & $0.97475$ \\
                \hline FGC16 & 151 & 151 & 148 & $0.98013$ & $0.98013$ \\
                \hline FGC21 & 300 & 277 & 247 & $0.82333$ & $0.8917$ \\
                \hline FGC23 & 109 & 76 & 75 & $0.68807$ & $0.98684$ \\
                \hline FGC25 & 135 & 128 & 124 & $0.91852$ & $0.96875$ \\
                \hline FGC30 & 202 & 168 & 157 & $0.77723$ & $0.93452$ \\
                \hline FGC32 & 135 & 135 & 135 & 1 & 1 \\
                \hline FGC34 & 303 & 347 & 285 & $0.94059$ & $0.82133$ \\
                \hline
                \end{tabular}
            \caption{Detection results of single camera 3 (iou=0.2)}
            \label{tab:6.8}
        \end{table}

        \begin{table}
            \centering        
        \begin{tabular}{|l|l|l|l|l|l|}
        \hline camera & & & & & iou=1e-6 \\
        \hline 3 & ballNum & gtNum & hitNum & precision & recall \\
        \hline FGC3 & 220 & 295 & 201 & $0.91364$ & $0.68136$ \\
        \hline FGC5 & 154 & 155 & 154 & 1 & $0.99355$ \\
        \hline FGC7 & 157 & 154 & 153 & $0.97452$ & $0.99351$ \\
        \hline FGC12 & 130 & 131 & 129 & $0.99231$ & $0.98473$ \\
        \hline FGC14 & 240 & 198 & 196 & $0.81667$ & $0.9899$ \\
        \hline FGC16 & 151 & 151 & 151 & 1 & 1 \\
        \hline FGC21 & 300 & 277 & 272 & $0.90667$ & $0.98195$ \\
        \hline FGC23 & 109 & 76 & 76 & $0.69725$ & 1 \\
        \hline FGC25 & 135 & 128 & 128 & $0.94815$ & 1 \\
        \hline FGC30 & 202 & 168 & 164 & $0.81188$ & $0.97619$ \\
        \hline FGC32 & 135 & 135 & 135 & 1 & 1 \\
        \hline FGC34 & 303 & 347 & 297 & $0.9802$ & $0.85591$ \\
        \hline
        \end{tabular}
            \caption{Detection results of single camera 3 (iou=1e-6)}
            \label{tab:6.9}
        \end{table}

        \begin{table}
            \centering 
        \begin{tabular}{|l|l|l|l|l|l|}
        \hline camera & & & & & dist $=50$ \\
        \hline 3 & ballNum & gtNum & hitNum & precision & recall \\
        \hline FGC3 & 220 & 295 & 220 & 1 & $0.74576$ \\
        \hline FGC5 & 154 & 155 & 154 & 1 & $0.99355$ \\
        \hline FGC7 & 157 & 154 & 153 & $0.97452$ & $0.99351$ \\
        \hline FGC12 & 130 & 131 & 130 & 1 & $0.99237$ \\
        \hline FGC14 & 240 & 198 & 196 & $0.81667$ & $0.9899$ \\
        \hline FGC16 & 151 & 151 & 151 & 1 & 1 \\
        \hline FGC21 & 300 & 277 & 276 & $0.92$ & $0.99639$ \\
        \hline FGC23 & 109 & 76 & 76 & $0.69725$ & 1 \\
        \hline FGC25 & 135 & 128 & 128 & $0.94815$ & 1 \\
        \hline FGC30 & 202 & 168 & 166 & $0.82178$ & $0.9881$ \\
        \hline FGC32 & 135 & 135 & 135 & 1 & 1 \\
        \hline FGC34 & 303 & 347 & 302 & $0.9967$ & $0.87032$ \\
        \hline
        \end{tabular}
            \caption{Detection results of single camera 3 (dist=50)}
            \label{tab:6.10}
        \end{table}

        \begin{table}
            \centering 
        \begin{tabular}{|l|l|l|l|l|l|}
        \hline camera & & & & & \\
        \hline 8 & ballNum & gtNum & hitNum & precision & recall \\
        \hline FGC3 & 295 & 306 & 291 & $0.98644$ & $0.95098$ \\
        \hline FGC5 & 60 & 64 & 61 & $1.01667$ & $0.95313$ \\
        \hline FGC7 & 159 & 155 & 137 & $0.86164$ & $0.88387$ \\
        \hline FGC12 & 32 & 35 & 32 & 1 & $0.91429$ \\
        \hline FGC14 & 226 & 227 & 211 & $0.93363$ & $0.92952$ \\
        \hline FGC16 & 145 & 151 & 144 & $0.9931$ & $0.95364$ \\
        \hline FGC21 & 253 & 259 & 197 & $0.77866$ & $0.76062$ \\
        \hline FGC23 & 257 & 256 & 250 & $0.97276$ & $0.97656$ \\
        \hline FGC30 & 198 & 198 & 188 & $0.94949$ & $0.94949$ \\
        \hline FGC32 & 128 & 130 & 113 & $0.88281$ & $0.86923$ \\
        \hline FGC34 & 260 & 269 & 250 & $0.96154$ & $0.92937$ \\
        \hline
        \end{tabular}
            \caption{Detection results of single camera 8 (iou=0.2)}
            \label{tab:6.11}
        \end{table}

        \begin{table}
            \centering 
        \begin{tabular}{|l|l|l|l|l|l|}
        \hline camera & & & & & iou=1e-6 \\
        \hline 8 & ballNum & gtNum & hitNum & precision & recall \\
        \hline FGC3 & 295 & 306 & 293 & $0.99322$ & $0.95752$ \\
        \hline FGC5 & 60 & 64 & 61 & $1.01667$ & $0.95313$ \\
        \hline FGC7 & 159 & 155 & 150 & $0.9434$ & $0.96774$ \\
        \hline FGC12 & 32 & 35 & 32 & 1 & $0.91429$ \\
        \hline FGC14 & 226 & 227 & 219 & $0.96903$ & $0.96476$ \\
        \hline FGC16 & 145 & 151 & 145 & 1 & $0.96026$ \\
        \hline FGC21 & 253 & 259 & 234 & $0.9249$ & $0.90347$ \\
        \hline FGC23 & 257 & 256 & 253 & $0.98444$ & $0.98828$ \\
        \hline FGC30 & 198 & 198 & 191 & $0.96465$ & $0.96465$ \\
        \hline FGC32 & 128 & 130 & 125 & $0.97656$ & $0.96154$ \\
        \hline FGC34 & 260 & 269 & 253 & $0.97308$ & $0.94052$ \\
        \hline
        \end{tabular}
            \caption{Detection results of single camera 8 (iou=1e-6)}
            \label{tab:6.12}
        \end{table}

        \begin{table}
            \centering 
        \begin{tabular}{|l|l|l|l|l|l|}
        \hline camera & & & & & \\
        \hline 8 & ballNum & gtNum & hitNum & precision & recall \\
        \hline FGC3 & 295 & 306 & 294 & $0.99661$ & $0.96078$ \\
        \hline FGC5 & 60 & 64 & 61 & $1.01667$ & $0.95313$ \\
        \hline FGC7 & 159 & 155 & 153 & $0.96226$ & $0.9871$ \\
        \hline FGC12 & 32 & 35 & 32 & 1 & $0.91429$ \\
        \hline FGC14 & 226 & 227 & 221 & $0.97788$ & $0.97357$ \\
        \hline FGC16 & 145 & 151 & 145 & 1 & $0.96026$ \\
        \hline FGC21 & 253 & 259 & 251 & $0.99209$ & $0.96911$ \\
        \hline FGC23 & 257 & 256 & 254 & $0.98833$ & $0.99219$ \\
        \hline FGC30 & 198 & 198 & 194 & $0.9798$ & $0.9798$ \\
        \hline FGC32 & 128 & 130 & 128 & 1 & $0.98462$ \\
        \hline FGC34 & 260 & 269 & 256 & $0.98462$ & $0.95167$ \\
        \hline
        \end{tabular}
            \caption{Detection results of single camera 8 (dist=50)}
            \label{tab:6.13}
        \end{table}

        From the results of multiple single-camera tests, information on the results of 36 cameras in each game needs to be counted so that the results can be targeted. This way the relevant results can be given in a targeted manner, and since scenarios and venues can vary, it is necessary to test in multiple test Since the scenarios and venues vary, it is necessary to conduct tests in multiple test sets to effectively consider that the single camera module can meet the requirements of the system, so that the entire system framework can be built and developed with a view to having a better understanding of the system. Only when each sub-module is efficient and useful can the system be built and developed without considering other sub-modules. Only when each sub-module is efficient and useful can other information not be considered when building the module.

    For the results of the above 36 cameras, the performance index of single camera in this game is shown in Table 6.14. It can be found that, except for most of the occlusion problems, in which the false detection and other situations can be effectively solved, the index of single camera can basically be achieved, and the index of single camera can also achieve a better accuracy and recall rate, then for multi-camera, on this basis, it will certainly be greatly improved

%% TODO to modify
        \begin{table}
            \centering 
        \begin{tabular}{|l|l|l|l|l|}
        \hline & camera12 & iou $=0.2$ & iou $=1 \mathrm{e}-6$ & $50 \mathrm{~cm}$ \\
        \hline \multirow{5}{*}{ Precision } & live3 & $0.887298748$ & $0.919499106$ & $0.933363148$ \\
        
        \hline & live7 & $0.885857033$ & $0.993082244$ & $0.995388163$ \\
        % \cline { 2 - 7 } & live8 & $0.930948833$ & $0.971684054$ & $0.988077496$ \\
        \hline { 2 - 7 } & live8 & $0.930948833$ & $0.971684054$ & $0.988077496$ \\

        \hline & live11 & $0.947676852$ & $0.973629134$ & $0.982838008$ \\
        \hline & live12 & $0.894137495$ & $0.913116828$ & $0.943061999$ \\
        \hline & average & $0.911675563$ & $0.956762282$ & $0.971129985$ \\

        \hline \multirow{5}{*}{ Recall } & live3 & $0.895711061$ & $0.928216704$ & $0.94221219$ \\
        \hline & live7 & $0.852440828$ & $0.955621302$ & $0.957840237$ \\
        \hline & live8 & $0.914146341$ & $0.954146341$ & $0.970243902$ \\
        \hline & live11 & $0.947676852$ & $0.973629134$ & $0.982838008$ \\
        \hline & live12 & $0.91934085$ & $0.93885516$ & $0.969644406$ \\
        \hline & average & $0.905863187$ & $0.950093728$ & $0.964555748$ \\
        \hline
        \end{tabular}
            \caption{Detection results of single camera 8 (dist=50)}
            \label{tab:6.14}
        \end{table}

        From the appeal statistics, it can be seen that the entire single camera product is very ineffective, because some cameras Due to the large number of occlusions, a large number of multiple cameras need to be tested in concert, because the synergy of multiple cameras, the increase in recall and accuracy due to error detection can be well solved some occlusions. The improvement of recall and accuracy due to some occlusions can be solved by the synergy of multiple cameras.

    \subsubsection{Test for multi-camera detection}
    The multi-camera test is mainly aimed at the whole system [41], in which, firstly, the single camera and camera parameters to reconstruct the motion capture of the soccer ball in 3D, due to the false and missed detections of the single camera, from which can lead to errors in the modeling process of multiple cameras. Also for the data pre-processed graph, the scene graph under multiple cameras is shown in Figure 6.4.
        \begin{figure*}
            \centering
            \includegraphics[width=0.8\textwidth]{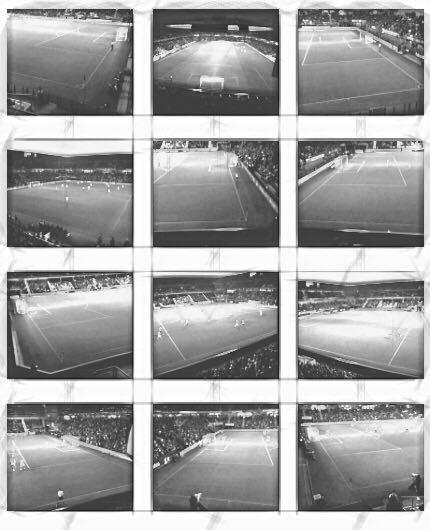}
            \caption{Course map with multiple cameras.}
            \label{fig:6.4}
        \end{figure*}

    The pitch map after the filtering algorithm is shown in Figure 6.5. The test shows that a large portion of the redundant information has been filtered out. Figure 6.5 shows the hind-view cut map of one frame in the stadium game video. It can be seen from the figure that after eliminating the redundant information from the rest of the stadium, the only information in the stadium is the player and the target ball, so that after setting the relevant simple pixel threshold, the possible target region of interest will be framed into the detector operation, which is the operation to determine the target location.

        \begin{figure*}
            \centering
            \includegraphics[width=0.8\textwidth]{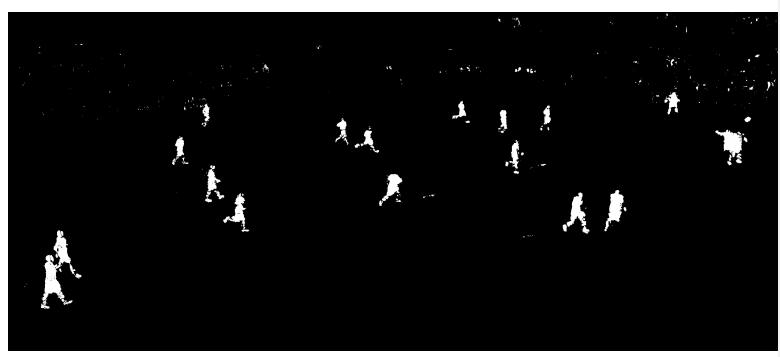}
            \caption{Rear view of the course after cutting.}
            \label{fig:6.5}
        \end{figure*}

\begin{figure}
            \centering
            \includegraphics[width=0.8\textwidth]{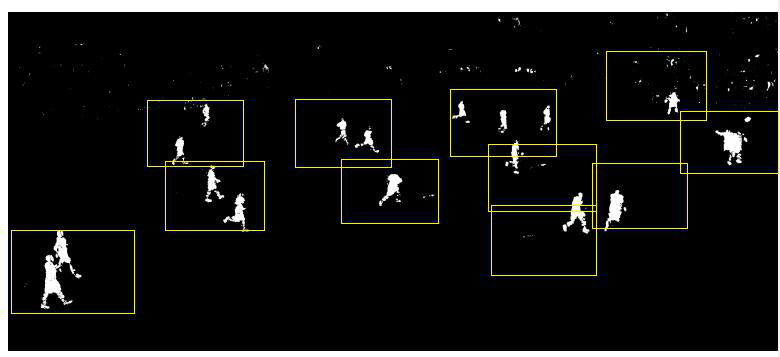}
            \caption{Block diagram of the rear view cut after the pitch is fed to the detector.}
            \label{fig:6.6}
        \end{figure}

        As shown in Figure \ref{fig:6.6}, the positions of the possible balls are fed into the YOLO detector, i.e., the yellow box in the figure Then after getting their positions determined, they are back-calculated into the original graph, i.e., the possible positions in the original graph must be fed into the detector. Because in the training, it is RGB graph, that is, the information of three channels, in the test also trained the model accuracy of a single channel, although the overall speed is improved, but because of the single channel singularity feature, there is no way to extract the information of the soccer, so after feeding all the boxes into the detector, the weighted to calculate the target prediction value, each box is calculated as a prediction value for threshold selection of the final value.

        Then how to obtain the accurate true value of each camera position in 3D by multiple single cameras? Through the previous section, the calibration information of the camera is returned by the camera modeler, that is, the position information of the camera in 3D is known, and the linear transformation is performed directly, where the spatial point $P$ of the target soccer in 3D is required, and its odd order coordinates are $\mathrm{P}=(\mathrm{X}, \mathrm{Y}, \mathrm{Z}, 1)^{\mathrm{T}}$, where the position matrix information of the camera is $\mathrm{K}$, then after obtaining the position of the target from a single camera, that is, if its coordinates under camera 1 are $\mathrm{x} 1=\left(\mathrm{u}_1, \mathrm{v}_1, 1\right)^{\mathrm{T} }$, then we get in three dimensions as:
        $$
        P=x_1 \bullet K
        $$
        At this point, the results are optimized by the cooperative detection of multiple cameras introduced in the previous section, which is visualized in the demo video screenshots shown in Figs. $6.7$ and $6.8$. It is important to note that this figure is for the difficult case of a single camera, where the scenario is due to the change in light intensity, i.e., the playing field is due to the sunlight coming down from the building above, and the whole playing field keeps changing due to the changing sunlight scene over time. This is well resolved in the final modeling.

        It can be seen from the figure that the single camera is not very accurate after acquiring the exact position of the ball, and after collaborating with multiple cameras by After collaborating with multiple cameras, i.e., as seen in Figure \ref{fig:6.8} and Figure \ref{fig:6.9}, multiple cameras perform the calculation of the optimal solution, i.e., the single camera The result of the single camera is corrected and updated, so it is the most accurate, and it is the guarantee that the whole system can operate effectively.

         \begin{figure}
            \centering
            \includegraphics[width=0.8\textwidth]{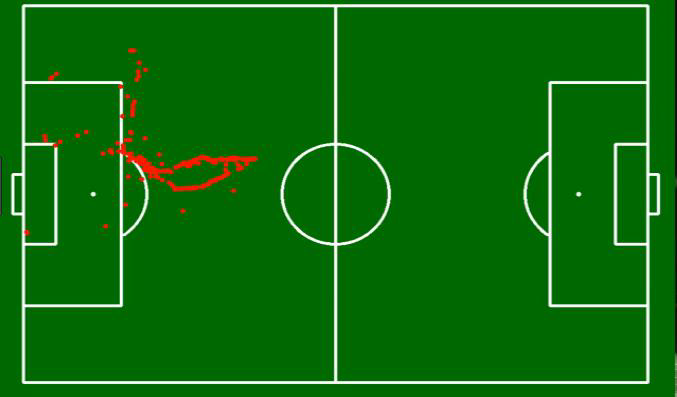}
            \caption{Single camera modeling 3D target map.}
            \label{fig:6.7}
        \end{figure}

        \begin{figure}
            \centering
            \includegraphics[width=0.8\textwidth]{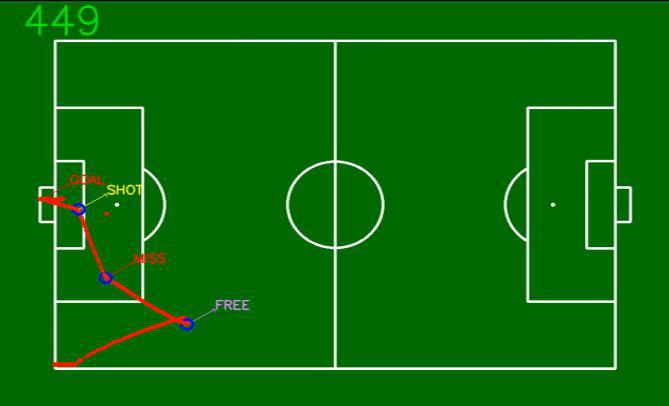}
            \caption{Multi-camera modeling of 3D target maps.}
            \label{fig:6.8}
        \end{figure}

        \begin{figure}
            \centering
            \includegraphics[width=0.8\textwidth]{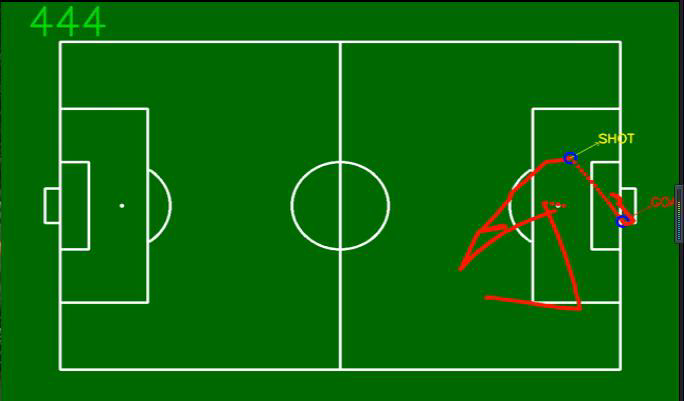}
            \caption{Multi-camera modeling of 3D target maps.}
            \label{fig:6.9}
        \end{figure}

        The strategies in the parameters are then tested one by one to see if the developed strategies are very effective in solving the problem of accuracy and speed of the system.

        \begin{itemize}
            \item Testing three scenarios in a multi-camera frame process:
                \begin{itemize}
                    \item M1: Every 5 frames are detected and the remaining frames are not processed, or if the keyframe detection fails, they are not processed.
                    \item M2: Detect every 5 frames, if there is a result, the subsequent frames are tracked, if the track fails, it is not processed; if the detection fails, it is also not processed.
                    \item M3: The first frame is detected, the subsequent trace, and if the trace fails, the frame is re-detected and not processed after the failed detection.
                \end{itemize}
            \item Based on the result of detect (track) of the above three solutions, 3D is calculated for each frame and then returned for optimization, and then local detection is performed as the final output.
            \item Linear interpolation of M1 and M2 outputs, followed by comparison with GT to calculate precision and recall. The output of M3 is not processed, and the comparison with GT is done directly to calculate the precision and recall.
        \end{itemize}

        According to the performance metrics of the whole system, the results of precision of multiple cameras under multiple strategies, the results of recall of multiple cameras under multiple strategies are pretty good.  The results of recall for multiple cameras under multiple strategies also go to a higher accuracy.
        As the test standard, the accuracy of 97.5\% can be achieved on all other lives.

        %% TODO add the multi-cam test
        \begin{table}
            \centering 
        \begin{tabular}{|l|l|l|l|l|l|}
        \hline precision & & & & & \\
        \hline & cam12 & method & iou $=0.2$ & iou=1e-6 & $50 \mathrm{~cm}$ \\
        \hline & live03 & 1 & $0.707789919$ & $0.744800846$ & $0.764892492$ \\
        \hline & & 2 & $0.730654762$ & $0.862417582$ & $0.772321429$ \\
        \hline & & 3 & $0.931653076$ & $0.947114482$ & $0.947114482$ \\
        \hline & live07 & 1 & $0.804702313$ & $0.858551384$ & $0.93629124$ \\
        \hline & & 1 & $0.897338403$ & $0.924799324$ & $0.972539079$ \\
        \hline & & 2 & $0.890527092$ & $0.924069296$ & $0.990416513$ \\
        \hline & & 3 & $0.986305493$ & $0.989769752$ & $0.989769752$ \\
        \hline & live08  & 1 & $0.78116939$ & $0.81900258$ & $0.84866724$ \\
        \hline & & 2 & $0.845274725$ & $0.862417582$ & $0.876043956$ \\
        \hline & & 3 & $0.876043956$ & $0.977114482$ & $0.977114482$ \\
        \hline & live11 & 1 & $0.897338403$ & $0.924799324$ & $0.972539079$ \\
        \hline & & 2 & $0.954013378$ & $0.969063545$ & $0.987458194$ \\
        \hline & & 3 & $0.988182198$ & $0.994897618$ & $0.994897618$ \\
        % \hline & live12 & 1 & $0.796603476$ & $0.834518167$ & $0.86492891$ \\
        \hline & livel2 & 1 & $0.796603476$ & $0.834518167$ & $0.86492891$ \\
        \hline & & 2 & $0.817404817$ & $0.842657343$ & $0.858974359$ \\
        \hline & & 3 & $0.891137566$ & $0.935277778$ & $0.955277778$ \\
        \hline & average & 1 & $0.7975207$ & $0.83633446$ & $0.877463792$ \\
        \hline & & 2 & $0.847574955$ & $0.89212507$ & $0.89704289$ \\
        \hline & & 3 & $0.969556064$ & $0.971634822$ & $0.98245216$ \\
        
        % \hline & & 3 & $0.858974359$ & $0.935277778$ & $0.955277778$ \\
        % \hline & average & 1 & $0.866621589$ & $0.908453171$ & $0.952706965$ \\
        % \hline & & 2 & $0.897984709$ & $0.922630628$ & $0.969031204$ \\
        % \hline & & 3 & $0.947361321$ & $0.954219004$ & $0.954219004$ \\
        % \hline & & 3 & $$ & $$ & $$ \\

        \hline
        \end{tabular}
            \caption{The result of precision of multiple cameras under multiple policies}
        \label{tab:6.15}
        \end{table}

        \begin{table}
            \centering 
        \begin{tabular}{|l|l|l|l|l|l|}
        \hline recall & & & & & \\
        \hline & cam12 & method & iou $=0.2$ & iou $=1 \mathrm{e}-6$ & $50 \mathrm{~cm}$ \\
        \hline & live03 & 1 & $0.906546275$ & $0.948387097$ & $0.979683973$ \\
        \hline & & 2 & $0.886681716$ & $0.916930023$ & $0.93724605$ \\
        \hline & & 3 & $0.921196388$ & $0.936930023$ & $0.936930023$ \\
          \hline & live07 & 1 & $0.784763314$ & $0.837278107$ & $0.913091716$ \\

        \hline &  & 2 & $0.893491124$ & $0.92714497$ & $0.993713018$ \\
        \hline & & 3 & $0.962928994$ & $0.962928994$ & $0.962928994$ \\
        \hline & live08 & 1 & $0.882201204$ & $0.928362573$ & $0.961988304$ \\
        \hline & & 2 & $0.937134503$ & $0.956140351$ & $0.971247563$ \\
          \hline & & 3 & $0.961773879$ & $0.961773879$ & $0.961773879$ \\
        \hline & live11& 1 & $0.889074927$ & $0.916282964$ & $0.963583089$ \\
        \hline & & 2 & $0.955211386$ & $0.970280452$ & $0.9886982$ \\
        \hline &  & 3 & $0.979769778$ & $0.986513185$ & $0.986513185$ \\
        \hline & live12 & 1 & $0.870522227$ & $0.911955114$ & $0.945187743$ \\
        \hline & & 2 & $0.817404817$ & $0.842657343$ & $0.954251187$ \\
        \hline & & 3 & $0.891137566$ & $0.923534743$ & $0.923534743$ \\
        \hline & average  & 1 & 0.866621589 & 0.908453171 & 0.952706965 \\
        \hline & & 2 & 0.897984709 & 0.922630628 & 0.969031204 \\
        \hline & & 3 & 0.947361321 & 0.954219004 & 0.954219004 \\
        
        \hline
        \end{tabular}
        \caption{The result of recall of multiple cameras under multiple policies}
        \label{tab:6.16}
        \end{table}

        The test accuracy of the multi-camera detection by the appeal has reached the performance on the product side, so the need to test is in each program strategy, about a sequence, here a sequence of 450 frames, for the overall time consuming In the multi-GPU environment, we can basically see that Scenario 3 has a big advantage in the overall In a multi-GPU environment, we can basically see that Option 3 has a great advantage in terms of the overall time, because more tracking and less detection will reduce very little time, less detection of the whole picture, and only for the previous frame in a picture This can greatly reduce the speed and also avoid more false detections.

        After the accuracy is guaranteed, the speed of the multi-camera algorithm needs to be further tested. The evaluation of the speed of the multi-camera algorithm is shown in Table \ref{tab:6.17}, and the test is performed in a sequence of 450 frames, with a few live images from cam12 and the whole 36 cameras. The 36 cameras are parallelized on the GPU, so the time consumed by the 36 cameras is basically the same as the time consumed by each camera. The time consumption of 36 cameras is basically the same as the time consumption of each single camera. How to deploy the 36 cameras on the GPU for As shown in Table \ref{tab:6.17}, it is about 36ms per frame, which is basically the same as the time consumed by each single camera. As shown in Table \ref{tab:6.17}, it is about 36ms per frame, which basically achieves the real-time effect.

        \begin{table}
            \centering 
        \begin{tabular}{|l|l|l|l|}
        \hline methods & Method-1 & Method-2 & Method-3 \\
        \hline Cam12 processing time & $19135 \mathrm{~ms}$ & $22853 \mathrm{~ms}$ & $16400 \mathrm{~ms}$ \\
        \hline 36 camera whole processing time & $20000 \mathrm{~ms}$ & $26878 \mathrm{~ms}$ & $17564 \mathrm{~ms}$ \\
        \hline average frame processing time  & $41 \mathrm{~ms}$ & $55 \mathrm{~ms}$ & $36 \mathrm{~ms}$ \\
        \hline
        \end{tabular}
        \caption{Multi-camera test time consumption evaluation}
        \label{tab:6.17}
        \end{table}

        \subsection{Summary}
        
        This section begins with an introduction to the configuration and use environment of the soccer detection system with multiple cameras, while The environmental parameters for the system to be deployed are also expanded for the subsequent tests. Then, this chapter starts with a test of the detector accuracy and speed, then tests the operation of the single-camera detection module and the multi-camera detection module. and then tests the operation of the single camera detection module and the multi-camera detection module, and by showing the final related screenshots and test tables, the whole system testing is fully solution.

\newpage
\newgeometry{
	left=32mm, right=18mm, top=20mm, bottom=18mm,
	marginparwidth=28mm, marginparsep=4mm}
%%%%%%%%%%%%%%%%%%%%%%%%%%%%%%%%%%%%%%%%%%%%%%%%%%%%%%%%%%%

 \section{Summary and Outlook}
 \vspace{10.5cm}

 \subsection{Summary}
 With the booming wave of artificial intelligence in academia and industry, deep learning-based computer technology technology, this system is mainly to solve the 3D small target detection technology that appears in the traditional method with cost high, low accuracy and inconvenient disadvantages.

 The main objective of this paper is to explore and solve the problem of soccer ball detection under 36-eye camera. A target soccer ball with a ratio of about $2x10^{-5}$ to the original image is detected by multiple cameras and its position in 3D is determined, and the model is optimally compressed to achieve load reduction and accuracy and speed improvement. The main work and innovations of this paper are the following.

\begin{itemize}
    \item First, the research on the detection of small targets was investigated, and the methods were fused according to the relevant findings We designed and implemented a scheme for a soccer detector.
    \item A computer vision deep learning algorithm adapted to this project is used for each monocular camera to continuously optimize the the best results that can be achieved with monocular cameras.
    \item Using the detection results of multiple cameras and the beam flow difference method to optimize the position of the target in 3D, and finally, the image and model are compressed and optimized to achieve the accuracy of capturing the motion target in real time. Finally, the image and model are compressed and optimized to capture the motion target accurately in real time.
\end{itemize}

\subsection{Outlook}
For the design of this system, the purpose of the system is the live broadcast system of ball sports, in view of the author in doing football matches, but also in doing live rugby matches, for once there is a large number of masking problems exist. The system is designed for the purpose of the live broadcast system of ball sports, in view of the fact that the author is doing the live broadcast of football matches and rugby matches at the same time. The system can't meet the effect well, then from the whole system requirements analysis, outline design, detailed System design from the whole system requirements analysis, outline design, detailed design and system testing, there will still be shortcomings, I think in these aspects I believe that in these areas can also be improved.

\begin{itemize}
    \item Model aspect. The detector's model is not effective in detecting the prolonged occlusion situation for the target, then the target value is derived by determining the key points based on the athletes' motion information. Also continue to optimize the model to be portable to run independently in the embedded system.
    \item 3D visual inspection aspect. When 3D modeling is performed, with the development of artificial intelligence technology, in the later phase it should be recovered directly from the image to the object pose in 3D, i.e., 3D vision will be greatly enhanced.
    \item This system can further do rendering. And it can be applied to large sports events, while the method can be directly migrated to basketball smart live, rugby smart live and other ball games.
\end{itemize}

\newpage

%%%%%%%%%%%%%%%%%%%%%%%%%%%%%%%%%%%%%%%%%%%%%%%%%%%%%%%%%%%
\section{Referencing and Documentation}
 \vspace{10.5cm}

 [1] Harold W A. Aircraft Warning System, U. S. Patent 3053932, September 1962. 
 
 [2] Papageorgiou C P, Oren M, Poggio T. A General Framework for object detection. In:
Proceedings of the 6th IEEE International Conference on Computer Vision. Bombay,
India:IEEE, 1998.555-562.

 [3] Viola P,Jones M J.Robust Real-time Object Detection.International Journal of Computer
Vision,2001,4:51-52.

 [4] Lowe D G. Distinctive Image Features from Scale-invariant Keypoints.International
Journal of Computer Vision,2004,60(2):91–110.DOI:10.1023/B:VISI.0000029664.99615.

 [5] Dalal N, Triggs B. Histograms of Oriented Gradients for Human Detection.In:
Proceedings of the 2005 IEEE Computer Society Conference on Computer Vision and
Pattern Recognition.San Diego,CA,USA:IEEE,2005.886-893.

 [6] Felzenszwalb P F, Girshick R B, McAllester D, Ramanan D. Object Detection with
Discriminatively Trained Part-based Models.IEEE Transactions on Pattern Analysis and
Machine Intelligence,2010,32(9):1627–1645.DOI:10.1109/TPAMI.2009.167.

 [7] Everingham M, Van Gool L, Williams C K I, Winn J, Zisserman A. The Pascal
Visual Object Classes (VOC) Challenge. International Journal of Computer Vision,2010,
88(2):303–338. DOI:10.1007/s11263-009-0275-4.

 [8] Alex Krizhevsky, Ilya Sutskever, and Geoffrey E Hinton. Imagenet Classifi-cation with
Deep Convolutional Neural Networks. In Advances in Neural Information Processing
Systems,pages1097-1105,2012.

 [9] Vedaldi A,Lenc K. MatConvNet-convolutional Neural Networks for MATLAB.arXiv:141
2.4564,2014.

 [10] Goodfellow I J, Warde-Farley D, Lamblin P, Dumoulin V, Mirza M, Pascanu R, Berg
J, Bastien F, Bengio Y. Pylearn2:A Machine Learning Research Library.arXiv:1308. 421
4,2013.

 [11] The Theano Development Team. Theano:A Python Framework for Fast Computation
of Mathematical Expressions. arXiv:1605.02688,2016.

[12] The Visual Object Tracking VOT2014 Challenge Results.In:Proceedings of The
European Conference on Computer Vision (ECCV 2014), Lecture Notes in Computer
Science. Zurich,Switzerland:Springer International Publishing,2015.191-217.

[13] Yin Hongpeng,Chen Bo,Chai Yi,Liu Zhaodong. A review of vision-based target detection and tracking[J]. Journal of Automation,2016,42(10):1466-1489.Doi:10.16383/j.aas.2016.c150823.

[14] TAN Min,WANG Shuo. Advances in robotics research[J]. Journal of Automation,2013,39(7):963-972.

[15] Li Yubo,Zhu Xuzhou,Lu Huimin,Zhang Hui. A Review of Visual Odometry Technology,Computer Application Research012,08:2801-2805
-2810.

[16] Krizhevsky A,Sutskever I,Hinton G E.Imagenet Classification with Deep Convolutional
Neural Networks[C]//Advances in Neural Information Processing Systems.2012:1097-11
05.

[17] Sun Y,Wang X,Tang X.Deep Learning Face Representation from Predicting 10,000 Classes[C]//Computer Vision and Pattern Recognition (CVPR),2014 IEEE Conference on.IEEE, 2014: 1891-1898.

[18] Kaiming He:Deep Learning for Object Detection,CVPR2017;Detection tutorial

[19] REDMON J,DIVVALA S,GIRSHICK R,et al.You Only Look Once:Unified,Real-Time
Object Detection[J],2015,:779-788.

[20] Liu W,Anguelov D,Erhan D, et al.SSD:Single Shot MultiBox Detector[J],2016,6321-3.

[21] Donggeun Yoo,Sunggyun Park,et al.AttentionNet: Aggregating Weak Directions for
Accurate Object Detection,2015.

[22] REDMON J,DIVVALA S,GIRSHICK R,et al.YOLOv3:An Incremental Improvement,
2018,CVPR2018.

[23] Xie, Feng-Ying. Digital Image Processing and Applications: Electronic Industry Press, 2014.

[24] Pedro Felzenszwalb.Object detection with discriminatively trained partbased models.
IEEE Trans.PAMI,32(9):1627-1645,2010.

[25] Hartley,Richard,and Andrew Zisserman  Author. Multiview Geometry in Computer Vision.Cambridge University Press,2003.159~161.

[26] Triggs,Bill,et al."Bundle adjustment—A Modern Synthesis."International workshop on
vision algorithms.Springer Berlin Heidelberg,1999.

[27] Liu Kaiyu, Xia Bin. Real-time human pose recognition based on Kinect [J]. Electronic Design Engineering 2014(19):31-34.

[28]  Li Jiusheng,Li Yongqiang,Zhou Di. Consistency analysis of EKF-based SLAM algorithms[J]. Computer Simulation,2008,06:155-160.

[29] Grisetti G,Stachniss C,Burgard W.Improved Techniques for Grid Mapping with Rao
blackwellized Particle Filters[J].Robotics,IEEE Transactions on,2007,23(1):34-46.

[30] Gil Briskin:Estimating Pose and Motion using Bundle Adjustment and Digital
Elevation Model Constraints,Technion-Computer Science Department-M.Sc.Thesis MSC-2014-16-2014.

[31] UIJLINGS J R,SANDE K E,GEVERS T,et al.Selective Search for Object Recognition
[J].International Journal of Computer Vision,2013,104(2):154-171.

[32] Jia Yangqing,Shelhamer,Evan,Donahue,Jeff,Karayev,Sergey,Long,Jonathan,Girshick,Ross,
Guadarrama,Sergio,andDarrell,Trevor.Caffe:Convolutional Architecture for Fast Feature
Embedding.arXiv:1408.5093,2014.

[33] Song Han,Huizi Mao,William J.Dally:Deep Compression:Compressing Deep Neura-l
Networks with Pruning,Trained Quantization and Huffman Coding,ICLR 2016.

[34] Chen,Wenlin,Wilson,James T,Tyree, Stephen, Weinberger, Kilian Q., and Chen, Yixin.
Compressing Neural Networks with the Hashing Trick.arXiv:1504.04788,2015.

[35] Kai Chen,Jiaqi Wang1.et al:Optimizing Video Object Detection via a Scale-Time
Latice,CVPR2018.

[36] Van Nguyen, Hien, Zhou, Kevin, and Vemulapalli, Raviteja. Cross-domain synthesis
of medical images using efficient location-sensitive deep network. In Medical Image
Computing and ComputerAssisted Intervention–MICCAI 2015,pp.677–684.Springer,2015.

[37] Arora, Sanjeev, Bhaskara, Aditya, Ge, Rong, and Ma, Tengyu. Provable bounds for
learning some deep representations.In Proceedings of the 31th International Conference
on Machine Learning,ICML 2014,pp.584-592,2014.

[38] REN S,HE K,GIRSHICK R, et al.Faster R-CNN: Towards Real-Time Object Detection with Region Proposal Networks[J]. IEEE Transactions on Pattern Analysis \& Machine Intelligence 2017 39(6): 1137-1149.

[39] Xiaogang Wang:Deep Learning for Object Detection in Videos, CVPR2016 Detection
tutorial, 2017.

[40] Pierre Baque´:Deep Occlusion Reasoning for Multi-camera Multi-Target Detection,
CVPR2018,2018.

[41] Wout Oude Elferink*:Multi-camera Tracking of Soccer Players Through Severe
Occlusions W.G.Oude Elferink M.Sc.IEEE Transactions on Pattern Analysis \& Machine
Intelligence 2017,39(6):1137-1115.

\newpage
\end{document}